%% file: main.tex
\pdfoutput=1  

\documentclass[runningheads]{llncs}

\usepackage[final,year=2026]{eccv}

\input{preamble}

\usepackage[breaklinks]{hyperref}
\usepackage{orcidlink}

\title{Intrinsic PAPR: Tackling Misattribution in 3D Intrinsic Decomposition via Proximity Attention Point Rendering}

\titlerunning{Intrinsic PAPR}

\author{Alireza Moazeni\inst{1}\orcidlink{0009-0002-6532-0112}\thanks{Corresponding author.} \and
Shichong Peng\inst{1}\orcidlink{0009-0005-8404-6392} \and
Yanshu Zhang\inst{1}\orcidlink{0009-0002-6046-3966} \and
Chirag Vashist\inst{1}\orcidlink{0000-0003-3270-1533} \and
Ke Li\inst{1,2,3}\orcidlink{0000-0002-3229-271X}}
\authorrunning{A. Moazeni et al.}
\institute{APEX Lab, School of Computing Science, Simon Fraser University, Canada \and
Alberta Machine Intelligence Institute (Amii), Canada \and
Canadian Institute for Advanced Research (CIFAR), Canada\\
\email{\{sam62, shichong\_peng, yanshu\_zhang, chirag\_vashist, keli\}@sfu.ca}}

\begin{document}

\input{arxiv_figuse}\maketitle

\input{tex/0_abstract}

\input{tex/1_intro}

\input{tex/2_related_work}
\input{tex/3_method}

\input{tex/4_experiments}

\input{tex/5_ablation_study}
\input{tex/6_discussion_and_conclusion}

\input{tex/8_ack}

\clearpage
\bibliographystyle{splncs04}
\bibliography{main}

\end{document}


\input{arxiv_figuse}\maketitle

\noindent
This document provides extended implementation details, additional quantitative and qualitative results, and further ablations for Intrinsic PAPR, complementing the main paper.

\setcounter{tocdepth}{2}
\tableofcontents
\clearpage

\input{tex/7_suppl}

\clearpage

\bibliographystyle{splncs04}
\bibliography{main}

%% file: preamble.tex
\AtBeginDocument{\raggedbottom}

\usepackage{multirow}
\usepackage{graphicx}
\usepackage{booktabs}

\providecommand{\Description}[1]{}

\IfFileExists{eccvabbrv.sty}{%
  \usepackage{eccvabbrv}%
}{%
}

\definecolor{appGreen}{RGB}{34,139,34}
\definecolor{appPurple}{RGB}{128,0,128}
\definecolor{appOrange}{RGB}{230,124,34}
\definecolor{appYellow}{RGB}{204,153,0}
\newcommand{\appgreen}[1]{{\color{appGreen}#1}}
\newcommand{\apppurple}[1]{{\color{appPurple}#1}}
\newcommand{\apporange}[1]{{\color{appOrange}#1}}
\newcommand{\appyellow}[1]{{\color{appYellow}#1}}
\definecolor{appBlue}{RGB}{0,102,204}
\newcommand{\appblue}[1]{{\color{appBlue}#1}}

%% file: tex/0_abstract.tex
\begin{abstract}
Recent point-based intrinsic decomposition and inverse rendering methods have advanced the modelling of the shading and albedo of 3D scenes. However, we identify a fundamental limitation: these methods suffer from a \textit{misattribution issue}, where individual primitives learn incorrect appearance features despite producing correct aggregated renderings. We show that the root cause lies in volume rendering, which aggregates translucent primitives along each ray and only supervises the final colour, preventing direct supervision of individual primitive features. To address this, we propose \textit{Intrinsic PAPR}, a robust intrinsic decomposition framework which leverages Proximity Attention Point Rendering (PAPR) to enable direct per-point supervision. Unlike volume rendering approaches, PAPR eliminates translucent primitives and directly predicts appearance at ray-surface intersections, enabling accurate supervision to the feature of each individual point. Our method incorporates a 2D albedo prior adapted with conditional Implicit Maximum Likelihood Estimation (cIMLE) to handle monocular ambiguities, and employs a space carving loss to ensure multi-view consistency. Extensive evaluations on synthetic and real-world datasets demonstrate that Intrinsic PAPR outperforms point-based inverse rendering, NeRF-based intrinsic decomposition, and diffusion-based PBR methods in novel view synthesis and albedo estimation while resolving the misattribution issue.

\keywords{Neural Rendering \and Intrinsic Decomposition \and Inverse Rendering \and Point-Based Rendering}
\end{abstract}

%% file: tex/1_intro.tex
\section{Introduction}
\label{sec:introduction}
\input{figs/main/teaser/teaser}
\input{figs/main/motivation/motivation_fig}

Recent advances in point-based neural rendering, such as 3D Gaussian Splatting~\cite{kerbl20233d}, have achieved impressive quality in novel view synthesis. However, these methods typically encode only the final emitted colour per primitive, limiting editing applications that require modifying scene properties like material albedo or shading. To enable such control, recent works have explored intrinsic decomposition~\cite{Xing2023IntrinsicAD} and inverse rendering~\cite{Liang2023GSIR3G}, which learn to represent distinct material and lighting components.

Surprisingly, applying edits to these decomposed models often yields unexpected and inconsistent results. As shown in Figure~\ref{fig:motivation_figure}a, if we copy the learnt albedo features from a source region to a target, the rendered target albedo fails to match the source. This indicates that the underlying primitives have learned incorrect albedo features, a phenomenon we term the \textbf{misattribution issue}, a critical barrier to creating robust and predictable 3D editing tools.

Our first key contribution is identifying the root cause of misattribution: the aggregation of semi-transparent primitives in volume rendering. In splat-based methods~\cite{kerbl20233d,Liang2023GSIR3G}, primitives are constrained to ellipsoid shapes, so they must be made translucent and overlaid in various orientations to yield near-opaque, non-ellipsoid regions. Volume rendering then composites the features of all primitives along a ray. This allows ambiguity: a set of primitives can learn incorrect individual features that, when aggregated, still produce the correct final pixel colour. Since no single primitive is directly supervised by the output pixel, there is no guarantee that individual features are correct. Notably, this ambiguity is not cured by capturing more views---near-coincident primitives render identically from every direction---a prediction we confirm empirically in Sec.~\ref{sec:experiments-intro}.

Our second key contribution is the insight that this issue can be resolved by fundamentally changing the rendering paradigm. We propose repurposing Proximity Attention Point Rendering (PAPR)~\cite{zhang2023papr}, which avoids the pitfalls of volumetric aggregation. Unlike splat-based primitives, PAPR's primitives are points without spatial extent; instead, it forms an opaque surface by connecting nearby points with a learned interpolation kernel, which takes the form of an attention mechanism, thereby eliminating the need for translucency. This approach allows PAPR to predict appearance directly at ray-surface intersections, rather than aggregating features along a ray. Furthermore, because PAPR learns a parsimonious point cloud, each point influences the rendered colour of at least one pixel, so its features are more directly supervised by the image, preventing incorrect, compensatory features and resolving the misattribution issue (Figure~\ref{fig:motivation_figure}b).

Building on this insight, we introduce \textbf{Intrinsic PAPR} (Figure~\ref{fig:teaser}), a novel framework for robust intrinsic decomposition and editable 3D appearance. Our method learns per-point albedo and shading features suitable for high-quality reflectance and illumination editing. Our third contribution is a novel technique to supervise this decomposition robustly. We use a pre-trained 2D intrinsic decomposition model~\cite{Careaga2024Colorful} as a prior, but a single image-to-albedo prediction is often ambiguous. We address this by adapting the 2D prior into a probabilistic model with conditional Implicit Maximum Likelihood Estimation (cIMLE)~\cite{Li2020MultimodalIS,peng2022chimle}, generating a distribution of plausible albedos per view. We use a mode selection loss akin to the space carving loss from~\cite{uy2023scade} to distill these multi-view distributions into a single, geometrically consistent 3D albedo representation.

While state-of-the-art inverse rendering methods increasingly rely on complex physically-based rendering (PBR) models to estimate SVBRDF parameters, these formulations are highly ill-posed and often fail to reliably estimate editable parameters in casual multi-view captures. We deliberately adopt a simpler albedo-shading intrinsic decomposition, avoiding the under-constrained nature of full physical inverse rendering. A richer decomposition can represent more complex effects but cannot reliably recover them from observations; by the bias-variance trade-off, for editing it is better to recover a few parameters accurately than many inaccurately. Our decomposition acts as an effective geometric regularizer, forcing the model to explain illumination effects through accurate geometry rather than baking them into the albedo. This not only improves decomposition but also leads to higher-quality novel view synthesis than the original PAPR.

We evaluate Intrinsic PAPR against state-of-the-art methods on synthetic and real-world datasets, demonstrating superior performance in both novel view synthesis and albedo estimation, with a $10.3\%$ and $8.0\%$ PSNR improvement, respectively, over the best inverse-rendering baseline. Quantitative feature-transfer experiments further confirm that our method mitigates the misattribution issue, learning more accurate and consistent per-primitive features and achieving more than a $4\times$ reduction in albedo- and shading-transfer error over a leading 3DGS-based method.

%% file: figs/main/teaser/teaser.tex
\newlength{\teasersubfigheight}
\newlength{\teaserimgwidth}
\setlength{\teasersubfigheight}{0.198\textheight}

\newcommand{\teasergrid}[2]{%
\begin{figure*}[t]
  \centering
  \setlength{\fboxsep}{0pt}%
  \setlength{\fboxrule}{1pt}%
  \setlength{\teaserimgwidth}{\dimexpr0.44\linewidth-2\fboxrule\relax}%
  \begin{subfigure}[t]{0.44\linewidth}
    \centering
    \fcolorbox{red}{white}{\includegraphics[width=\teaserimgwidth,height=\teasersubfigheight]{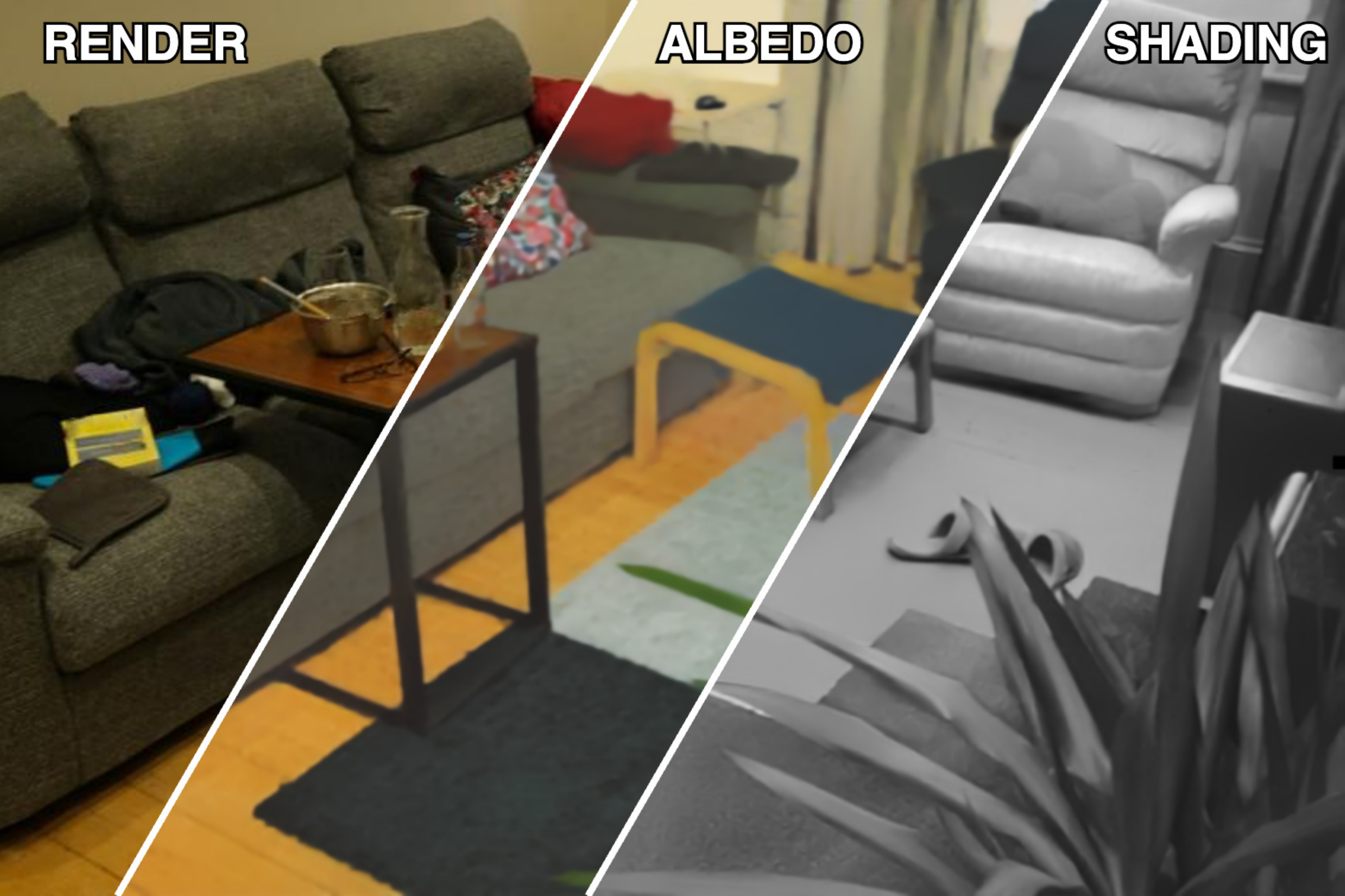}}
  \end{subfigure}
  \begin{subfigure}[t]{0.44\linewidth}
    \centering
    \fcolorbox{green}{white}{\includegraphics[width=\teaserimgwidth,height=\teasersubfigheight]{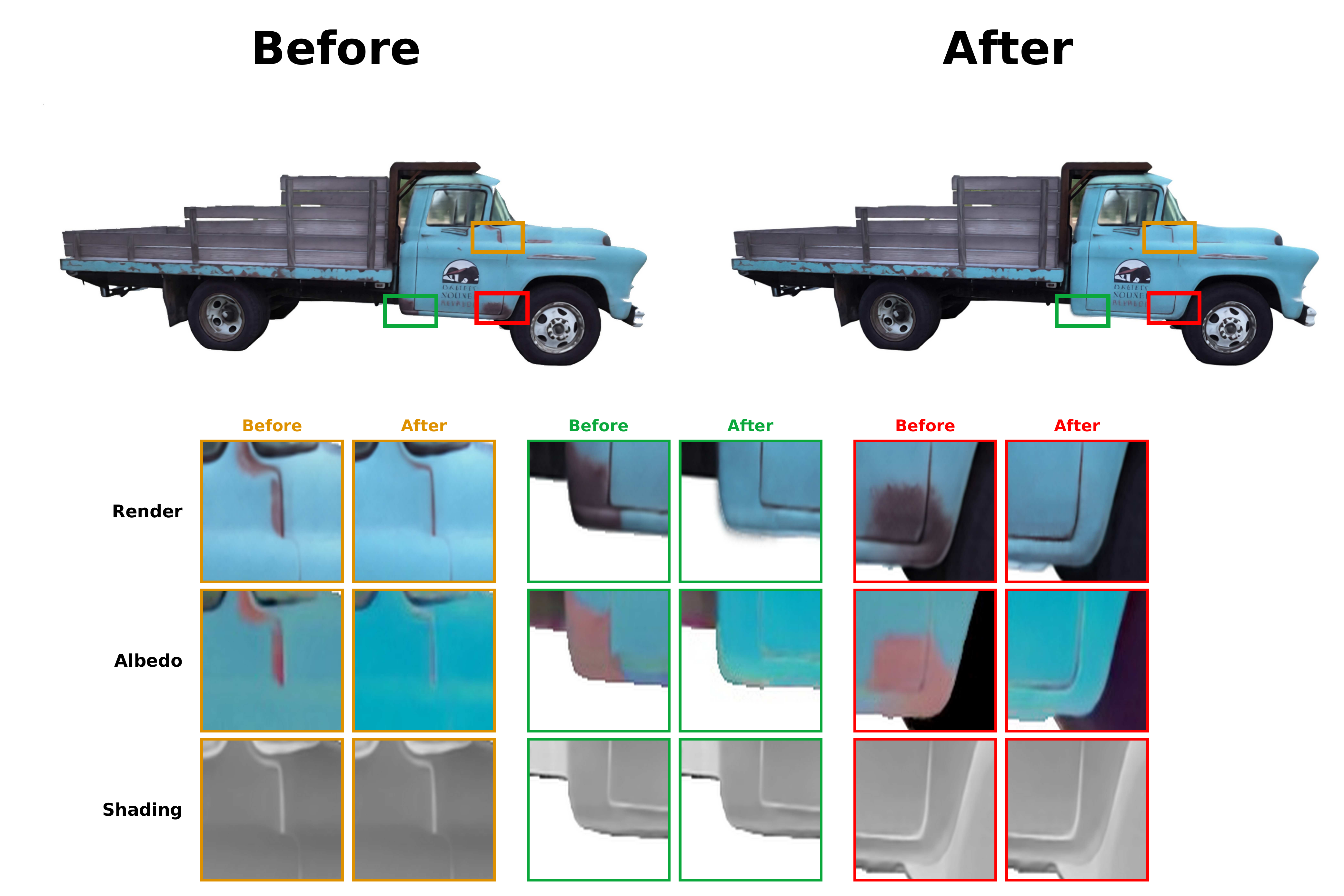}}
  \end{subfigure}
  \\
  \begin{subfigure}[t]{0.44\linewidth}
    \centering
    \fcolorbox{blue}{white}{\includegraphics[width=\teaserimgwidth,height=\teasersubfigheight]{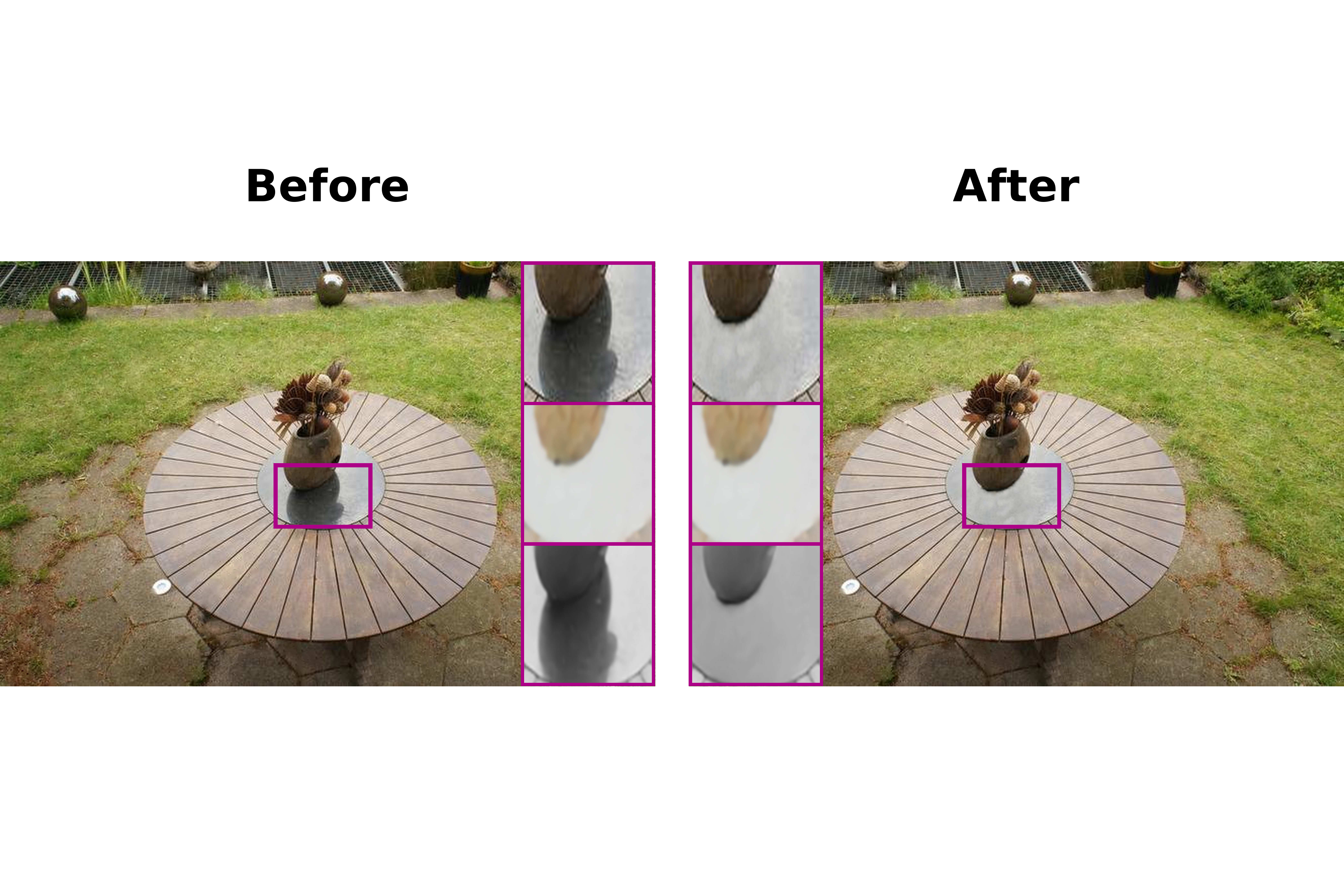}}
  \end{subfigure}
  \begin{subfigure}[t]{0.44\linewidth}
    \centering
    \fcolorbox{yellow}{white}{\includegraphics[width=\teaserimgwidth,height=\teasersubfigheight]{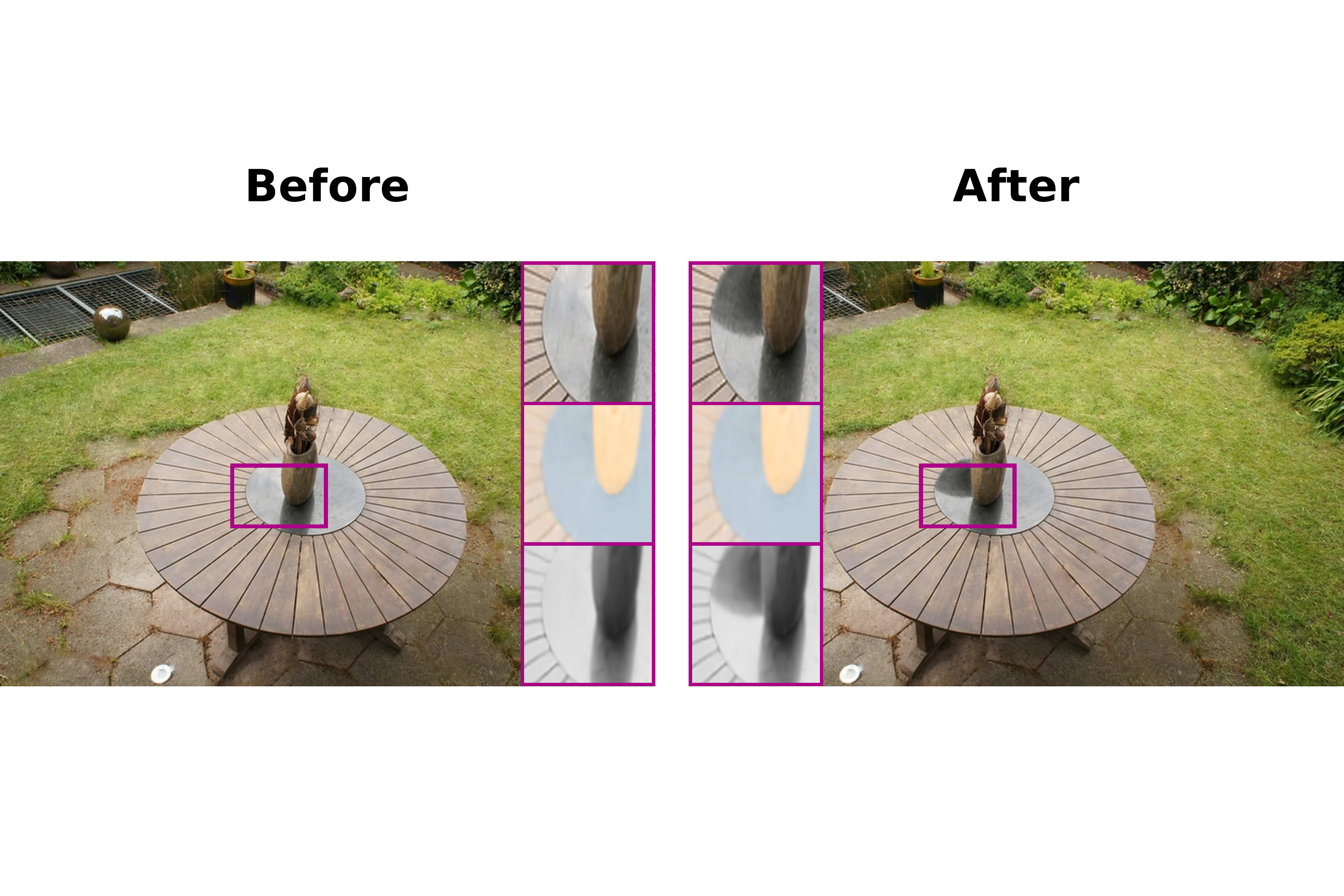}}
  \end{subfigure}
  \caption{#2}
  \Description{Teaser on real scenes: intrinsic decomposition (render, albedo, shading); freeform albedo editing (rust removal on a truck); user-defined shading removal and addition. Each editing panel shows large before/after images with coloured region boxes and per-region before/after render, albedo and shading crops.}
  \label{#1}
\end{figure*}
}

\newcommand{\teasercaptionfull}{We present \textbf{\textit{Intrinsic PAPR}}, which tackles misattribution in 3D intrinsic decomposition and enables robust editing on real-world and synthetic scenes.
\textcolor{red}{\textbf{(Top-Left)}} Our method accurately decomposes a 3D scene into multi-view consistent albedo and shading components.
It supports \textcolor[rgb]{0,0.5,0}{\textbf{(Top-Right)}} freeform albedo editing (e.g., rust removal on a truck) and allows geometry-aware shading manipulation, seamlessly removing \textcolor{blue}{\textbf{(Bottom-Left)}} or adding \textcolor[rgb]{0.7,0.7,0}{\textbf{(Bottom-Right)}} shadows in user-defined 3D regions while preserving the underlying scene structure. Each editing panel shows the large \emph{before} and \emph{after} images with coloured region boxes, and per-region \emph{before}/\emph{after} render, albedo and shading crops.}

\teasergrid{fig:teaser}{\teasercaptionfull}

%% file: figs/main/motivation/motivation_fig.tex
\begin{figure}[t]
  \centering
  \includegraphics[width=0.8\linewidth]{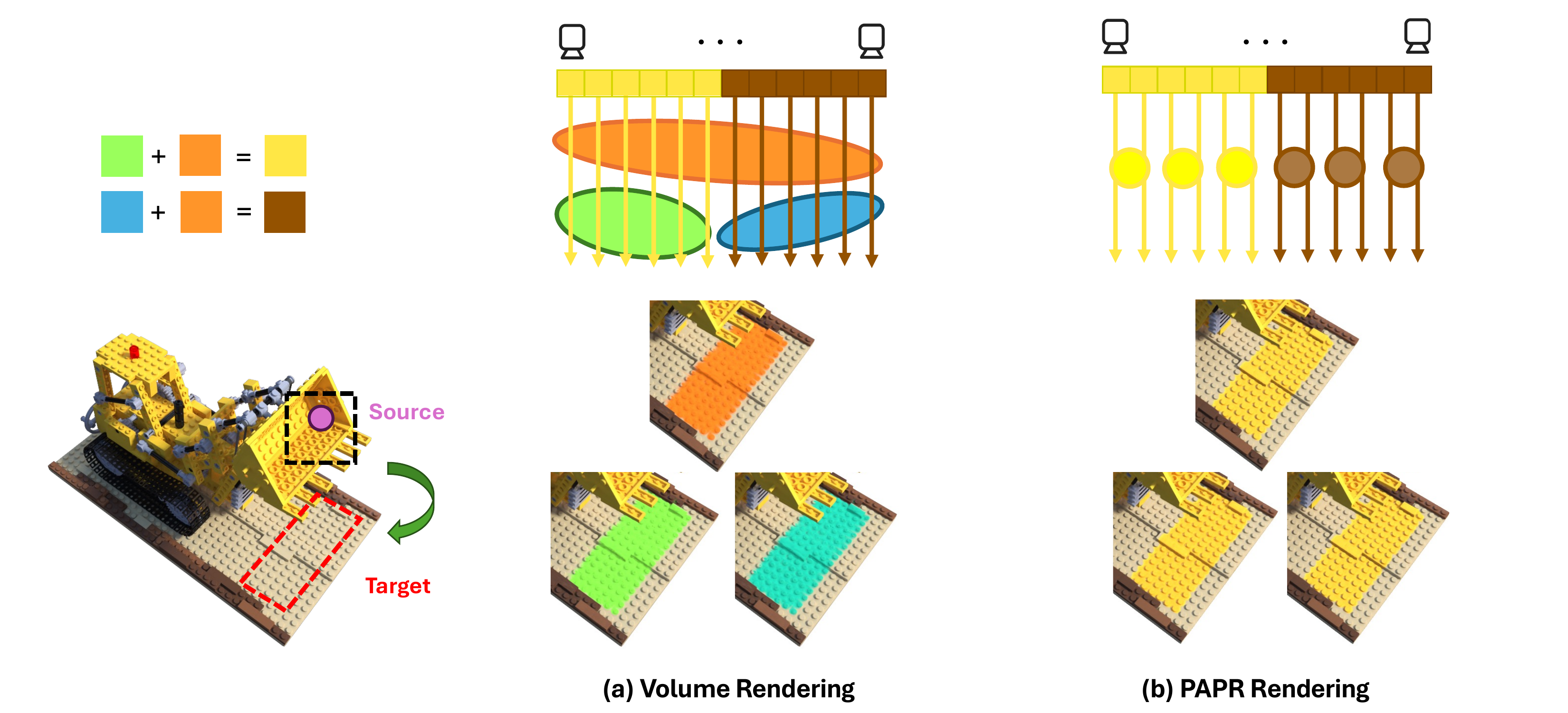}
  \caption{The misattribution issue in point-based rendering. (a) When using volume rendering with translucent primitives, individual primitives can learn incorrect features while still producing plausible results. (b) Our approach using PAPR ensures accurate per-point feature learning by directly supervising the features at each point.}
  \Description{Two side-by-side illustrations showing the misattribution issue in point-based rendering. The left image (a) demonstrates how volume rendering with translucent primitives can lead to incorrect feature learning, while the right image (b) shows how our PAPR approach ensures accurate feature learning through direct supervision.}
  \label{fig:motivation_figure}
\end{figure} 

%% file: tex/2_related_work.tex
\section{Related Work}
\label{sec:related_work}
\subsection{Neural Scene Representations}
In recent years, there has been a surge in interest in using neural networks for novel view synthesis from multi-view RGB images~\cite{Mildenhall2020NeRFRS,Yu2021PlenoxelsRF,Mller2022InstantNG,chen2022tensorrf,Xu2022PointNeRFPN,kerbl20233d,zhang2023papr}. Here, we discuss two widely-used scene representations relevant to our setting: NeRF-based and point-based representations. 

NeRF~\cite{Mildenhall2020NeRFRS} achieves impressive rendering quality but is computationally costly, evaluating multiple samples per ray. Precise point-level editing in NeRF is also non-trivial, requiring changes to scene properties at infinitely many points within a volume and careful specification of that volume's boundary.

In contrast, point-based representations render with fewer samples and gain increasing attention for their high rendering quality and efficiency~\cite{Lassner2021PulsarES,Ost2021NeuralPL,Zuo2022ViewSW,Zhang2022DifferentiablePR,kerbl20233d,zhang2023papr}. Because they store local scene properties at each point, precise point-level edits simply modify the properties stored at points within the desired region. Among these, splat-based techniques~\cite{Lassner2021PulsarES,Zuo2022ViewSW,Zhang2022DifferentiablePR,kerbl20233d} use radial basis functions to compute point contributions to rays and volume rendering to combine the per-point scene properties into the output. Another line of works uses attention-based model to interpolate properties stored at neighbouring surface points for rendering~\cite{Ost2021NeuralPL,zhang2023papr}. In particular, Proximity Attention Point Rendering (PAPR)~\cite{zhang2023papr} stands out for its ability to learn a parsimonious point cloud from scratch. In this work, we adopt PAPR as our scene representation and rendering method. 

\subsection{Intrinsic Decomposition}
The problem of image intrinsic decomposition aims to decompose images into reflectance and shading components. Early methods focused on predicting ordinal relationships between the albedos of pixel pairs~\cite{Narihira2015LearningLF,Zhou2015LearningDR,Zoran2015LearningOR}, while later methods directly regress to continuous shading and albedo values~\cite{Li2019InverseRF,Luo2020NIIDNetAS,Sengupta2019NeuralIR,Zhou2019GLoSHGS,Zhu2022IRISformerDV,Das2022PIENetPI}. Recent single-image intrinsic methods also leverage diffusion priors for material/intrinsic estimation~\cite{kocsis2024intrinsicdiffusion}. For evaluation, the IIW dataset and WHDR metric~\cite{bell2014iiw} remain standard for assessing reflectance ordering and can complement our multi-view albedo consistency analysis.

Whereas these methods consider 2D inputs, PoInt-Net~\cite{Xing2023IntrinsicAD} performs intrinsic decomposition on RGB-D. A major challenge is generalization to real imagery; Careaga and Aksoy first propose an ordinal-shading model~\cite{Careaga2023IntrinsicID} and then a colourful diffuse intrinsic decomposition~\cite{Careaga2024Colorful}, combining synthetic pretraining with multi-illumination data to generate dense pseudo-ground-truth intrinsic components; we use the latest version as our 2D prior. Multi-view intrinsic transformers such as IDT~\cite{idt2025} instead enforce consistency directly in the 2D image domain; in contrast, we distill ambiguity-aware 2D priors into a unified 3D point-based representation.

For 3D settings, intrinsic decomposition of neural scenes~\cite{wang2023fegr,ye2023intrinsicnerf,Sarkar2023LitNeRFIR,zhang2021nerfactor} estimates albedo and shading jointly with geometry. IntrinsicNeRF~\cite{ye2023intrinsicnerf} uses iterative reflectance clustering; LitNeRF~\cite{Sarkar2023LitNeRFIR} targets faces with few views. In the Gaussian Splatting family, Intrinsic-GS~\cite{intrinsicgs2024} performs multi-view intrinsic decomposition using colour-/shading-invariant priors to stabilize albedo across views. However, NeRF-/GS-based volume aggregation often encodes scenes in networks or splats, making localized edits challenging and susceptible to misattribution. Other works~\cite{Wu2022PaletteNeRFPC,wang2023seal3d,lee2023icenerf,radl2024laenerf} enable colour edits without explicit shading, which can lead to visually inconsistent edits under illumination changes.

A common limitation of these methods is that they either operate in 2D (lacking multi-view geometric consistency) or, when extended to 3D via NeRF or GS, remain susceptible to misattribution from volumetric feature aggregation. Our method bridges this gap with a point-attention framework (PAPR) that stores albedo and shading per point and aggregates them without volumetric blending. We supervise with a 2D intrinsic prior adapted via cIMLE and distill it across views into a single 3D representation, yielding multi-view consistent albedo and shading.

\subsection{Inverse Rendering}
Inverse rendering recovers geometry, materials, and illumination from images. NeRF-based approaches~\cite{Srinivasan2020NeRVNR,verbin2022refnerf,Zhang2022ModelingII,jin2023tensoir,chen2024intrinsicanything} model view-dependent reflectance, sometimes incorporating diffusion priors, while explicit/hybrid methods such as nvdiffrec~\cite{munkberg2022extracting} and Mitsuba~2~\cite{nimier2019mitsuba2} optimize meshes, SVBRDFs, and lighting via differentiable path tracing. NeILF/NeILF++~\cite{yao2022neilf,zhang2023neilfpp} improve disentanglement by explicitly modelling incident light fields. 

In point-/Gaussian-based IR, several methods operate in 3DGS/point primitives, including GS-IR~\cite{Liang2023GSIR3G}, DPIR~\cite{chung2024differentiable}, and GaussianShader~\cite{jiang2023gaussianshader}. Recent works emphasize robust lighting (VMINer~\cite{fei2024vminer}), glossy materials~\cite{wang2024glossypleno}, and relightable Gaussians with BRDFs~\cite{gao2024relightable3dgaussians}. Building on 3DGS, IRGS~\cite{irgs2025} introduces a differentiable 2D Gaussian ray tracer for full inter-reflections under the complete rendering equation; DiscretizedSDF~\cite{discretizedsdf2025} encodes a discretized signed distance field inside each Gaussian for relightable assets; SVG-IR~\cite{svgir2025} equips Gaussians with spatially varying material and normal attributes. GAINS~\cite{gains2025} leverages intrinsic-image and diffusion priors to stabilize inverse rendering from sparse multi-view captures, and GS-ID~\cite{gsid2025} focuses on illumination decomposition on Gaussians with diffusion-guided material priors. 

Domain-specific extensions include physically-based human IR~\cite{wang2024intrinsicavatar}, robust shape/illumination recovery~\cite{shinobi2024}, single-shot material estimation~\cite{merlin2024}, and controllable GS editing without explicit decomposition~\cite{wang2024gaussianeditor,chen2023gaussianeditorSwift}.

While these IR approaches enable component-wise editing, the estimation problem is highly ill-posed and typically relies on strong priors, controlled capture (e.g., OLAT), or restrictive assumptions that limit robustness in casual multi-view capture~\cite{bi2024rgs}. In contrast, we deliberately target a simpler intrinsic factorization (per-point albedo and shading) with a surface-attentional renderer (PAPR), trading physical completeness for robustness and editability; this rendering shift is not merely a design preference but a structural prerequisite for resolving misattribution. We leverage an ambiguity-aware 2D intrinsic prior distilled into a multi-view-consistent 3D representation that favours stable point-level edits with predictable behaviour across views. 

%% file: tex/3_method.tex
\section{Intrinsic PAPR}
\label{sec:intrinsic-papr}
\subsection{Representation and Rendering}
\label{sec:repr-render}
We build on Proximity Attention Point Rendering (PAPR)~\cite{zhang2023papr}. To enable intrinsic decomposition, we augment PAPR's per-point features by partitioning them into two orthogonal components: an albedo component capturing reflectance and a shading component capturing illumination. Each point thus stores separate albedo and shading information while retaining local scene detail, which is essential for downstream editing. Figure~\ref{fig:method_1_pipeline_figure} presents an overview of our rendering pipeline.

Specifically, our representation is denoted as $P = \left\{ \left( \mathbf{p}_i, \mathbf{a}_i, \mathbf{h}_i, \tau_i \right) \right\}_{i=1}^N$, where $\mathbf{a}_i \in \mathbb{R}^n$ represents the albedo feature vector and $\mathbf{h}_i \in \mathbb{R}^m$ represents the shading feature vector for each point. These two feature vectors serve as inputs to their corresponding value branches in the attention mechanism to produce an albedo feature map $A_{\mathcal{C}} \in \mathbb{R}^{H\times W\times d_\text{albedo}}$ and a shading feature map $S_{\mathcal{C}} \in \mathbb{R}^{H\times W\times d_\text{shading}}$.

Given a camera $\mathcal{C}$ and pixel $(i,j)$ with ray $r_{ij}$, we follow PAPR to select a small top-$K$ set of points $\mathcal{N}(r_{ij})$ nearest to the ray, compute proximity-based attention scores, and aggregate features to obtain per-pixel albedo and shading features:
\begin{align}
    s_k^{ij} &= f_\theta\!\big(\mathbf{p}_k,\, r_{ij},\, \tau_k\big), \quad
    w_k^{ij} = \mathrm{softmax}_k\!\left(s_k^{ij}\right), \nonumber \\
    A_{\mathcal{C}}^{ij} &= \sum\nolimits_{k \in \mathcal{N}(r_{ij})} w_k^{ij}\,\mathbf{a}_k,\quad
    S_{\mathcal{C}}^{ij} = \sum\nolimits_{k \in \mathcal{N}(r_{ij})} w_k^{ij}\,\mathbf{h}_k. \label{eq:attn-agg}
\end{align}
Here $f_\theta$ is PAPR's learnable scoring function, mapping the point position $\mathbf{p}_k$ and its perpendicular and along-ray distances to the ray $r_{ij}$ to a query--key attention score on positional encodings, re-weighted by the per-point influence $\tau_k$~\cite{zhang2023papr}; $\mathcal{N}(r_{ij})$ is the top-$K$ set selected by perpendicular point--ray distance, and the weights are softmax-normalized across it.

Under proximity attention, the score increases as the perpendicular point--ray distance shrinks. Consequently, each surface point admits a non-trivial cone of rays for which its normalized weight dominates ($w_k^{ij} \approx 1$). On those rays, the photometric residual backpropagates primarily to that point, making the features of every point identifiable. In contrast, volumetric splatting distributes transmittance multiplicatively along the ray: supervision constrains only the integrated colour, so the feature of each individual primitive can be incorrect as long as the aggregated colour is correct, since the error in one primitive can compensate for the errors of others and cancel out. The geometric locality of PAPR, together with a parsimonious point set, suppresses such compensations and thereby mitigates misattribution (see Figure~\ref{fig:motivation_figure}).

\input{figs/main/method_1_pipeline/pipeline_fig}

Following the intrinsic factorization $\mathbf{I} \approx \mathbf{A} \odot \mathbf{S}$, we adopt a minimalist supervision strategy: supervise RGB and albedo, and learn shading implicitly. This is sufficient by construction: under the factorization, any two of $(\mathbf{I}, \mathbf{A}, \mathbf{S})$ determine the third, so constraining $\mathbf{I}$ and $\mathbf{A}$ fully determines $\mathbf{S}$ without an additional shading loss. Moreover, the rendering loss enforces multi-view RGB consistency and the space carving loss enforces multi-view albedo consistency, so the implicit shading is forced to be stable and view-consistent rather than an unconstrained parameter absorbing residual colours. Direct shading supervision would moreover be ill-posed in regions where $\mathbf{A} \approx \mathbf{0}$, because many values of $\mathbf{S}$ yield the same product. To supervise the rendered image, we reuse PAPR's reconstruction objective (MSE+LPIPS) given by:
\begin{align}
    \mathcal{L}_{\text {recon }} = \alpha\, \mathrm{MSE}\!\left(\hat{\mathbf{I}}, \mathbf{I}_{gt}\right) + \beta\, \mathrm{LPIPS}\!\left(\hat{\mathbf{I}}, \mathbf{I}_{gt}\right).
    \label{eq:papr_loss_fn}
\end{align}
To supervise the decomposed reflectance, we augment this objective with a scale-invariant albedo term, introduced next.
\subsection{Scale-Invariant Albedo Loss}
\label{sec:handling-scale-variations}
In image intrinsic decomposition, the albedo is defined only up to a multiplicative scale: a decomposition into albedo $A$ and shading $S$ is as valid as one into $\gamma A$ and $S/\gamma$ for any positive $\gamma$. To make our albedo loss scale-invariant, we introduce a learnable scalar $\lambda$ per training view. Before computing the loss between the predicted albedo $\hat{\mathbf{A}}$ and the pseudo-ground-truth albedo $\mathbf{A}_{p-\text{gt}}$, we scale $\mathbf{A}_{p-\text{gt}}$ by its $\lambda$. Jointly optimizing these scalars with the model parameters aligns the pseudo-ground-truth and predicted scales, ensuring scale-invariant albedo across views.

\subsection{Ambiguity-aware Albedo Estimates}
In single-view 2D image intrinsic decomposition, the absence of geometric information introduces inherent ambiguities in determining albedo; different reflectance properties can explain the same observed image, leading to multiple plausible albedo interpretations. For instance, the dotted pattern on the baseboard of a Lego bulldozer could be shadows cast by raised dots, or a flat board with a dotted texture.

Since the prior model produces a single prediction per input image, predictions across views may be inconsistent due to per-view ambiguity. A view-consistent solution instead requires a range of plausible predictions per view. Inspired by the ambiguity-aware depth estimation of \cite{uy2023scade}, we adapt the prior model~\cite{Careaga2024Colorful} to be ambiguity-aware by training it with conditional Implicit Maximum Likelihood Estimation (cIMLE), avoiding GAN mode collapse~\cite{goodfellow2014generative}. Concretely, we sample $M$ latent codes per view, injecting each via AdaIN to yield $M$ albedo hypotheses; see Appendix for details.

Here the modes represent the plausible albedo configurations consistent with the input image. Since the training dataset provides only one observed albedo per input, the samples for a given input share the scale factor of Sec.~\ref{sec:handling-scale-variations}, isolating the effect of scale and input ambiguity on the prediction variations. Further details on ambiguity-aware prior training are in the supplement.

\input{figs/main/method_2_SPL/scl_figure}
\subsection{Albedo Fusion via Space Carving Loss}
\label{sec:albedo-space-carving-loss}
To distill the view-wise albedo hypotheses into a single, globally consistent reflectance, we formulate a probabilistic space carving for reflectance, illustrated in Figure~\ref{fig:method_2_SPL_figure}, in the spirit of ambiguity-aware space carving for depth~\cite{uy2023scade}. For each view $v$ and pixel $(i,j)$, the cIMLE prior provides a finite set of albedo hypotheses $\{p_{v,k}^{ij}\}$. We define the fused albedo $\hat{p}^{ij}$ to be the mode that is jointly consistent across views up to the per-view scale indeterminacy of albedo, i.e., we select one hypothesis per view and minimize $\sum_v \lVert \hat{p}^{ij} - \lambda_v\, p_{v,k_v}^{ij} \rVert_2^2$, with $\lambda_v$ learned to absorb scale. This induces the sample-based objective below, which is the reflectance analogue of SCADE's mode selection loss and supervises modes rather than moments:
\begin{equation}
\mathcal{L}_{\text{SC}} = \sum_{(i,j) \in \Omega} \min_{k \in [M]} \left\| \hat{p}^{ij} - p_k^{*ij} \right\|_2^2,
\end{equation}

where $\hat{p}^{ij}$ is the predicted albedo at pixel $(i,j)$, $p_k^{*ij}$  represents the $k$-th scaled albedo hypothesis with $\lambda$ from the ambiguity-aware prior, $\Omega$ denotes all pixel positions, and $M$ is the number of albedo samples per image.

Although the objective performs per-pixel selection among hypotheses, the continuity of the albedo decoder preserves local smoothness in practice without requiring additional regularizers.

The total loss to optimize and train our model is the sum of reconstruction loss and weighted space carving loss presented in Eq.~\ref{eq:our_loss_fn}: 
\begin{align}
     \mathcal{L} = \mathcal{L}_{\text {recon }} +  w_{SC} \mathcal{L}_{\text{SC}}  \label{eq:our_loss_fn}
\end{align}

\paragraph{Implementation Details.} We provide full implementation and training details, including all hyperparameters, in the supplement.

%% file: figs/main/method_1_pipeline/pipeline_fig.tex
\begin{figure*}[t]
  \centering
  \includegraphics[width=\linewidth]{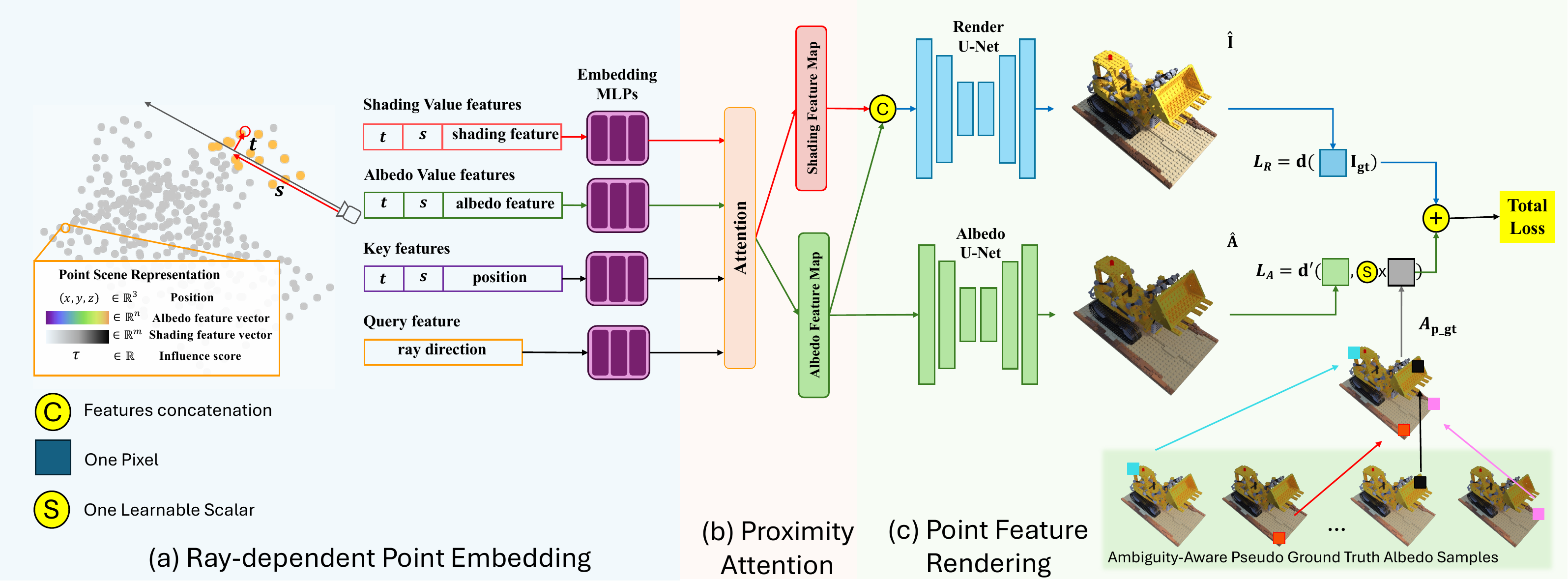}
  \caption{Rendering pipeline overview. Each scene point stores position, albedo, shading features, and an influence score. (a) Ray-dependent embeddings combine values (albedo, shading), keys (position, $t$, $s$), and queries (ray directions) as inputs to the attention model. (b) The attention model selects and aggregates albedo and shading into feature maps. (c) The albedo feature map is rendered (\textcolor{green}{green}) to produce the albedo image; both albedo and shading feature maps are concatenated (\textcolor{blue}{blue}) to generate the final colour image. The model is trained end-to-end with supervision on both albedo and colour outputs.}
  \Description{An illustration of the rendering pipeline showing three main components: (a) ray-dependent embeddings generation, (b) attention-based feature aggregation, and (c) final rendering with separate albedo and colour outputs. The diagram shows how scene points with position, albedo, and shading features are processed through the pipeline to generate both albedo and final colour images.}
  \label{fig:method_1_pipeline_figure}
\end{figure*} 

%% file: figs/main/method_2_SPL/scl_figure.tex
\begin{figure}[t]
  \centering
  \includegraphics[width=\linewidth]{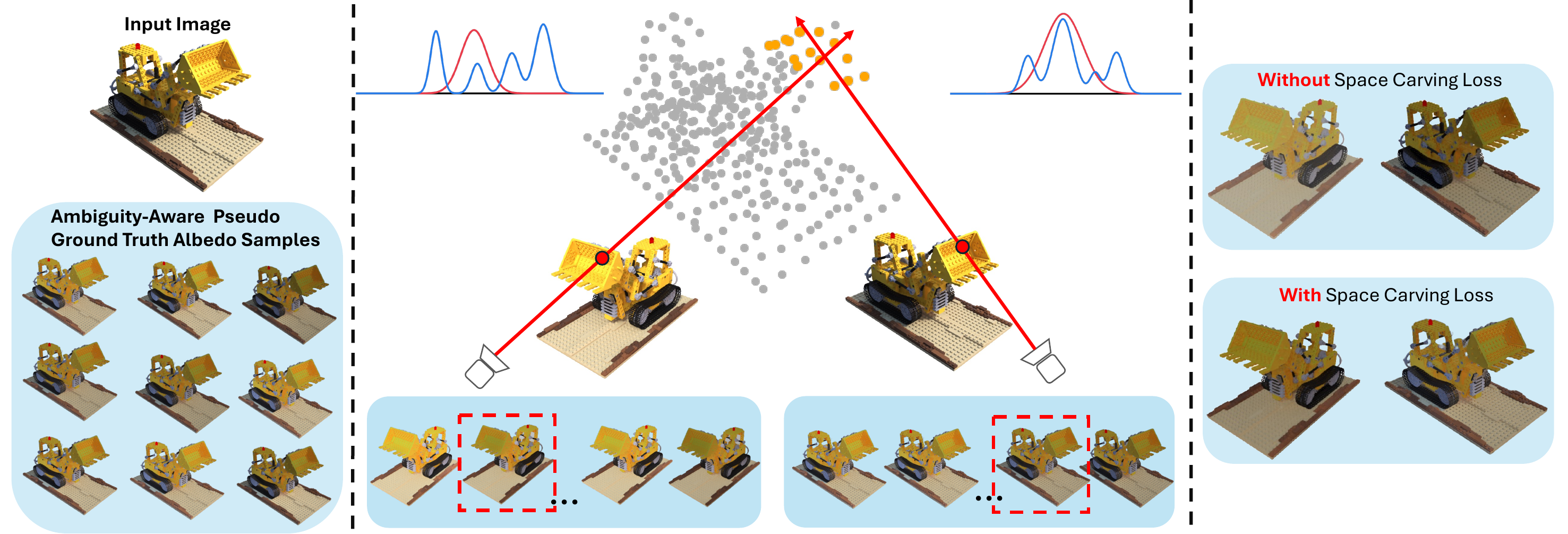}
  \caption{Illustration of Space Carving (SC) Loss. \textbf{Left}: Multiple pseudo-ground-truth albedo samples per view generated by our ambiguity-aware 2D prior model capture estimation ambiguities. \textbf{Middle}: SCL selects consistent albedo modes along rays across views to resolve these ambiguities. \textbf{Right}: Predicted albedo maps exhibit improved cross-view consistency for 3D rendering.}
  \Description{Three-panel illustration showing the Space Carving Loss process. The left panel shows multiple albedo samples from different views, the middle panel demonstrates how consistent modes are selected across views, and the right panel shows the resulting consistent albedo predictions.}
  \label{fig:method_2_SPL_figure}
\end{figure} 

%% file: tex/4_experiments.tex
\section{Experiments}
\label{sec:experiments-intro}

\textbf{Evaluation Setup.} We evaluate our method on NeRF Synthetic~\cite{Mildenhall2020NeRFRS} and TensoIR~\cite{jin2023tensoir} (synthetic) datasets, as well as Tanks \& Temples~\cite{Knapitsch2017} and Mip-NeRF 360~\cite{barron2022mipnerf360} (real-world). To quantify performance across novel view synthesis (NVS) and albedo reconstruction, we report PSNR, SSIM, and LPIPS~\cite{Zhang2018TheUE}. Following prior work (e.g., NeRFactor, GS-IR), we apply per-view scale-and-shift alignment before computing albedo metrics.

\textbf{Baselines.} We compare against state-of-the-art baselines spanning volumetric versus surface representations and physically-based versus intrinsic models: point- and Gaussian-based inverse rendering methods (DPIR~\cite{chung2024differentiable}, GS-IR~\cite{Liang2023GSIR3G}, IRGS~\cite{irgs2025}, DiscretizedSDF~\cite{discretizedsdf2025}), a diffusion-prior PBR approach (MaterialFusion~\cite{litman2025materialfusion}), and a NeRF-based method (Intrinsic-NeRF~\cite{ye2023intrinsicnerf}). We use official implementations and default configurations for all baselines, plus vanilla PAPR~\cite{zhang2023papr} as a pure reconstruction baseline.

\textbf{Results.} As reported in Table~\ref{tab:render_albedo_quantitative_comparison}, Intrinsic PAPR delivers higher-fidelity NVS and more accurate albedo reconstructions than all inverse-rendering baselines across both synthetic and real-world scenes; Figure~\ref{fig:albedo-novel-reconstruction-comparison} shows this qualitatively on NeRF Synthetic. The supplement provides per-scene breakdowns and also shows that simply adding 2D prior supervision to baselines does not improve their albedo quality, highlighting the role of PAPR's representation. For positioning against scene-level intrinsic decomposition, MAIR++~\cite{choi2025mairpp} on our nine synthetic scenes reaches NVS\,/\,albedo PSNR of $24.42\,/\,22.28$~dB, trailing our $35.00\,/\,31.21$~dB, a gap due to domain mismatch: MAIR++ targets indoor SVBRDF estimation and excludes object-centric scenes.

Our framework also surpasses vanilla PAPR~\cite{zhang2023papr} in NVS, by supervising albedo and final RGB separately. Explicit albedo supervision resolves texture-illumination ambiguities, forcing the model to explain image formation via a more accurate shading component; since shading is tightly coupled to geometry, this disentanglement yields cleaner geometry and thus higher-quality NVS. Further qualitative NVS comparisons against vanilla PAPR and depth visualizations are in the supplement.
\input{tables/main/render_albedo_quantitative_comparison}


\input{figs/main/render_albedo_comparison/render_albedo_comparison_fig}
\subsection{Albedo and Shading Feature Quality Evaluation}

To evaluate how our design addresses the misattribution issue, we compare against our point-based inverse rendering baselines, DPIR~\cite{chung2024differentiable} and GS-IR~\cite{Liang2023GSIR3G}, on per-point intrinsic transfer tasks (albedo and shading).

\textbf{Protocol.} For both albedo and shading transfers, the edited target should match the source primitive’s attribute. We render the edited scene for each method, decompose into albedo/shading with~\cite{lee2024dmp}, project source/target 3D points to image pixels, and compute the transfer error:
$\mathrm{TE}=\frac{1}{N}\sum_{i=1}^N \lVert \hat{A}_E(p_t^{(i)}) - A_{\text{GT}}(p_s^{(i)}) \rVert_2^2$, where $A\in\{\text{albedo},\text{shading}\}$. For synthetic scenes, $A_{\text{GT}}$ at $p_s$ is read from Blender. Source points are sampled uniformly at random from the source mask; we report mean $\pm$ std over $N{=}100$ trials. When used, the decoupling error measures edit locality: $\mathrm{DE}=\frac{1}{|R|}\sum_{p\in R}\lVert \hat{A}_{\text{post}}(p)-\hat{A}_{\text{pre}}(p)\rVert_2^2$ in regions $R$ that exclude the edited area. Masks are defined with our UI and shared across methods; per-scene visualizations appear in the supplement.

\subsubsection{Albedo Transfer}

Albedo transfer isolates primitive-level reflectance attribution: the desired outcome is that the target reproduces the source albedo independent of shading. Volumetric baselines, whose features are supervised only after transmittance-weighted aggregation, admit compensatory errors and manifest as colour drift and high variability across transfers (Figure~\ref{fig:albedo-shading-transfer-merged}, \textcolor{red}{left}). In contrast, our surface point renderer assigns high attention to the source primitive along some rays, providing direct per-primitive supervision. The transferred albedo closely matches the source with markedly lower variance (Table~\ref{tab:transfer-comparison}), evidencing correct attribution rather than aggregation compensation. This advantage is structural, not a matter of training data: sweeping training views from 25 to 200 keeps our per-primitive transfer error $5.04$--$5.23\times$ below GS-IR at every view count, and at 25 views we already undercut GS-IR's 200-view result ($0.078$ vs.\ $0.314$). More views cannot help, because near-coincident primitives render identically from every direction; no amount of multi-view supervision can disambiguate features that volumetric aggregation has rendered observationally equivalent (per-view-count breakdown in the supplement).

\input{figs/main/albedo_shading_transfer/albedo_shading_transfer}
\input{tables/transfer_comparison_combined}

\subsubsection{Shading Transfer}

Shading transfer probes whether illumination effects are correctly bound to individual primitives and local geometry. DPIR and GS-IR exhibit inconsistencies and artefacts indicative of mixed attributions along the ray, whereas our method produces geometry-consistent shading that tracks the source primitive (Figure~\ref{fig:albedo-shading-transfer-merged}, \textcolor{blue}{right}). This successful transfer proves our indirectly supervised shading branch captures genuine illumination rather than absorbing residual errors, which would otherwise yield arbitrary artefacts upon transfer. Quantitatively, the error and variance are lowest (Table~\ref{tab:transfer-comparison}), confirming precise per-primitive shading attribution.


\subsection{Freeform Intrinsic Editing}
Precise per-primitive albedo and shading attribution enables practical, freeform editing tools. Users paint arbitrary regions on an input view with our UI (details in the supplement); we map these pixels to the underlying 3D points and modify their albedo or shading features, so edits live in the 3D representation and remain multi-view consistent, geometry-aware, and localized.

Figure~\ref{fig:applications_figure} \appgreen{(a)} illustrates local albedo and shading edits in user-drawn regions, such as writing letters or drawing circular patterns on surfaces. The modified appearance follows the surface across views and blends smoothly with unedited areas, providing intuitive ``paint-like'' control of material and shading.

Building on this, we use the same per-primitive shading features to control illumination strength. In Figure~\ref{fig:applications_figure} \apppurple{(b)}, we edit shading intensity by scaling the shading feature vectors of selected primitives before transferring them to a target region, yielding predictable changes from subtle brightening to strong darkening. Applying a uniform scale to the shading features of all primitives (Figure~\ref{fig:applications_figure} \apporange{(c)}) produces coherent scene-level shading adjustments while leaving geometry and albedo unchanged.

Finally, our representation supports more structured edits. Figure~\ref{fig:applications_figure} \appyellow{(d)} demonstrates geometry-aware shadow addition and removal: users select a 2D region, and we adjust shading magnitudes only for primitives projecting into that region, producing shadows that align with object boundaries and occlusions. On the albedo side, disentanglement from shading allows us to synthesize colours not present in the original scene (Figure~\ref{fig:applications_figure} \appblue{(e)}) by forming convex combinations of existing albedo features; we expose these combinations as a continuous palette in the UI, enabling recolouring with unseen yet realistic hues.

\input{figs/main/applications/applications}

%% file: tables/main/render_albedo_quantitative_comparison.tex
\begin{table}[tb]
  \centering
\caption{Novel-view synthesis and albedo reconstruction. Left: NVS on synthetic datasets (NeRF Synthetic~\cite{Mildenhall2020NeRFRS}, TensoIR~\cite{jin2023tensoir}) and real-world datasets (Tanks \& Temples~\cite{Knapitsch2017}, Mip-NeRF 360~\cite{barron2022mipnerf360}). Right: Albedo on NeRF Synthetic vs. Blender ground truth. Intrinsic PAPR achieves the best performance across all metrics and datasets.}
  \label{tab:render_albedo_quantitative_comparison}
  \resizebox{\linewidth}{!}{%
  \begin{tabular}{@{}lccc|ccc@{}}
    \toprule
    \multicolumn{7}{c}{\textbf{Novel View Synthesis (Image Quality Metrics)}}\\
    \midrule
    \multirow{2}{*}{Method} & \multicolumn{3}{c}{Synthetic} & \multicolumn{3}{c}{Real-world} \\
    \cmidrule(lr){2-4} \cmidrule(lr){5-7}
     & \textit{PSNR} $\uparrow$ & \textit{SSIM} $\uparrow$ & $\textit{LPIPS}_{vgg}$ $\downarrow$ & \textit{PSNR} $\uparrow$ & \textit{SSIM} $\uparrow$ & $\textit{LPIPS}_{vgg}$ $\downarrow$ \\
    \midrule
  DPIR~\cite{chung2024differentiable} & $26.47$ & $0.867$ & $0.133$ & $20.17$ & $0.756$ & $0.290$ \\
  MaterialFusion~\cite{litman2025materialfusion} & $28.59$ & $0.902$ & $0.103$ & $25.18$ & $0.801$ & $0.237$ \\
  Intrinsic-NeRF~\cite{ye2023intrinsicnerf} & $29.51$ & $0.936$ & $0.051$ & $23.79$ & $0.768$ & $0.252$ \\
  GS-IR~\cite{Liang2023GSIR3G} & $30.40$ & $0.945$ & $0.055$ & $26.44$ & $0.851$ & $0.185$ \\
  IRGS~\cite{irgs2025} & $31.35$ & $0.949$ & $0.048$ & $27.26$ & $0.878$ & $0.157$ \\
  DiscretizedSDF~\cite{discretizedsdf2025} & $31.72$ & $0.959$ & $0.038$ & $27.86$ & $0.876$ & $0.147$ \\
  PAPR~\cite{zhang2023papr} & $32.84$ & $0.973$ & $0.035$ & $28.03$ & $0.865$ & $0.159$ \\
  Intrinsic PAPR (Ours) & \textbf{35.00} & \textbf{0.978} & \textbf{0.021} & \textbf{29.99} & \textbf{0.912} & \textbf{0.094} \\
    \bottomrule
  \end{tabular}

  \begin{tabular}{@{}lccc@{}}
    \toprule
    \multicolumn{4}{c}{\textbf{Albedo Reconstruction (Synthetic)}}\\
    \midrule
    Method & \textit{PSNR} $\uparrow$ & \textit{SSIM} $\uparrow$ & $\textit{LPIPS}_{vgg}$ $\downarrow$ \\
    \midrule
    DPIR~\cite{chung2024differentiable}  & $22.89$ & $0.854$ & $0.215$ \\
    Intrinsic-NeRF~\cite{ye2023intrinsicnerf}  & $24.86$ & $0.880$ & $0.108$ \\
    MaterialFusion~\cite{litman2025materialfusion}  & $25.36$ & $0.899$ & $0.097$ \\
    GS-IR~\cite{Liang2023GSIR3G}  & $26.33$ & $0.877$ & $0.127$ \\
    IRGS~\cite{irgs2025}  & $28.44$ & $0.913$ & $0.095$ \\
    DiscretizedSDF~\cite{discretizedsdf2025}  & $28.89$ & $0.921$ & $0.097$ \\
    Intrinsic PAPR (Ours) & $\mathbf{31.21}$ & $\mathbf{0.949}$ & $\mathbf{0.056}$ \\
    \bottomrule
  \end{tabular}
  }
\end{table}

%% file: figs/main/render_albedo_comparison/render_albedo_comparison_fig.tex
\begin{figure*}
\setlength\tabcolsep{1pt}
\footnotesize
\centering
\resizebox{\linewidth}{!}{
\begin{tabular}{c}
\includegraphics[width=\linewidth]{figs/main/render_albedo_comparison/render_albedo_comparison.pdf}
\end{tabular}
}
\caption{
Qualitative albedo comparison on the NeRF Synthetic dataset~\cite{Mildenhall2020NeRFRS}. Per-view albedo PSNR is shown per method; Intrinsic PAPR achieves the best average. Ground truth is from Blender. To visualize the albedo maps, we follow the approaches of NeRFactor~\cite{zhang2021nerfactor} and GS-IR~\cite{Liang2023GSIR3G}, computing a scaling factor for each RGB channel.}
\Description{A comparison figure showing albedo maps from different methods on the NeRF Synthetic dataset, with PSNR metrics displayed for each view. The visualization includes ground truth maps from Blender and demonstrates the effectiveness of different approaches in reconstructing surface albedo.}
\label{fig:albedo-novel-reconstruction-comparison}
\end{figure*} 

%% file: figs/main/albedo_shading_transfer/albedo_shading_transfer.tex
\begin{figure*}[t]
\setlength\tabcolsep{1pt}
\footnotesize
\centering
\begin{minipage}[t]{0.495\linewidth}
\includegraphics[width=\linewidth]{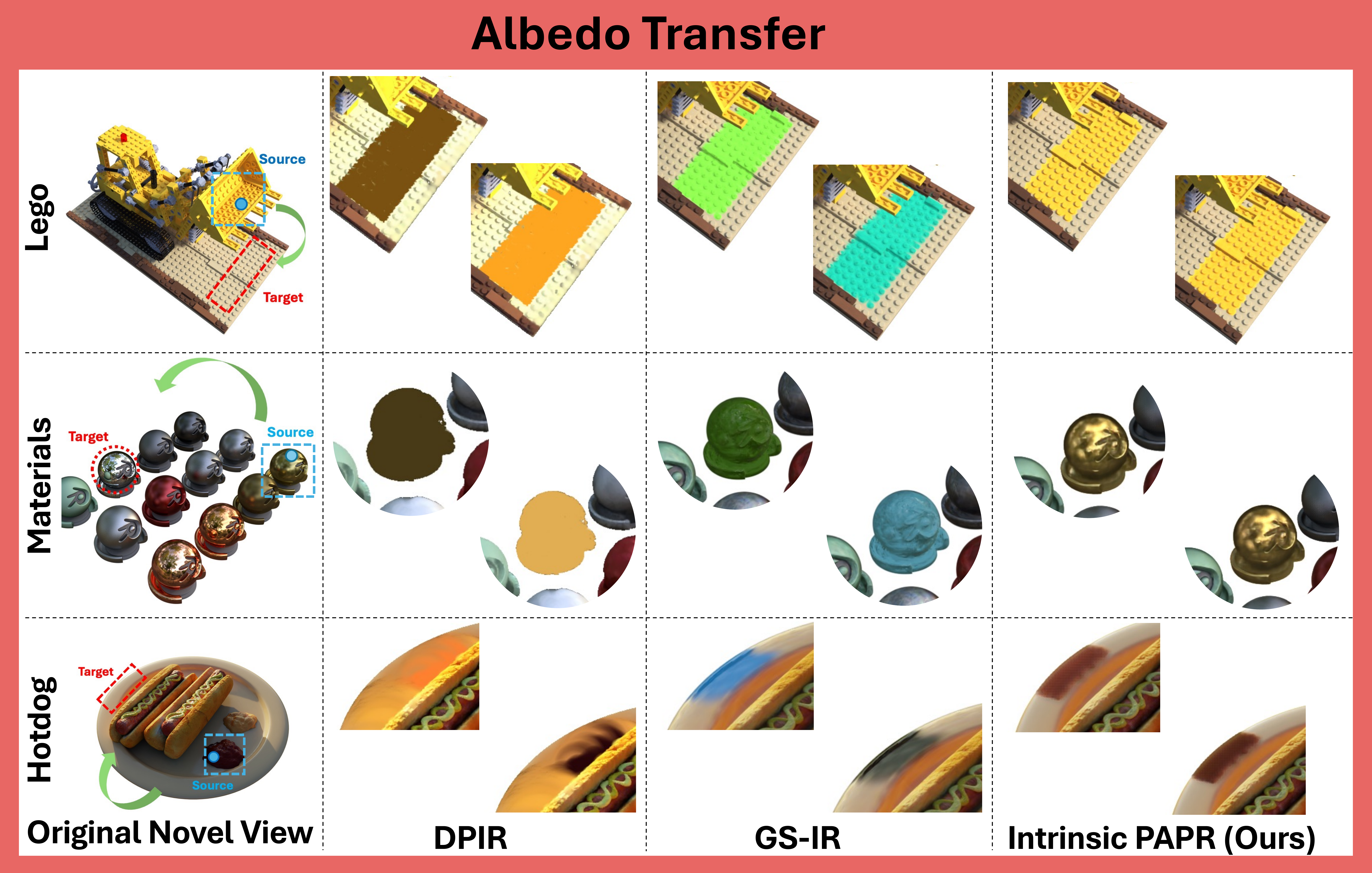}
\end{minipage}\hfill
\begin{minipage}[t]{0.495\linewidth}
\includegraphics[width=\linewidth]{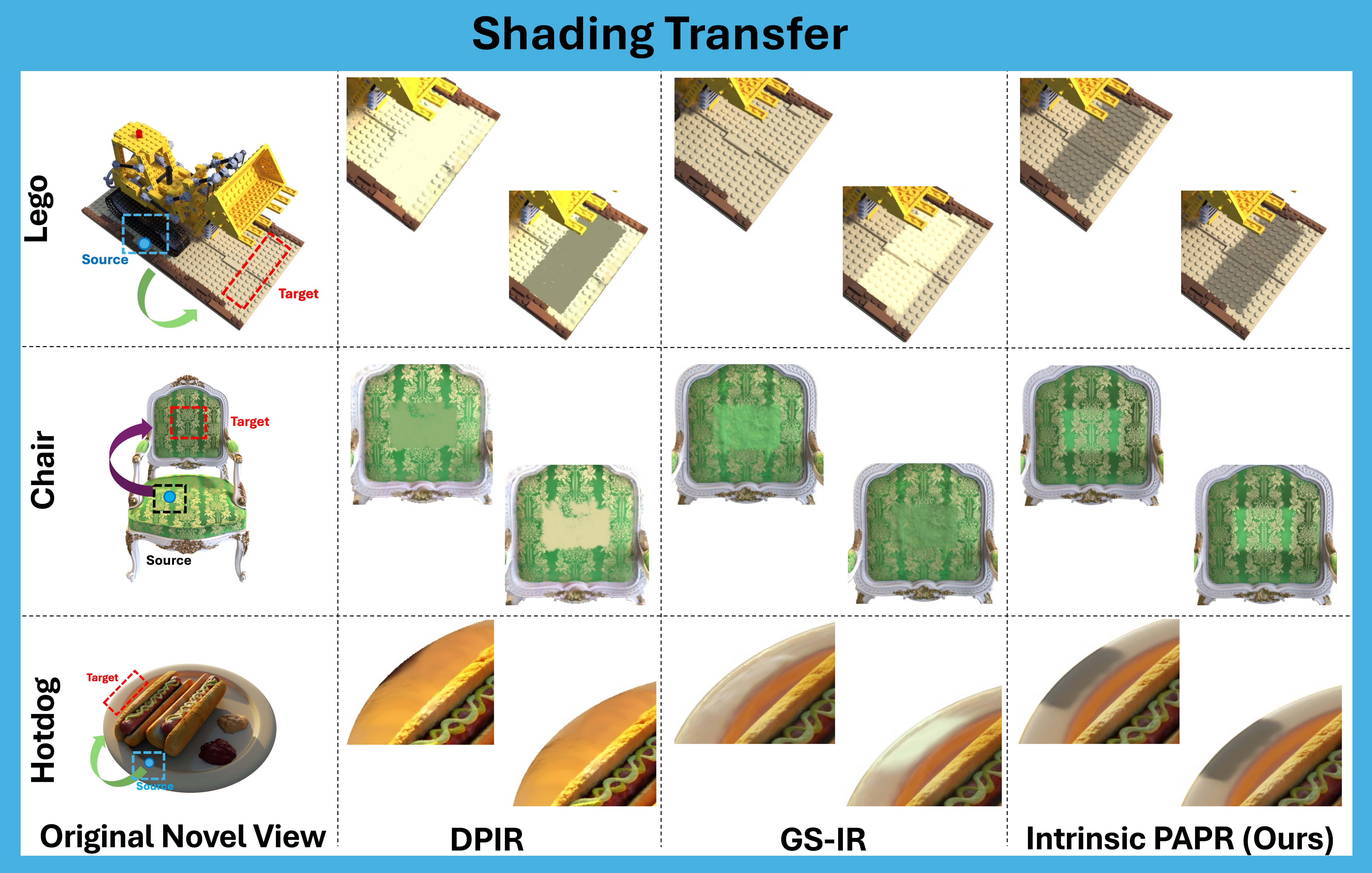}
\end{minipage}
\caption{Per-point intrinsic transfer on NeRF-Synthetic. \textcolor{red}{\textbf{Left:} Albedo transfer}—our method preserves primitive-level reflectance across random source points, while volumetric baselines show colour drift and high variability. \textcolor{blue}{\textbf{Right:} Shading transfer}—our method produces geometry-consistent illumination tied to the source primitive, whereas DPIR and GS-IR show inconsistent shading and artefacts from mixed attributions.}
\Description{Side-by-side qualitative results: left panel shows albedo transfer, right panel shows shading transfer. Our method yields consistent, per-primitive albedo and shading, while baselines display drift, inconsistency, and artefacts.}
\label{fig:albedo-shading-transfer-merged}
\end{figure*}

%% file: tables/transfer_comparison_combined.tex
\begin{table}[tb]
  \caption{Per-point intrinsic attribution via transfer. Mean and standard deviation of MSE over 100 random transfers per scene, for albedo (left) and shading (right). Our method achieves substantially lower error and variance than volumetric baselines, indicating that per-point albedo/shading features are correctly learned.}
  \label{tab:transfer-comparison}
  \centering
  \footnotesize
  \begin{minipage}[t]{0.48\linewidth}
    \centering
    \resizebox{\linewidth}{!}{%
    \begin{tabular}{l l c c c c c c}
      \toprule
      \multicolumn{8}{c}{\textbf{Albedo Transfer (MSE: Avg $\downarrow$ / STD $\downarrow$)}} \\
      \midrule
      & & \multicolumn{2}{c}{Lego} & \multicolumn{2}{c}{Materials} & \multicolumn{2}{c}{Hotdog} \\
      \cmidrule(lr){3-4} \cmidrule(lr){5-6} \cmidrule(lr){7-8}
      & Method & Avg & STD & Avg & STD & Avg & STD \\
      \midrule
      & DPIR~\cite{chung2024differentiable} & 0.236 & 0.179 & 0.221 & 0.106 & 0.291 & 0.135 \\
      & GS-IR~\cite{Liang2023GSIR3G} & 0.381 & 0.219 & 0.318 & 0.175 & 0.280 & 0.141 \\
      & Ours & \textbf{0.053} & \textbf{0.016} & \textbf{0.067} & \textbf{0.026} & \textbf{0.071} & \textbf{0.033} \\
      \bottomrule
    \end{tabular}%
    }
  \end{minipage}\hfill
  \begin{minipage}[t]{0.48\linewidth}
    \centering
    \resizebox{\linewidth}{!}{%
    \begin{tabular}{l l c c c c c c}
      \toprule
      \multicolumn{8}{c}{\textbf{Shading Transfer (MSE: Avg $\downarrow$ / STD $\downarrow$)}} \\
      \midrule
      & & \multicolumn{2}{c}{Lego} & \multicolumn{2}{c}{Chair} & \multicolumn{2}{c}{Hotdog} \\
      \cmidrule(lr){3-4} \cmidrule(lr){5-6} \cmidrule(lr){7-8}
      & Method & Avg & STD & Avg & STD & Avg & STD \\
      \midrule
      & DPIR~\cite{chung2024differentiable} & 0.328 & 0.223 & 0.425 & 0.135 & 0.482 & 0.119 \\
      & GS-IR~\cite{Liang2023GSIR3G} & 0.414 & 0.157 & 0.319 & 0.101 & 0.540 & 0.187 \\
      & Ours & \textbf{0.074} & \textbf{0.029} & \textbf{0.059} & \textbf{0.022} & \textbf{0.041} & \textbf{0.033} \\
      \bottomrule
    \end{tabular}%
    }
  \end{minipage}
\end{table}

%% file: figs/main/applications/applications.tex
\begin{figure*}[t]
\setlength\tabcolsep{2pt}
\footnotesize
\centering
\begin{tabular}{cc}
\includegraphics[width=0.49\linewidth]{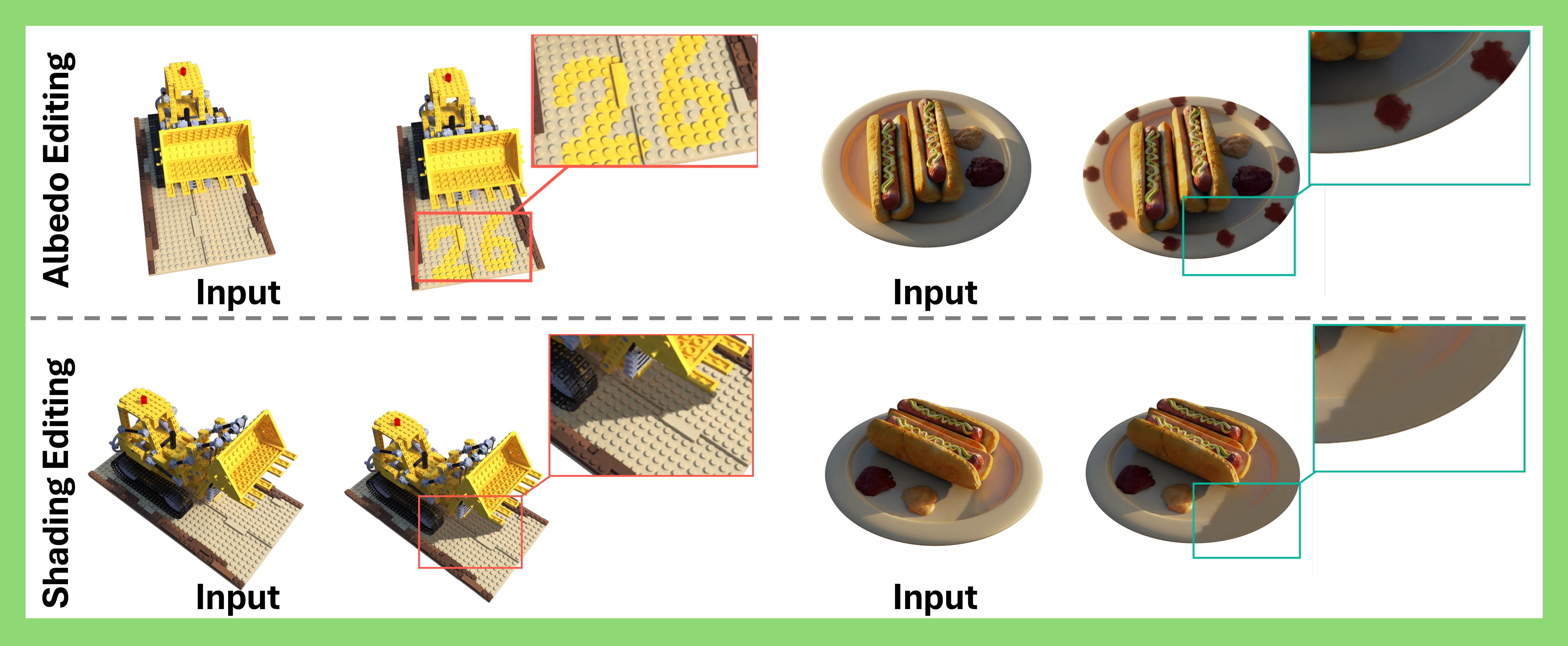}
&\includegraphics[width=0.49\linewidth]{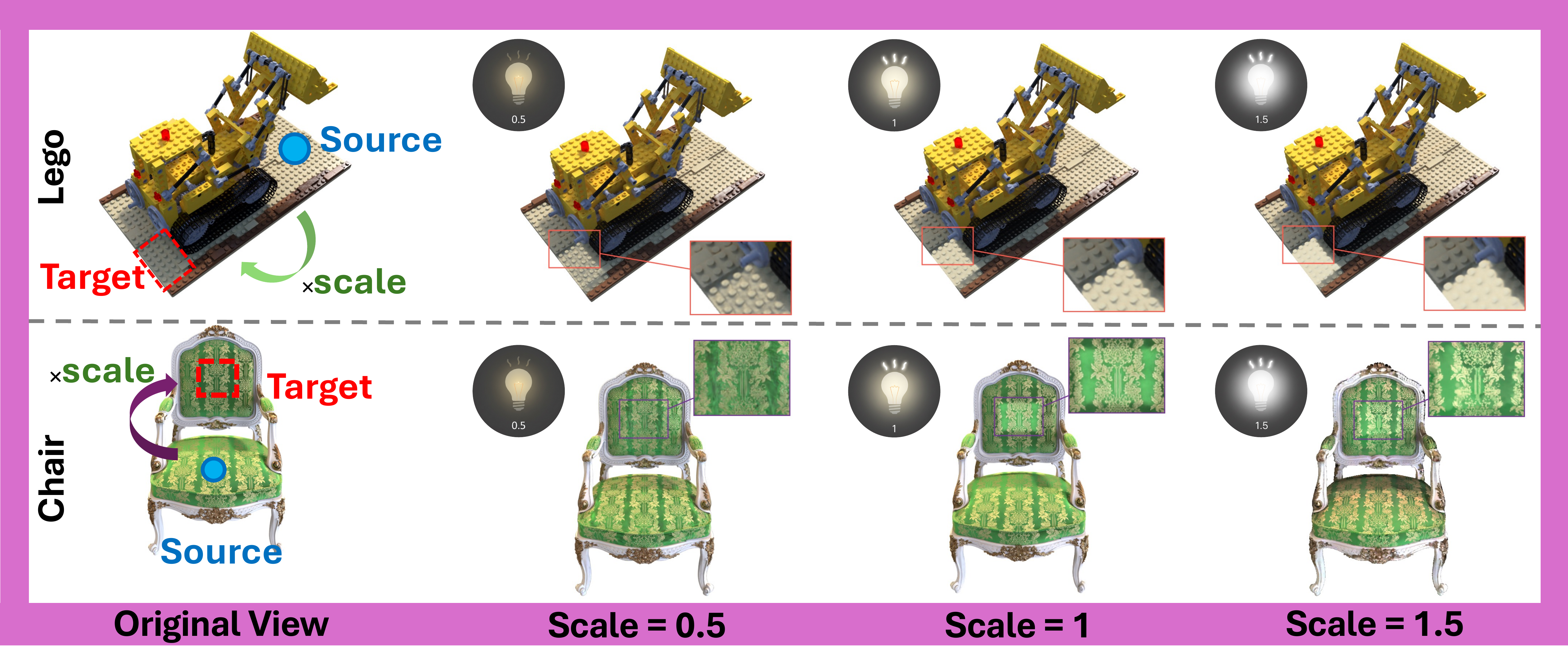} \\
\includegraphics[width=0.49\linewidth]{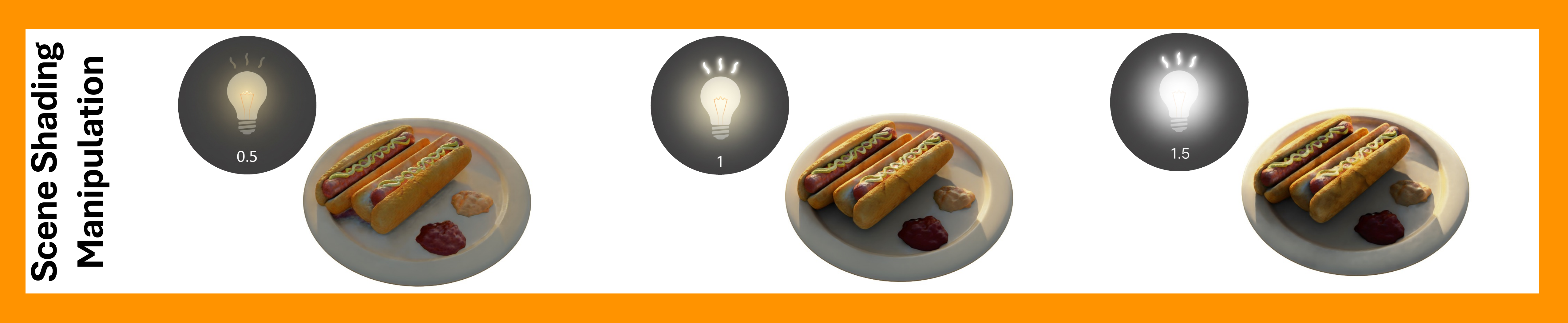}
&\includegraphics[width=0.49\linewidth]{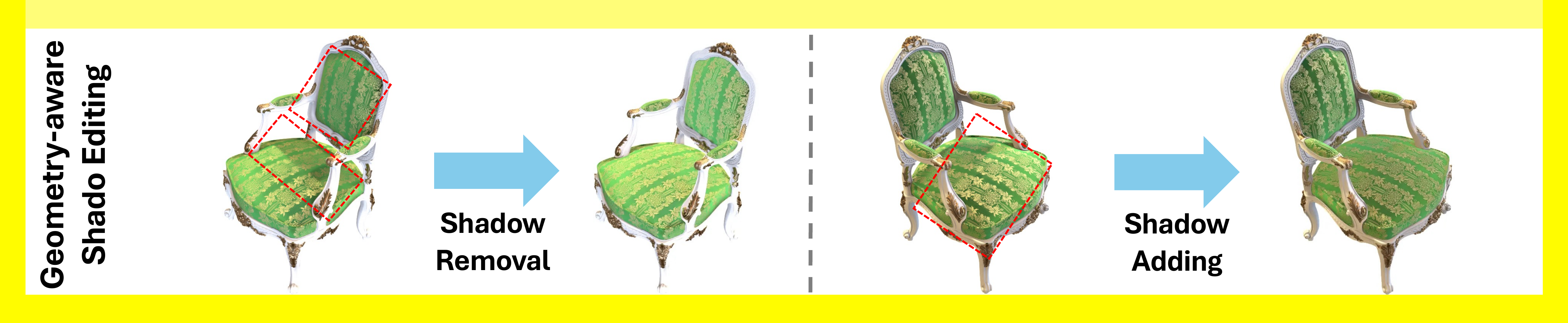} \\
\multicolumn{2}{c}{{\setlength{\fboxrule}{3.5pt}\fcolorbox{blue}{white}{\includegraphics[width=0.96\linewidth]{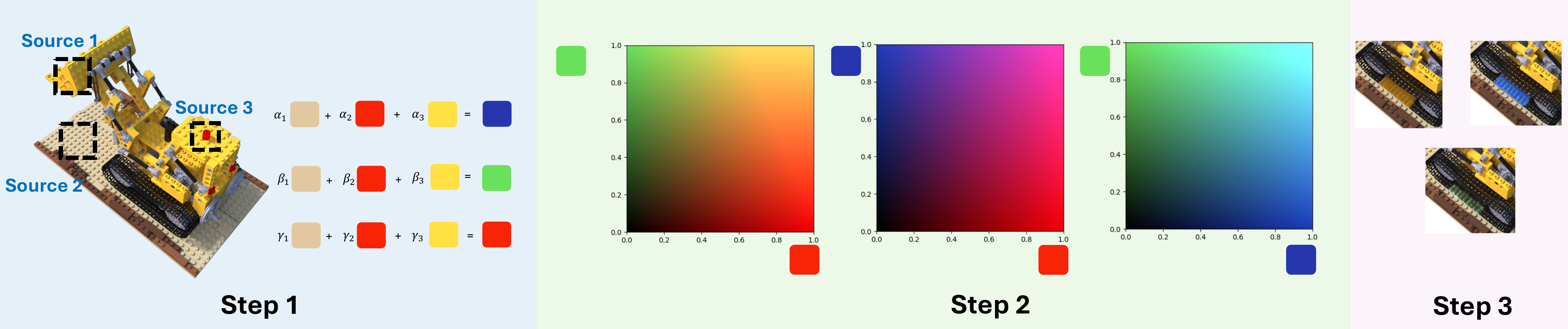}}}}
\end{tabular}
\caption{Freeform intrinsic editing enabled by precise per-primitive attribution. \appgreen{(a) Freeform region edits} on \emph{albedo} and \emph{shading} produce 3D-consistent results that adhere to object boundaries and occlusions. \apppurple{(b) Point-level \emph{shading intensity control}}: scaling shading features modulates contrast locally in a predictable, geometry-consistent manner. \apporange{(c) \emph{Scene-level shading manipulation}} through uniform scaling of shading features adjusts global contrast while respecting geometry. \appyellow{(d) \emph{Shadow addition/removal}:} geometry-aware edits specified in 2D yield coherent shadows that align with scene structure. \appblue{(e) Illustration of unseen colour generation} via albedo feature interpolation: (1) deriving primary colour coefficients, (2) generating colour palettes by convex combinations of different pairs of primary colours, and (3) applying unseen colours.}
\Description{A 2x2 grid: (a) freeform albedo/shading edits in arbitrary regions; (b) point-level shading intensity control via scaling; (c) scene-level shading manipulation via scaling; (d) geometry-aware shadow addition and removal with user-specified regions.}
\label{fig:applications_figure}
\end{figure*}

%% file: tex/5_ablation_study.tex
\section{Ablation Study}
\label{sec:ablation-study}

We ablate Intrinsic PAPR along two axes: the contribution of each component, and the choice of decomposition scope. Throughout, we report novel-view-synthesis (NVS) PSNR, albedo PSNR, and cross-view albedo consistency, the last measured by the \textbf{mean albedo consistency error (MACE)}: we track $K$ surface points over $T$ consecutive frames of a rendered camera trajectory and measure the drift in their rendered albedo,
$\mathrm{MACE}=\frac{1}{K(T-1)}\sum_{i=1}^K \sum_{j=1}^{T-1} \lVert \hat{\mathbf{x}}_{ij} - \hat{\mathbf{x}}_{i,j+1}\rVert_2^2$,
where $\hat{\mathbf{x}}_{ij}$ is the rendered albedo for point $i$ in frame $j$ (lower is better).

\paragraph{Per-component Ablation.} Starting from the vanilla PAPR backbone, we add one component at a time---the albedo/shading split with the 2D prior, the cIMLE-adapted prior, the space-carving loss (Sec.~\ref{sec:intrinsic-papr}), and the learnable per-view scalar---with all other settings held fixed (Table~\ref{tab:ablation}, top). The improvement is monotone across all rungs---NVS $32.84\!\rightarrow\!35.00$~dB, albedo $29.11\!\rightarrow\!31.21$~dB, MACE $0.019\!\rightarrow\!0.015$---and the space-carving loss contributes the largest single albedo gain ($+0.84$~dB) and the largest reduction in cross-view drift, confirming that resolving multi-view inconsistencies in the 2D prior---not merely adding supervision---turns albedo supervision into an effective geometric regularizer. Per-scene results, and a per-scene MACE comparison against the raw 2D prior, are in the supplement.
\input{tables/main/ablation_scl_merged}

\paragraph{Decomposition Scope.} We next test whether a richer material model helps. Adding a specular branch, or replacing our shading with a Cook-Torrance microfacet BRDF~\cite{cook1982reflectance} and spherical-harmonic lighting~\cite{ramamoorthi2001efficient}, lowers NVS, albedo, and consistency alike (Table~\ref{tab:ablation}, bottom). The extra parameters make the already ill-posed decomposition harder to estimate---a bias--variance trade-off---so our simpler factorization is better-conditioned.

%% file: tables/main/ablation_scl_merged.tex
\begin{table}[t]
  \centering
  \input{tables/main/ablation_ladder}
\end{table}

%% file: tables/main/ablation_ladder.tex
\centering
\small
\caption{Ablation study on our synthetic scenes (NVS PSNR, albedo PSNR, and albedo consistency, MACE). \textit{Top:} component ablation---starting from vanilla PAPR, we add one component at a time. As a reference, the off-the-shelf 2D prior run per view (no 3D fusion) gives a MACE of $0.020$; our Full model lowers this to $0.015$, showing that multi-view fusion, not the prior alone, makes the albedo view-consistent. \textit{Bottom:} decomposition scope---adding a specular branch, or replacing our shading with a Cook-Torrance microfacet BRDF and spherical-harmonic lighting, lowers all three metrics, supporting our simpler factorization.}
\label{tab:ablation}
\begin{tabular}{@{}lccc@{}}
  \toprule
  Variant & NVS PSNR$\uparrow$ & Albedo PSNR$\uparrow$ & MACE$\downarrow$ \\
  \midrule
  Vanilla PAPR & 32.84 & -- & -- \\
  + albedo/shading split + 2D prior & 33.71 & 29.11 & 0.019 \\
  + cIMLE-adapted prior ($M{=}10$) & 34.36 & 29.74 & 0.018 \\
  + space-carving loss & 34.83 & 30.58 & 0.016 \\
  + per-view scalar (Full) & \textbf{35.00} & \textbf{31.21} & \textbf{0.015} \\
  \midrule
  \multicolumn{4}{@{}l}{\textit{Decomposition scope (alternatives to Full)}} \\
  + specular branch & 33.20 & 30.95 & 0.017 \\
  + Cook-Torrance + SH lighting & 33.40 & 29.40 & 0.021 \\
  \bottomrule
\end{tabular}

%% file: tex/6_discussion_and_conclusion.tex
\section{Discussion and Conclusion}
\label{sec:discussion}

\paragraph{Limitations.}
\label{sec:limitations}
Our key design choice of reducing the number of free parameters for more accurate recovery also means that some effects, such as full relighting and subsurface scattering, cannot be modelled; and because we assume surface opacity, highly semi-transparent phenomena (e.g., smoke or clear glass) are likewise not handled. Integrating stronger priors and extending the decomposition to more expressive yet stable appearance models are key future directions.

\paragraph{Conclusion.}
This work revisits point-based intrinsic decomposition and inverse rendering through the lens of misattribution in volume rendering: aggregating semi-transparent primitives along rays fundamentally limits reliable supervision of per-primitive appearance, so we repurpose Proximity Attention Point Rendering (PAPR) as a surface-aligned alternative that restores point-wise identifiability. On this basis, Intrinsic PAPR pairs a minimalist albedo--shading factorization with an ambiguity-aware 2D intrinsic prior and a space carving loss that distills multi-view albedo hypotheses into consistent 3D reflectance. Across synthetic and real scenes, it yields more accurate albedo, higher-quality novel view synthesis, and more reliable feature transfer than recent point-based and NeRF-based baselines, a step toward robust, predictable 3D editing tools and toward extending PAPR to richer appearance models and downstream applications.

%% file: tex/8_ack.tex
\section*{Acknowledgements}
This research was enabled in part by NSERC, the Canada Foundation for Innovation, the Canada CIFAR AI Chairs program, the BC DRI Group and the Digital Research Alliance of Canada.

\subsubsection*{Disclosure of Interests.}
The authors have no competing interests to declare that are relevant to the content of this article.

%% file: tex/7_suppl.tex
\section{Implementation Details}
\label{sec-supp-train-details}

\subsection{PAPR Renderer and Intrinsic Adaptation}
Figure~\ref{fig:supp-arch-mlps} and Figure~\ref{fig:supp-arch-UNet} provide the architectural details of our intrinsic variant of the PAPR renderer~\cite{zhang2023papr}. Each 3D point stores two learnable feature vectors, an albedo feature $\mathbf{a}_i$ and a shading feature $\mathbf{h}_i$, together with its position and influence score. For a given camera ray, we follow PAPR and use a proximity-based attention module to select the top-$K$ nearby points, compute a scalar attention logit from the point position, its perpendicular and along-ray distances to the ray, and its influence, and normalize these logits into weights along the ray. We then aggregate $\mathbf{a}_i$ and $\mathbf{h}_i$ separately with the same weights, yielding per-pixel albedo and shading feature maps $A_{\mathcal{C}}$ and $S_{\mathcal{C}}$ as defined in Sec.~3 of the main paper. The decoder U-Net in Figure~\ref{fig:supp-arch-UNet} takes the concatenation $[A_{\mathcal{C}}, S_{\mathcal{C}}]$ as input and predicts the final RGB image; the per-pixel albedo map is decoded separately from the albedo feature map $A_{\mathcal{C}}$.
\input{figs/supp/fig_arch_mlps}

\clearpage
\subsection{Ambiguity-aware Albedo Estimates}

Our method relies on an ambiguity-aware 2D albedo prior to supply per-view hypothesis sets for the space-carving loss. The training and inference of this prior are strictly decoupled from the Intrinsic PAPR optimization pipeline, ensuring that the 2D prior does not introduce any computational or memory overhead during the main 3D training phase. The preparation and integration of the 2D prior follow a three-step offline procedure:

\noindent\textbf{1. Architecture Modification and Fine-tuning.} Figure~\ref{fig:prior_albedo_arch} provides the architectural details of our ambiguity-aware 2D albedo prior, which extends the intrinsic network of Careaga et al.~\cite{Careaga2024Colorful}. We insert AdaIN layers into the encoder blocks of the original prior and predict their scale and shift parameters from a latent code $\mathbf{z}\sim\mathcal{N}(\mathbf{0},\mathbf{I})$ using a lightweight MLP, so that different samples of $\mathbf{z}$ modulate the encoder features into distinct albedo hypotheses. During training, for each input image we draw $M$ latent codes, obtain a set of candidate albedo maps $\{G(I,z_i)\}_{i=1}^M$, and apply a cIMLE-style objective~\cite{Li2020MultimodalIS,peng2022chimle}: we measure the reconstruction error to the pseudo-ground-truth albedo for all candidates and backpropagate only through the best-matching one. This procedure encourages the prior to represent the full range of plausible albedo explanations—capturing ambiguities such as albedo–shading tradeoffs or texture vs.\ geometry—while remaining faithful to the supervision. This fine-tuning is performed once, independently of any specific scene, and requires approximately 1 hour for 100 steps on a single NVIDIA RTX 3090 GPU using the original training dataset of the prior model.

\noindent\textbf{2. Offline Sample Generation.} Using the single fine-tuned model from the previous step, we perform a one-time offline inference for each scene independently of Intrinsic PAPR training. We feed the training views of a given scene into the fine-tuned ambiguity-aware model, generate $M$ candidate samples per view, and save them to disk. This procedure is run once per scene and takes approximately 2 minutes for 100 views on an RTX 3090 GPU. These saved samples constitute the per-view hypothesis set utilized by our space-carving loss in Sec.~3 of the main paper.

\noindent\textbf{3. Integration with Intrinsic PAPR.} During the actual Intrinsic PAPR optimization on a 3D scene, we simply load all pre-computed albedo samples directly into memory. Because we do not perform any 2D network inference or maintain the prior's weights in GPU memory during the 3D phase, our method introduces zero computational time overhead from the 2D prior.
\input{figs/supp/fig_albedo_prior_arch}
\clearpage

\subsection{Data-preprocessing}
For the ground truth colour and albedo images, we preprocess them to ensure that the intensities in linear space are within the range of [0,1], and then we convert them to the logarithmic domain for improved numerical stability during training.

\subsection{Computational Cost for Editing}
Editing in our method is performed by directly modifying the learnable point features ($\mathbf{a}_i$ for albedo, $\mathbf{h}_i$ for shading) in the already-optimized scene representation. Given a user-defined region, the features of source points are transferred to target points in a single pre-render pass, with no gradient computation, no network re-training, and no further optimization of any kind. The edited scene is then rendered through the standard PAPR pipeline, which is unchanged. Consequently, an edited scene renders at the same speed as the original, approximately 20~FPS on a single GPU. This stands in contrast to methods such as Seal-3D~\cite{wang2023seal3d}, which require fine-tuning network weights to apply each edit.

\subsection{Training Hyperparameters}
We set the feature dimensions $m=n=32$ and $d_\text{albedo}=d_\text{shading}=16$, initialize each view's learnable scalar to $0.85$, use $M=10$ albedo samples per view, and optimize with the Adam optimizer~\cite{Kingma2014AdamAM}. For all real-world datasets, we set $w_{SC}=0.125$ and for synthetic datasets, we set the Space Carving loss weight, $w_{SC}=0.085$.

\subsection{Training Complexity}
Intrinsic PAPR is primarily designed to advance the accuracy and reliability of 3D intrinsic decomposition; training and inference efficiency were not primary design objectives of this work. That said, the following reports the practical costs of our implementation.

\noindent\textbf{Training.} Our method follows the same 250K-iteration training schedule as vanilla PAPR~\cite{zhang2023papr}, completing in approximately 3 GPU-hours per scene on a single NVIDIA L40S GPU. The integration of the ambiguity-aware 2D prior introduces no per-iteration overhead: as described in Sec.~\ref{sec-supp-train-details}, the $M$ albedo hypotheses for all training views are pre-computed offline and loaded into memory prior to training. Consequently, the overall training time of Intrinsic PAPR is comparable to that of the original PAPR renderer.

\noindent\textbf{Inference.} At test time, our method uses the same rendering pipeline as vanilla PAPR without any modification. Rendering proceeds at approximately 20~FPS on a single GPU, identical to vanilla PAPR.

\noindent\textbf{Peak memory.} GPU memory peaks at $8$--$15$~GB with the default top-$K=20$ proximity attention and $13$--$24$~GB with a larger top-$K=64$ neighbourhood, depending on scene complexity.

\noindent\textbf{Comparison to GS-IR.} Gaussian-splatting inverse-rendering methods such as GS-IR~\cite{Liang2023GSIR3G} render faster than our attention-based pipeline, but are far less accurate for point-level editing: our per-primitive albedo-transfer error stays $5.04$--$5.23\times$ below GS-IR across $25$ to $200$ training views (Table~\ref{tab:view-count-transfer}). We deliberately trade rendering speed for markedly more accurate, directly editable per-primitive intrinsics.
\clearpage

\section{Additional Results}

\subsection{Attention Concentration Analysis}
To substantiate the claim that Intrinsic PAPR's proximity attention leads to highly concentrated supervision---and thereby improved per-point identifiability and reduced misattribution---we quantitatively analyse the attention concentration during inference. For each foreground ray, we compute the maximum normalized attention weight ($w_k^{ij}$) and the Shannon entropy of the attention distribution over the top-$K$ nearest points in the point cloud.

As shown in Figure~\ref{fig:attention-concentration}, the proximity attention is extremely concentrated. The attention entropy histogram has most of its mass at zero (Figure~\ref{fig:attention-concentration}b), indicating that the rendering of a pixel is almost entirely determined by a single 3D point for most rays. Correspondingly, a significant majority of foreground rays place around $90\%$ or more of their attention mass on a single point---a maximum attention weight near $1.0$ (Figure~\ref{fig:attention-concentration}a).

Furthermore, we verify that each foreground point is supervised by a narrow cone of rays by counting, for each point, the number of rays that strongly supervise it---i.e., rays where that point has the highest attention weight and the maximum weight exceeds 0.8. As illustrated in Figure~\ref{fig:attention-concentration}c, this distribution is highly concentrated: approximately 90\% of the foreground points are strongly supervised by exactly one ray, with the remaining foreground points receiving supervision from only a handful of rays.

This geometric locality is the primary mechanism behind Intrinsic PAPR's ability to mitigate misattribution. Unlike volumetric splatting or standard NeRFs, where transmittance is distributed multiplicatively along a long ray, our formulation ensures that the photometric residual and albedo supervision backpropagate primarily to a single, specific surface point. This concentrated, point-wise identifiability prevents compensatory effects where multiple fuzzy primitives jointly recreate a pixel colour, resulting in the crisp, view-consistent albedo and shading factorizations demonstrated throughout our experiments.
\input{figs/supp/Attention_analysis/fig_attention_concentration}
\clearpage

\subsection{Per-scene Quantitative Evaluation on Synthetic Datasets}
In Table~\ref{tab:novel_view_eval_per_scene_synthetic}, we present a detailed scene-by-scene comparison of novel view reconstruction on synthetic datasets, including NeRF Synthetic~\cite{Mildenhall2020NeRFRS} and TensoIR~\cite{jin2023tensoir}. In Table~\ref{tab:albedo_eval_per_scene}, we provide the corresponding per-scene breakdown for the quantitative evaluation of albedo quality on the NeRF Synthetic dataset. Quantitative albedo evaluation is performed on synthetic datasets because real-world captures do not provide albedo ground truth; this follows standard practice in prior work~\cite{zhang2021nerfactor,Liang2023GSIR3G}. Following prior work~\cite{zhang2021nerfactor,Liang2023GSIR3G}, albedo metrics are computed after per-view scale-and-shift alignment, applied uniformly to all compared methods. For a fair, apples-to-apples comparison, we re-run every baseline under this single unified protocol rather than copying their published numbers; reported values therefore differ from the originals wherever the original setups differ from ours---in test splits, scene re-renderings under different environment maps, and per-channel albedo rescaling. Together, these per-scene tables complement the averaged metrics of the main paper by providing a detailed view of how performance varies across individual scenes.
\input{tables/supp/novel_view_eval_per_scene_synthetic}
\input{tables/supp/albedo_eval_per_scene}

\subsection{Per-scene Quantitative Evaluation on Real-world Dataset}
In Table~\ref{tab:NVS_real_scenes_per_scene} we present a per-scene breakdown of novel view reconstruction quality on our real-world benchmarks, comprising the Tanks \& Temples subset~\cite{Knapitsch2017} and the real-scene Mip-NeRF 360 dataset~\cite{barron2022mipnerf360}.
\input{tables/supp/NVS_real_scenes}

\subsection{Per-component Ablation: Per-scene Results}
The main paper reports the component ablation averaged over the synthetic scenes; Table~\ref{tab:ablation-ladder-per-scene} gives the per-scene breakdown for all three metrics. The ordering is consistent across scenes---the full model attains the best NVS PSNR, albedo PSNR, and MACE on every scene---and the largest NVS gains appear on the most geometrically and texturally complex scenes, where the 2D prior is least reliable and multi-view fusion has the most to correct.
\input{tables/supp/ablation_ladder_per_scene}

Figure~\ref{fig:ablation-per-component-qual} shows the same progression qualitatively. Each component contributes a visible improvement: the 2D prior and the cIMLE objective recover finer surface detail, the space-carving loss suppresses the residual cross-view artefacts, and the full model yields the cleanest reconstructions, in agreement with the per-scene metrics above.
\begin{figure}[t]
  \centering
  \includegraphics[width=\textwidth]{figs/supp/ablation_per_component/supp_ablation_grid_compressed.pdf}
  \caption{\textbf{Qualitative per-component ablation.} Novel-view renderings as the components of Intrinsic PAPR are added cumulatively from left to right, starting from vanilla PAPR~\cite{zhang2023papr} and ending with our full model, for two representative scenes. Insets magnify a fine-detail region in each scene. Reconstruction quality improves at every stage, consistent with the per-scene numbers in Table~\ref{tab:ablation-ladder-per-scene}.}
  \label{fig:ablation-per-component-qual}
\end{figure}

The component ablation above is reported on synthetic scenes, where albedo ground truth is available. Table~\ref{tab:scl-realworld-nvs} additionally reports the effect of the space-carving loss on real-world novel-view synthesis (Tanks \& Temples and Mip-NeRF~360): it improves PSNR, SSIM, and LPIPS on real scenes as well, confirming that the gains are not specific to synthetic data.
\input{tables/supp/scl_realworld_nvs}

\subsection{Depth and Novel View Synthesis Comparison with PAPR}
To substantiate the PSNR gains of Intrinsic PAPR over the vanilla PAPR~\cite{zhang2023papr} reported in the main paper, we compare the two methods on held-out views and their corresponding depth reconstructions (Figure~\ref{fig:supp-depth-pcd-papr}). In the left panels, depth visualizations reveal that PAPR frequently produces blurred, ``foggy'' geometry with texture-baked structures, while Intrinsic PAPR recovers sharper surfaces and a cleaner scene layout. In the right panels, novel view synthesis results show that these geometric improvements translate into crisper reconstructions with fewer shading and texture artefacts. Taken together, these observations are consistent with our interpretation in Sec.~4 that explicit albedo supervision and intrinsic factorization encourage geometrically faithful reconstructions that benefit both depth and appearance.
\input{figs/supp/fig_depth_pcd_comparison_vs_papr}

\subsection{Evaluating the Impact of Pseudo-Ground Truth Albedo Supervision on Baselines}
A natural question is whether the improvements of Intrinsic PAPR stem primarily from the additional pseudo-ground-truth albedo supervision provided by the 2D prior, rather than from PAPR's scene representation itself. To answer this, we conduct a controlled ablation in which every baseline receives the same albedo supervision as our method.

This experiment is straightforward to set up because all compared baselines (DPIR~\cite{chung2024differentiable}, GS-IR~\cite{Liang2023GSIR3G}, IRGS~\cite{irgs2025}, DiscretizedSDF~\cite{discretizedsdf2025}, and Intrinsic-NeRF~\cite{ye2023intrinsicnerf}) are inverse rendering methods that natively predict a diffuse or albedo map as part of their decomposition pipeline. We therefore add an $\ell_2$ loss between each baseline's existing albedo output head and the same per-view pseudo-albedo used by our method, after applying per-view global scale--shift alignment. All other losses, learning rate schedules, and hyper-parameters remain unchanged, ensuring that the only difference from the baselines' original training is the additional albedo supervision signal.

As shown in Table~\ref{tab:albedo-supervision-effect-per-scene}, providing the same pseudo-albedo supervision does not improve reconstruction quality for any of the baselines. This result demonstrates that the additional supervision signal alone is insufficient—performance gains require a representation that can effectively leverage per-view albedo guidance. Specifically, volume-rendering-based methods aggregate features from multiple overlapping primitives along each ray, so per-view pixel-level albedo losses cannot unambiguously supervise individual primitive attributes, and the misattribution issue persists. In contrast, PAPR's proximity-attention rendering concentrates each pixel's contribution on a small set of surface points, enabling the albedo supervision to directly constrain per-primitive features and yield the observed improvements.
\input{tables/supp/albedo_supervision_effect_per_scene}

\subsection{Albedo Consistency Across Different Views}
In Section~5 of the main paper, we introduced the mean albedo consistency error (MACE) to measure how stable the rendered albedo of each surface point remains as the viewpoint changes. Specifically, we track $K$ surface points across $T$ consecutive frames of a rendered camera trajectory, and compute MACE as the average squared albedo difference between consecutive frames (lower is better). A method with truly view-consistent albedo should produce near-zero MACE, whereas view-dependent shading leakage or misattributed features inflate this error.

Table~\ref{tab:abbl-space-carving-albedo-consis-per-scene} presents a per-scene breakdown of this metric under three conditions: (i)~Intrinsic PAPR \emph{without} the space carving loss (No SCL), (ii)~Intrinsic PAPR \emph{with} the space carving loss (With SCL), and (iii)~the vanilla 2D prior model~\cite{Careaga2024Colorful} used as a reference. Across all scenes, enabling the space carving loss consistently yields the lowest MACE, demonstrating that fusing multi-view albedo hypotheses into a single geometrically consistent representation effectively suppresses view-dependent artefacts. Notably, the variant without space carving produces MACE values comparable to those of the raw 2D prior, confirming that the multi-view consistency enforced by the space carving loss — rather than the albedo supervision alone — is the primary driver of improved cross-view albedo stability.
\input{tables/supp/tab_albedo_consistency}
\clearpage

\subsection{Decomposition Results on Real-world Scenes}
While quantitative albedo evaluation on real-world data is infeasible due to the absence of ground-truth reflectance, we provide qualitative decomposition results to demonstrate that Intrinsic PAPR produces plausible intrinsic factorizations on unconstrained scenes. Figure~\ref{fig:real-world-decomposition} shows rendered images alongside their corresponding albedo and shading maps for representative scenes from the Mip-NeRF 360 dataset~\cite{barron2022mipnerf360}. The decomposed albedo maps successfully remove illumination effects such as cast shadows and specular highlights, recovering spatially uniform reflectance on surfaces with consistent material. The corresponding shading maps capture smooth, geometry-coherent illumination variations without residual texture content, indicating that the factorization does not suffer from the shading leakage commonly observed in volume-rendering-based methods.
\input{figs/supp/render_albedo_shading/fig_render_albedo_shading}
\clearpage

\subsection{Albedo Transfer}
\label{sec:supp-additional-albedo-transfer}
We present additional qualitative results for the point-level albedo transfer task. Figure~\ref{fig:qualitative-supp-albedo-transfer} showcases transfers on the NeRF Synthetic dataset~\cite{Mildenhall2020NeRFRS}, Figure~\ref{fig:qualitative-supp-albedo-transfer-TT} demonstrates real-world albedo editing on the Tanks \& Temples~\cite{Knapitsch2017} subset, and Figure~\ref{fig:qualitative-supp-albedo-transfer-mipnerf} illustrates the effectiveness of our albedo transfer on unbounded real-world scenes from the Mip-NeRF 360 dataset~\cite{barron2022mipnerf360}. Additionally, we include a comparison with Seal-3D~\cite{wang2023seal3d}, a recolouring baseline that directly assigns RGB values through a user-defined brush tool, to demonstrate the broader impact of our intrinsic decomposition-based editing approach. For Seal-3D, we provide a representative qualitative result on the NeRF Synthetic dataset.
\input{figs/supp/fig_albedo_transfer}
\input{figs/supp/albedo_editing_MipNeRF/fig_albedo_editing_MipNeRF}
\input{figs/supp/fig_real_world_albedo_transfer}
\clearpage

\subsection{Shading Transfer}
To further showcase our method's capabilities, we include qualitative results for point-level shading transfer. Figure~\ref{fig:qualitative-supp-shading-transfer} demonstrates shading edits on the NeRF Synthetic dataset~\cite{Mildenhall2020NeRFRS}, and Figure~\ref{fig:qualitative-supp-shading-transfer-TT} highlights real-world applicability with shading edits on the Tanks \& Temples~\cite{Knapitsch2017} subset. Furthermore, Figure~\ref{fig:qualitative-supp-shading-transfer-mipnerf} demonstrates our method's ability to edit shading in complex, unbounded real-world scenes from the Mip-NeRF 360 dataset~\cite{barron2022mipnerf360}.
\input{figs/supp/fig_shading_transfer}
\input{figs/supp/fig_real_world_shading_transfer}
\input{figs/supp/shading_editing_MipNeRF/fig_shading_editing_MipNeRF}

\subsection{Transfer Error vs.\ Number of Training Views}
To verify that our per-primitive attribution advantage is structural rather than a consequence of the training-data regime, we sweep the number of training views and re-measure the per-primitive transfer error against GS-IR. As Table~\ref{tab:view-count-transfer} shows, the advantage is essentially invariant to view count---the albedo-transfer ratio stays within $5.04$--$5.23\times$ across $25$ to $200$ views, and at $25$ views we already beat GS-IR at $200$ views---consistent with the main paper: near-coincident primitives render identically from every direction, so additional views cannot disambiguate them.
\input{tables/supp/view_count_transfer}
\clearpage

\subsection{Scene-level Shading Intensity Control}
In Figure \ref{fig:qualitative-supp-shading-control} we present more examples of how our method is able to adjust the overall intensity of shading in a scene.
\input{figs/supp/fig_shading_control}
\clearpage

\subsection{Custom Editing Tool}
All freeform editing results presented in this paper—including the teaser, the editing figures in Section~4, and the supplementary examples—were produced using a custom graphical user interface that we developed specifically for this work (Figure~\ref{fig:albedo-shadow-editing-app}).

The editing workflow proceeds as follows. Given a trained Intrinsic PAPR scene, the user loads a reference view and uses the brush tool to paint over the region they wish to edit (the \emph{target region}). The tool provides fine-grained control over brush colour, size, softness, and shape, as well as an eraser for corrections. The user also selects a \emph{source region} whose albedo or shading attributes should be transferred to the target. Once both regions are defined, the interface exports the corresponding 2D pixel coordinates.

During the subsequent rendering pass, these pixel coordinates are back-projected to identify the underlying 3D surface points in the trained scene representation. The albedo or shading feature vectors of the target points are then replaced with those of the source points—a direct feature transfer that requires no gradient computation, no network re-training, and no further optimization (as detailed in the Computational Cost for Editing section). Because the edit modifies the 3D point features themselves, the changes are automatically multi-view consistent: re-rendering the scene from any novel viewpoint reflects the edit with correct geometry and occlusion handling.
\input{figs/supp/albedo_shading_editing_UI/albedo_shading_editing_tool_fig}
\clearpage

\subsection{Robustness, Limitations, and Future Work}
\label{sec:supp-robustness-limitations}

\paragraph{Out-of-Distribution Robustness.}
A key practical concern is whether our method generalizes beyond the training distribution of the 2D albedo prior~\cite{Careaga2024Colorful}, which was trained predominantly on indoor furniture scenes. Our evaluation deliberately covers a broad range of domains: synthetic objects with diverse materials and geometries (NeRF Synthetic~\cite{Mildenhall2020NeRFRS}, TensoIR~\cite{jin2023tensoir}), large-scale industrial and outdoor scenes (Tanks \& Temples~\cite{Knapitsch2017}), and unbounded real-world environments (Mip-NeRF 360~\cite{barron2022mipnerf360}). Despite significant domain gaps, the method consistently produces high-quality novel view synthesis and plausible albedo decompositions across all benchmarks (Tables~\ref{tab:novel_view_eval_per_scene_synthetic}--\ref{tab:NVS_real_scenes_per_scene}). This robustness is not coincidental: our space carving loss (Sec.~3) explicitly handles prior unreliability by generating multiple albedo hypotheses per view and selecting those that are geometrically consistent across views, effectively filtering out erroneous single-view predictions. Quantitatively, this filtering recovers albedo quality despite the domain gap: relative to the deterministic-prior variant without multi-view filtering, the cIMLE and space-carving filter chain raises albedo PSNR from $29.11$ to $31.21$~dB and lowers cross-view drift (MACE) from $0.019$ to $0.015$, confirming that the prior's unreliable predictions are rejected rather than baked into the 3D albedo.

\paragraph{Robustness on Diverse Materials.}
Our method deliberately adopts a Lambertian reflectance assumption and a surface-opacity rendering paradigm. To validate the robustness of these design choices under materials that violate them, we evaluate the decomposition on challenging cases encompassing strong specularity and complex inter-reflections. Figure~\ref{fig:ood-materials} presents the predicted albedo and shading for two such scenarios.

\noindent\textbf{Glossy and Specular Surfaces (Drums).}
The Drums scene contains metallic cymbals (near-mirror) alongside glossy-diffuse drum bodies. On the drum bodies (Figure~\ref{fig:ood-materials}, Regions 3 and 4), the decomposition is clean: specular highlights are correctly attributed to the shading branch, and the albedo recovers a uniform, shadow-free reflectance. On the metallic cymbals (Region 1), the model assigns a low, approximately neutral albedo and absorbs the view-dependent golden reflections into shading. While this is not physically exact (a PBR model would decompose this into metallic and roughness parameters), it yields a \emph{functionally usable} albedo: editing the cymbal's albedo will not inadvertently alter highlights, because they reside in the shading branch. This case demonstrates that the Lambertian assumption degrades \emph{gracefully} on moderately specular surfaces.

\noindent\textbf{Complex Inter-reflections (Real-world Indoor).}
The bottom row of Figure~\ref{fig:ood-materials} combines metallic cookware, glass containers, reflective tabletops, and fabric upholstery in a single real-world frame, exercising multiple assumptions simultaneously. Opaque diffuse surfaces decompose correctly with clean albedo and smooth shading. Specular objects (metal container, Region 1) exhibit graceful degradation similar to the drum scene, with partial environment-reflection leakage into albedo. Additionally, where coloured light bounces between objects (Regions 1 and 4), the ordinal (monochromatic) shading model occasionally attributes the reflected colour to albedo rather than illumination. Despite these localized artefacts, the overall decomposition remains coherent and does not catastrophically fail, supporting the method's practical applicability even on complex real-world scenes.

\noindent\textbf{Known Limitations on Extreme Materials.}
While our model exhibits robust practical performance, predominantly specular surfaces (e.g., perfect mirrors) and highly transparent objects (e.g., clear glass, as seen in Region 3 of the indoor scene) represent the expected boundaries of our assumptions. For near-mirror surfaces, the space carving loss cannot perfectly isolate the minimal diffuse reflectance, leading to environment reflections partially leaking into the albedo. For transparent objects, PAPR's surface-aligned proximity attention lacks a mechanism for compositing content visible \emph{through} a surface, causing the albedo to absorb a blend of the foreground tint and the background. As discussed in the main text, these are necessary architectural trade-offs accepted in exchange for resolving the misattribution issue on the vast majority of opaque, primarily diffuse real-world surfaces.
\input{figs/supp/Special_cases/special_2_cases_fig}

\paragraph{Ordinal Shading and Non-Monochromatic Illumination.}
The 2D prior model~\cite{Careaga2024Colorful} supports both ordinal (greyscale) and colourful (chromatic) shading decomposition. In the current implementation, we use its \emph{ordinal shading} output, which assumes monochromatic illumination. This is a deliberate design choice that simplifies the albedo--shading factorization and is appropriate for the majority of scenes under neutral or white lighting.

However, under strongly coloured illumination, this assumption causes the light colour to be absorbed into the predicted albedo rather than the shading component. Figure~\ref{fig:limitation} illustrates this effect: when a green point light is added to a synthetic scene, both the raw 2D prior and our method bake the green tint into the albedo. Importantly, this is not a fundamental limitation of the Careaga et al.\ prior---it is a consequence of our current choice to use the ordinal output. Upgrading to the colourful shading branch of the same prior model would directly resolve this artefact, and we plan this as an immediate extension.

\paragraph{Future Directions.}
The most immediate extension is replacing the ordinal shading output with the colourful shading branch of the same 2D prior~\cite{Careaga2024Colorful}, which would enable decomposition under non-monochromatic illumination with minimal architectural changes. Beyond this, our prior-based supervision paradigm naturally accommodates richer material properties: additional per-point feature channels for roughness or metallicity can be introduced alongside PBR-aware priors, following the same supervision strategy used for albedo. More broadly, as 2D intrinsic decomposition models continue to advance, our framework can directly benefit from more accurate and domain-general priors without requiring changes to the 3D representation or training procedure.
\input{figs/supp/fig_limitation}

%% file: figs/supp/fig_arch_mlps.tex
\begin{figure*}
\centering
\resizebox{\linewidth}{!}{%
\begin{minipage}[]{.7\textwidth}
    \includegraphics[width=1.0\textwidth]{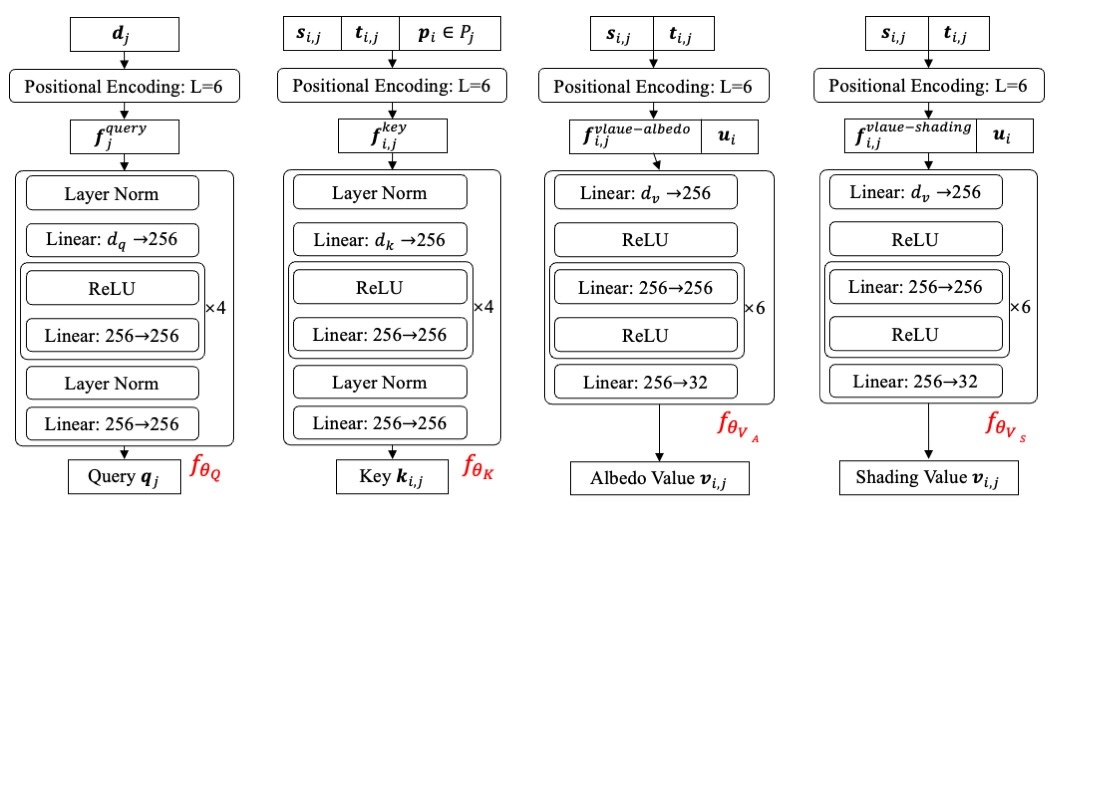}
    \caption{Embedding MLPs architecture details showing the network structure and connections.}
    \Description{A detailed architectural diagram of the Embedding MLPs network, showing the layer connections and structure of the neural network.}
    \label{fig:supp-arch-mlps}
\end{minipage}%
\begin{minipage}[]{.28\textwidth}
\centering
\footnotesize
    \includegraphics[width=1.0\textwidth]{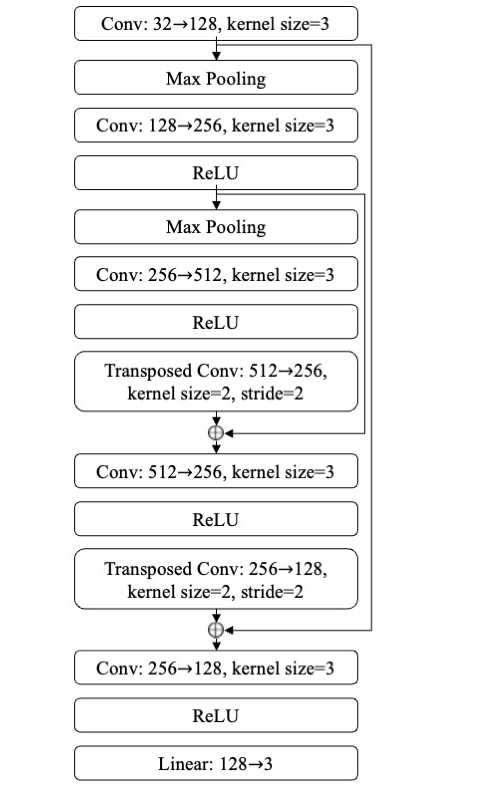}
    \caption{U-Net architecture details showing the encoder-decoder structure.}
    \Description{A detailed view of the U-Net architecture, illustrating the encoder-decoder structure and skip connections.}
    \label{fig:supp-arch-UNet}
\end{minipage}
}
\end{figure*} 

%% file: figs/supp/fig_albedo_prior_arch.tex
\begin{figure*}
  \centering
  \includegraphics[width=\linewidth]{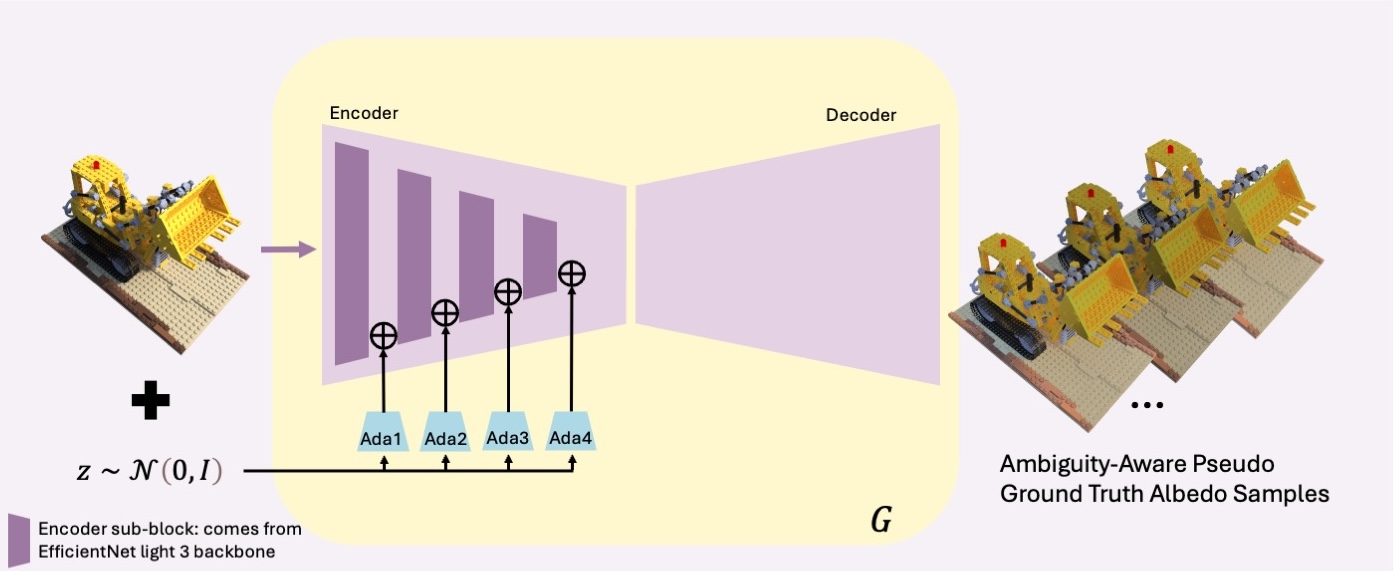}
  \caption{Architecture of our ambiguity-aware albedo estimation module. The network incorporates AdaIN layers that apply scale and shift based on a latent code $z \sim \mathcal{N}(\mathbf{0}, \mathbf{I})$, enabling generation of diverse albedo maps.}
  \Description{Architecture diagram showing the ambiguity-aware albedo estimation network with AdaIN layers for generating diverse albedo maps.}
  \label{fig:prior_albedo_arch}
\end{figure*} 

%% file: figs/supp/Attention_analysis/fig_attention_concentration.tex
\begin{figure}[tb]
    \centering
    \begin{subfigure}{0.32\linewidth}
        \includegraphics[width=\linewidth]{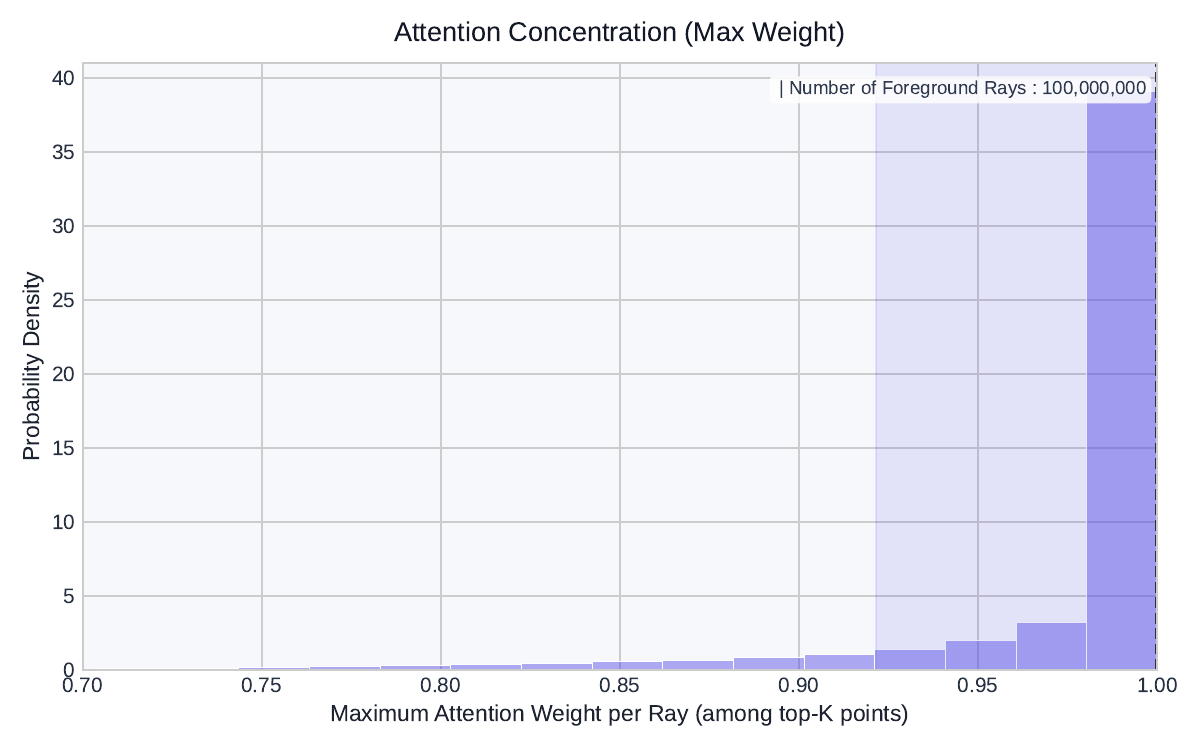}
        \caption{Max Attention Weight}
    \end{subfigure}\hfill
    \begin{subfigure}{0.32\linewidth}
        \includegraphics[width=\linewidth]{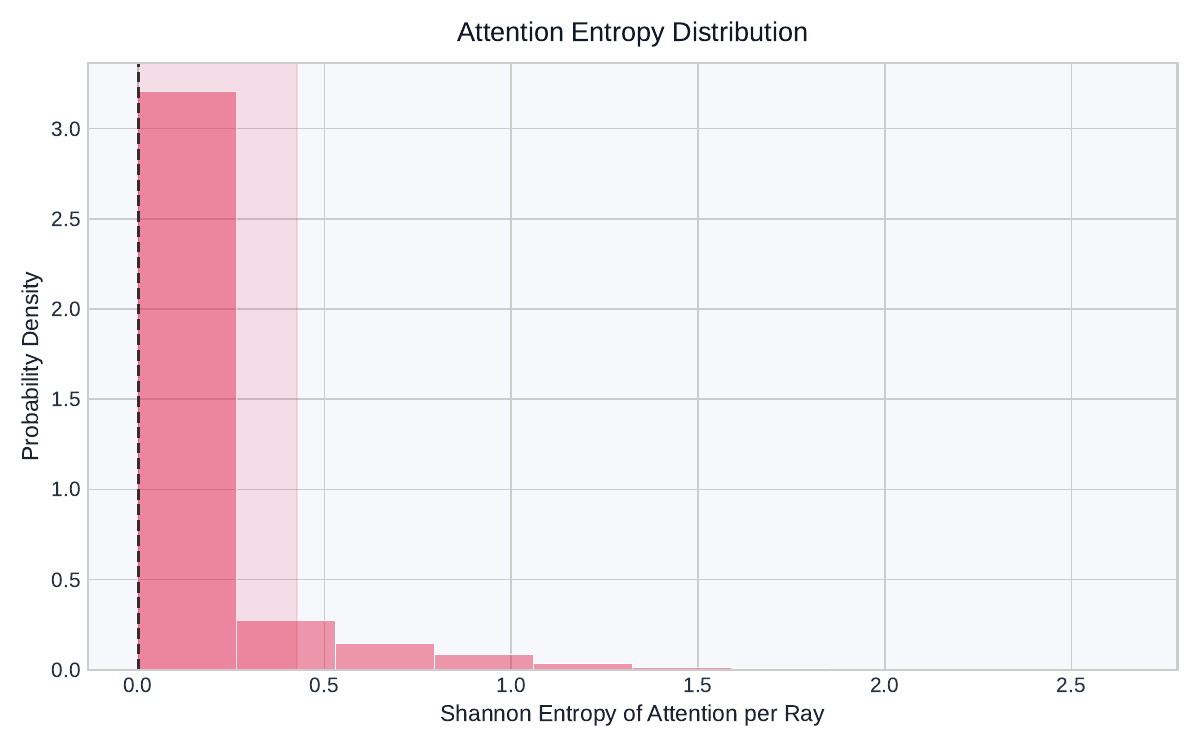}
        \caption{Attention Entropy}
    \end{subfigure}\hfill
    \begin{subfigure}{0.32\linewidth}
        \includegraphics[width=\linewidth]{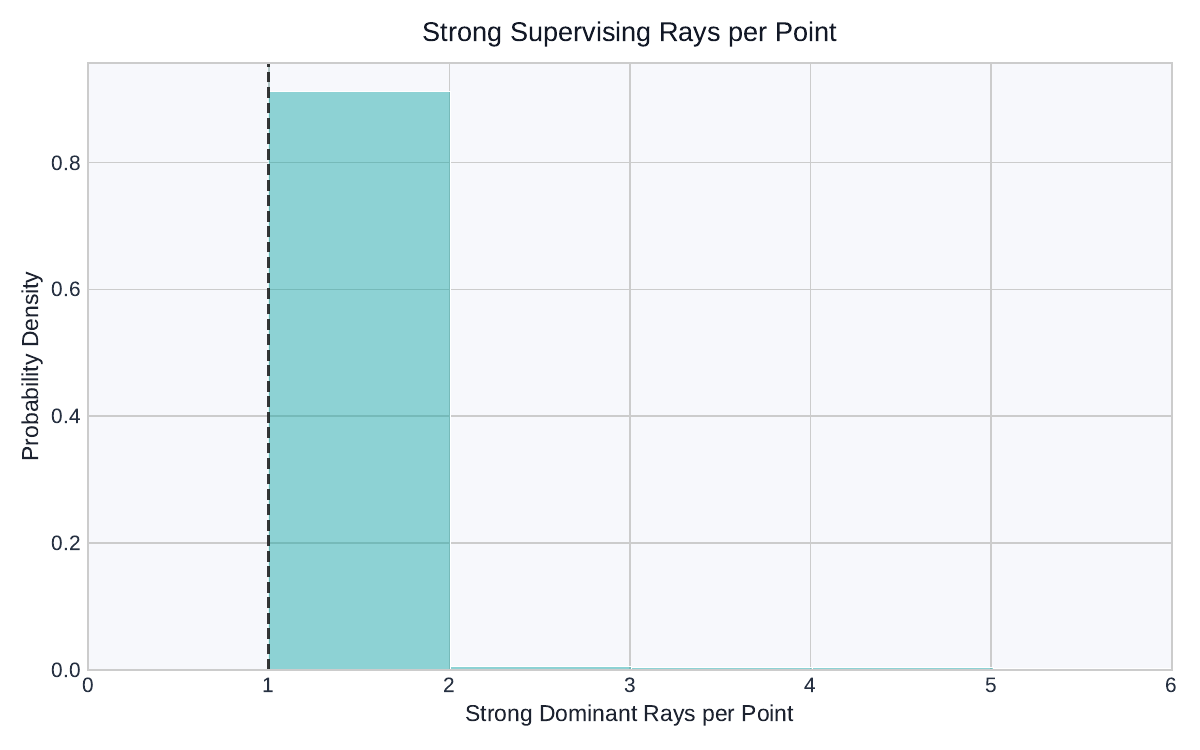}
        \caption{Strong Dominant Rays per Point}
    \end{subfigure}
    \caption{\textbf{Attention Concentration Analysis.} We quantify the proximity attention weights during the rendering process to demonstrate the geometric locality of Intrinsic PAPR. \textbf{(a) Max Attention Weight:} The histogram of maximum attention weights per ray shows a strong peak near 1.0, indicating that a single point strongly dominates the rendered colour of most pixels. \textbf{(b) Attention Entropy:} The histogram of attention entropy confirms that the distribution over the top-$K$ points is highly concentrated, with most mass at zero. \textbf{(c) Strong Dominant Rays per Point:} For each point, we count rays where it is dominant and the max attention weight $\geq 0.8$. Approximately 90\% of foreground points are strongly supervised by exactly one ray, demonstrating that backpropagated supervision is confined to a tight, local cone. Together, these metrics confirm that our proximity-based representation enforces point-wise identifiability, which directly mitigates the misattribution of albedo and shading commonly observed in volumetric scattering methods.}
    \label{fig:attention-concentration}
\end{figure}

%% file: tables/supp/novel_view_eval_per_scene_synthetic.tex
\begin{table*}
  \centering
  \footnotesize
  \caption{Per-scene evaluating novel view synthesis and image quality metrics (PSNR, SSIM, and LPIPS~\cite{Zhang2018TheUE}) on synthetic (NeRF Synthetic~\cite{Mildenhall2020NeRFRS} and TensoIR~\cite{jin2023tensoir}) datasets, broken down by scene. Higher PSNR and SSIM scores are better, while lower LPIPS scores are better. Our method outperforms the baselines across all metrics on both datasets.}
  \label{tab:novel_view_eval_per_scene_synthetic}
  \resizebox{\linewidth}{!}{
    \begin{tabular}{@{}lcccccccccc@{}}
      \toprule
                                                              & Chair          & Drums          & Ficus          & Hotdog         & Lego           & Materials      & Mic            & Ship           & Armadillo      & Avg.           \\
      \midrule
      \multicolumn{10}{c}{\textbf{PSNR}$\uparrow$}                                                                                                                                                                                      \\
      \midrule
      \textit{DPIR~\cite{chung2024differentiable}}            & 26.11          & 23.16          & 24.54          & 25.43          & 27.35          & 26.40          & 27.34          & 25.50          & 32.41          & 26.47          \\
      \textit{MaterialFusion~\cite{litman2025materialfusion}} & 30.98          & 25.26          & 27.12          & 27.12          & 28.62          & 28.60          & 30.46          & 24.77          & 34.40          & 28.59          \\
      \textit{Intrinsic-NeRF~\cite{ye2023intrinsicnerf}}      & 29.37          & 24.48          & 28.90          & 34.58          & 29.15          & 26.60          & 31.13          & 26.21          & 35.16          & 29.51          \\
      \textit{GS-IR~\cite{Liang2023GSIR3G}}                   & 31.10          & 24.50          & 30.77          & 33.82          & 32.79          & 25.79          & 31.69          & 26.74          & 36.40          & 30.40          \\
      \textit{IRGS~\cite{irgs2025}}                           & 33.31          & 24.91          & 33.83          & 34.02          & 32.85          & 26.46          & 32.63          & 27.06          & 37.11          & 31.35          \\
      \textit{DiscretizedSDF~\cite{discretizedsdf2025}}       & 33.42          & 25.02          & 34.73          & 34.36          & 32.91          & 27.02          & 32.89          & 27.24          & 37.92          & 31.72          \\
      \textit{PAPR~\cite{zhang2023papr}}                      & 33.59          & 25.35          & 36.50          & 36.40          & 32.62          & 29.54          & 35.64          & 26.92          & 38.98          & 32.84          \\
      \textit{Intrinsic PAPR (Ours)}                          & \textbf{37.71} & \textbf{27.81} & \textbf{38.87} & \textbf{36.89} & \textbf{34.95} & \textbf{30.96} & \textbf{36.33} & \textbf{28.16} & \textbf{43.35} & \textbf{35.00} \\

      \midrule
      \multicolumn{10}{c}{\textbf{SSIM}$\uparrow$}                                                                                                                                                                                      \\
      \midrule
      \textit{DPIR~\cite{chung2024differentiable}}            & 0.822          & 0.864          & 0.813          & 0.805          & 0.798          & 0.861          & 0.946          & 0.837          & 0.955          & 0.867          \\
      \textit{MaterialFusion~\cite{litman2025materialfusion}} & 0.902          & 0.897          & 0.927          & 0.936          & 0.888          & 0.916          & 0.958          & 0.777          & 0.921          & 0.902          \\
      \textit{Intrinsic-NeRF~\cite{ye2023intrinsicnerf}}      & 0.948          & 0.921          & 0.958          & 0.969          & 0.928          & 0.935          & 0.975          & 0.819          & 0.972          & 0.936          \\
      \textit{GS-IR~\cite{Liang2023GSIR3G}}                   & 0.964          & 0.928          & 0.966          & 0.967          & 0.963          & 0.912          & 0.971          & 0.857          & 0.980          & 0.945          \\
      \textit{IRGS~\cite{irgs2025}}                           & 0.971          & 0.932          & 0.971          & 0.973          & 0.969          & 0.922          & 0.963          & 0.862          & 0.976          & 0.949          \\
      \textit{DiscretizedSDF~\cite{discretizedsdf2025}}       & 0.980          & 0.943          & 0.982          & 0.980          & 0.972          & 0.934          & 0.970          & 0.882          & 0.990          & 0.959          \\
      \textit{PAPR~\cite{zhang2023papr}}                      & 0.986          & 0.951          & 0.994          & 0.988          & 0.981          & 0.972          & 0.993          & 0.904          & 0.989          & 0.973          \\

      \textit{Intrinsic PAPR (Ours)}                          & \textbf{0.990} & \textbf{0.962} & \textbf{0.996} & \textbf{0.991} & \textbf{0.986} & \textbf{0.974} & \textbf{0.995} & \textbf{0.916} & \textbf{0.996} & \textbf{0.978} \\

      \midrule
      \multicolumn{10}{c}{\textbf{LPIPS}$_{Vgg}$$\downarrow$}                                                                                                                                                                           \\
      \midrule
      \textit{DPIR~\cite{chung2024differentiable}}            & 0.122          & 0.145          & 0.279          & 0.136          & 0.140          & 0.121          & 0.132          & 0.113          & 0.077          & 0.133          \\
      \textit{MaterialFusion~\cite{litman2025materialfusion}} & 0.079          & 0.105          & 0.080          & 0.080          & 0.111          & 0.096          & 0.064          & 0.262          & 0.054          & 0.103          \\
      \textit{Intrinsic-NeRF~\cite{ye2023intrinsicnerf}}      & 0.061          & 0.067          & 0.018          & 0.031          & 0.031          & 0.039          & 0.025          & 0.156          & 0.027          & 0.051          \\
      \textit{GS-IR~\cite{Liang2023GSIR3G}}                   & 0.030          & 0.064          & 0.028          & 0.049          & 0.035          & 0.079          & 0.031          & 0.150          & 0.032          & 0.055          \\
      \textit{IRGS~\cite{irgs2025}}                           & 0.027          & 0.053          & 0.039          & 0.032          & 0.027          & 0.049          & 0.029          & 0.148          & 0.029          & 0.048          \\
      \textit{DiscretizedSDF~\cite{discretizedsdf2025}}       & 0.025          & 0.043          & 0.029          & 0.029          & 0.032          & 0.041          & 0.025          & 0.093          & 0.021          & 0.038          \\
      \textit{PAPR~\cite{zhang2023papr}}                      & 0.018          & 0.055          & 0.010          & 0.021          & 0.027          & 0.036          & 0.007          & 0.129          & 0.014          & 0.035          \\
      \textit{Intrinsic PAPR (Ours)}                          & \textbf{0.009} & \textbf{0.033} & \textbf{0.004} & \textbf{0.011} & \textbf{0.013} & \textbf{0.019} & \textbf{0.005} & \textbf{0.082} & \textbf{0.011} & \textbf{0.021} \\

      \bottomrule
    \end{tabular}%
  }
\end{table*}

%% file: tables/supp/albedo_eval_per_scene.tex
\begin{table*}
  \centering
  \footnotesize
  \caption{Per-scene quantitative comparison of albedo quality on the NeRF Synthetic~\cite{Mildenhall2020NeRFRS} dataset. For albedo reconstruction, we follow the methods of NeRFactor~\cite{zhang2021nerfactor} and GS-IR~\cite{Liang2023GSIR3G} to compute the scaling factor for each RGB channel. As shown in the table, our albedo reconstruction quality surpasses all baselines across all scenes based on the specified metrics.}
  \label{tab:albedo_eval_per_scene}
  \resizebox{\linewidth}{!}{
    \begin{tabular}{@{}lcccccccccc@{}}
      \toprule
      & Chair & Drums & Ficus & Hotdog & Lego & Materials & Mic & Ship & Armadillo & Avg. \\
      \midrule
      \multicolumn{10}{c}{\textbf{PSNR}$\uparrow$} \\
      \midrule
      \textit{DPIR~\cite{chung2024differentiable}}   & 19.91 & 20.54 & 27.02 & 21.20 & 20.29 & 22.51 & 21.38 & 20.56 & 32.62 & 22.89 \\

      \textit{Intrinsic-NeRF~\cite{ye2023intrinsicnerf}}   & 18.84 & 21.97 & 26.34 & 25.62 & 23.00 & 18.41 & 30.24 & 20.90 & 38.41 & 24.86 \\

      \textit{MaterialFusion~\cite{litman2025materialfusion}} & 22.77 & 25.69 & 28.72 & 27.60 & 26.58 & 24.26 & 26.73 & 22.60 & 23.28 & 25.36 \\
      
      \textit{GS-IR~\cite{Liang2023GSIR3G}}         & 23.91 & 24.21 & 30.87 & 26.75 & 24.96 & 22.46 & 23.31 & 21.91 & 38.57 & 26.33 \\

      \textit{IRGS~\cite{irgs2025}}         & 24.50 & 25.61 & 31.30 & 27.83 & 27.49 & 26.20 & 30.30 & 24.04 & 38.67 & 28.44 \\

      \textit{DiscretizedSDF~\cite{discretizedsdf2025}}         & 25.38 & 26.12 & 30.96 & 27.92 & 28.13 & 27.37 & 30.19 & 25.14 & 38.79 & 28.89 \\
      
      \textit{Intrinsic PAPR (Ours)}       & \textbf{28.01} & \textbf{29.77} & \textbf{32.08} & \textbf{29.12} & \textbf{30.69} & \textbf{30.74} & \textbf{31.63} & \textbf{29.05} & \textbf{39.80} & \textbf{31.21} \\
      
      \midrule
      \multicolumn{10}{c}{\textbf{SSIM}$\uparrow$} \\
      \midrule
      \textit{DPIR~\cite{chung2024differentiable}}   & 0.851 & 0.796 & 0.943 & 0.863 & 0.849 & 0.771 & 0.908 & 0.706 & 0.996 & 0.854 \\

      \textit{Intrinsic-NeRF~\cite{ye2023intrinsicnerf}}   & 0.855 & 0.833 & 0.940 & 0.962 & 0.905 & 0.774 & 0.959 & 0.705 & 0.987 & 0.880 \\

      \textit{MaterialFusion~\cite{litman2025materialfusion}} & 0.862 & 0.826 & 0.944 & 0.968 & 0.934 & 0.762 & 0.949 & 0.944 & 0.905 & 0.899 \\

      \textit{GS-IR~\cite{Liang2023GSIR3G}}         & 0.816 & 0.797 & 0.948 & 0.941 & 0.889 & 0.850 & 0.815 & 0.850 & 0.986 & 0.877 \\

      \textit{IRGS~\cite{irgs2025}}         & 0.872 & 0.815 & 0.947 & 0.953 & 0.904 & 0.862 & 0.953 & 0.932 & 0.981 & 0.913 \\

      \textit{DiscretizedSDF~\cite{discretizedsdf2025}}         & 0.868 & 0.876 & 0.942 & 0.960 & 0.901 & 0.872 & 0.962 & 0.930 & 0.979 & 0.921 \\
      
      \textit{Intrinsic PAPR (Ours)}       & \textbf{0.898} & \textbf{0.931} & \textbf{0.953} & \textbf{0.972} & \textbf{0.935} & \textbf{0.930} & \textbf{0.969} & \textbf{0.960} & \textbf{0.997} & \textbf{0.949} \\
      
      \midrule
      \multicolumn{10}{c}{\textbf{LPIPS}$_{Vgg}$$\downarrow$} \\
      \midrule
      \textit{DPIR~\cite{chung2024differentiable}}   & 0.155 & 0.182 & 0.066 & 0.178 & 0.849 & 0.198 & 0.096 & 0.198 & 0.009 & 0.215 \\

      \textit{Intrinsic-NeRF~\cite{ye2023intrinsicnerf}}   & 0.137 & 0.148 & 0.033 & 0.097 & 0.129 & 0.185 & 0.051 & 0.169 & 0.022 & 0.108 \\

      \textit{MaterialFusion~\cite{litman2025materialfusion}} & 0.130 & 0.176 & 0.046 & 0.050 & 0.078 & 0.187 & 0.048 & 0.046 & 0.111 & 0.097 \\

      \textit{GS-IR~\cite{Liang2023GSIR3G}}         & 0.136 & 0.164 & 0.053 & 0.088 & 0.143 & 0.208 & 0.095 & 0.205 & 0.051 & 0.127 \\

      \textit{IRGS~\cite{irgs2025}}         & 0.124 & 0.152 & 0.042 & 0.068 & 0.097 & 0.152 & 0.063 & 0.126 & 0.034 & 0.095 \\

      \textit{DiscretizedSDF~\cite{discretizedsdf2025}}         & 0.130 & 0.148 & 0.046 & 0.072 & 0.103 & 0.168 & 0.075 & 0.103 & 0.028 & 0.097 \\
      
      \textit{Intrinsic PAPR (Ours)}       & \textbf{0.082} & \textbf{0.115} & \textbf{0.024} & \textbf{0.030} & \textbf{0.068} & \textbf{0.113} & \textbf{0.034} & \textbf{0.032} & \textbf{0.002} & \textbf{0.056} \\
      
      \bottomrule
    \end{tabular}%
  }
\end{table*}

%% file: tables/supp/NVS_real_scenes.tex
\begin{table*}
    \centering
    \footnotesize
    \caption{Per-scene novel view synthesis and image quality metrics (PSNR, SSIM and LPIPS~\cite{Zhang2018TheUE}) on real-world datasets. Columns list all scenes from the Tanks \& Temples~\cite{Knapitsch2017} subset and the real-scene Mip-NeRF 360 dataset~\cite{barron2022mipnerf360}.}
    \label{tab:NVS_real_scenes_per_scene}
    \resizebox{\linewidth}{!}{%
        \begin{tabular}{@{}lcccccccccccccc@{}}
            \toprule
                                                                    & \multicolumn{5}{c}{Tanks \& Temples} & \multicolumn{7}{c}{Mip-NeRF 360} &                                                                                                                                                                                          \\
            \cmidrule(lr){2-6} \cmidrule(lr){7-13}
                                                                    & Family                               & Truck                           & Caterpillar    & Ignatius       & Barn           & Kitchen        & Room           & Garden         & Bicycle        & Bonsai         & Stump          & Counter        & Average        \\
            \midrule
            \multicolumn{14}{c}{\textbf{PSNR}$\uparrow$}                                                                                                                                                                                                                                                                                \\
            \midrule
            \textit{DPIR~\cite{chung2024differentiable}}            & 19.86                                & 18.70                           & 19.17          & 18.18          & 19.87          & 22.88          & 19.91          & 20.27          & 18.94          & 22.83          & 19.85          & 21.62          & 20.17          \\
            \textit{Intrinsic-NeRF~\cite{ye2023intrinsicnerf}}      & 27.49                                & 23.47                           & 18.80          & 26.37          & 23.96          & 25.69          & 25.71          & 23.42          & 19.73          & 25.61          & 21.13          & 24.11          & 23.79          \\
            \textit{MaterialFusion~\cite{litman2025materialfusion}} & 28.60                                & 24.33                           & 23.06          & 27.76          & 23.60          & 27.10          & 27.14          & 23.96          & 21.86          & 26.56          & 22.40          & 25.75          & 25.18          \\
            \textit{GS-IR~\cite{Liang2023GSIR3G}}                   & 30.16                                & 25.84                           & 23.47          & 26.24          & 25.52          & 27.99          & 28.79          & 25.72          & 23.80          & 28.18          & 25.37          & 26.22          & 26.44          \\
            \textit{IRGS~\cite{irgs2025}}                           & 33.24                                & 26.21                           & 25.48          & 26.91          & 25.93          & 28.56          & 29.10          & 26.54          & 23.75          & 28.72          & 25.78          & 26.88          & 27.26          \\
            \textit{DiscretizedSDF~\cite{discretizedsdf2025}}       & 33.98                                & 27.01                           & 26.28          & 27.34          & 26.12          & 28.72          & 29.56          & 27.17          & 24.10          & 29.54          & 26.76          & 27.73          & 27.86          \\
            \textit{PAPR~\cite{zhang2023papr}}                      & 34.39                                & 26.98                           & 26.79          & 28.40          & 27.06          & 28.31          & 29.36          & 28.40          & 24.21          & 29.12          & 26.23          & 27.13          & 28.03          \\
            \textit{Intrinsic PAPR (Ours)}                          & \textbf{36.38}                       & \textbf{28.46}                  & \textbf{27.50} & \textbf{32.18} & \textbf{29.88} & \textbf{29.91} & \textbf{31.19} & \textbf{29.68} & \textbf{25.68} & \textbf{31.35} & \textbf{27.76} & \textbf{29.86} & \textbf{29.99} \\
            \midrule
            \multicolumn{14}{c}{\textbf{SSIM}$\uparrow$}                                                                                                                                                                                                                                                                                \\
            \midrule
            \textit{DPIR~\cite{chung2024differentiable}}            & 0.874                                & 0.844                           & 0.828          & 0.949          & 0.791          & 0.631          & 0.728          & 0.693          & 0.662          & 0.736          & 0.646          & 0.691          & 0.756          \\
            \textit{Intrinsic-NeRF~\cite{ye2023intrinsicnerf}}      & 0.920                                & 0.837                           & 0.770          & 0.917          & 0.760          & 0.717          & 0.781          & 0.730          & 0.648          & 0.722          & 0.632          & 0.778          & 0.768          \\
            \textit{MaterialFusion~\cite{litman2025materialfusion}} & 0.974                                & 0.884                           & 0.885          & 0.880          & 0.782          & 0.751          & 0.792          & 0.712          & 0.703          & 0.756          & 0.697          & 0.799          & 0.801          \\
            \textit{GS-IR~\cite{Liang2023GSIR3G}}                   & 0.961                                & 0.909                           & 0.878          & 0.920          & 0.857          & 0.867          & 0.867          & 0.804          & 0.706          & 0.883          & 0.716          & 0.839          & 0.851          \\
            \textit{IRGS~\cite{irgs2025}}                           & 0.973                                & 0.916                           & 0.912          & 0.934          & 0.893          & 0.889          & 0.831          & 0.893          & 0.756          & 0.891          & 0.806          & 0.847          & 0.878          \\
            \textit{DiscretizedSDF~\cite{discretizedsdf2025}}       & 0.979                                & 0.931                           & 0.919          & 0.942          & 0.888          & 0.834          & 0.812          & 0.867          & 0.779          & 0.894          & 0.811          & 0.852          & 0.876          \\
            \textit{PAPR~\cite{zhang2023papr}}                      & 0.983                                & 0.931                           & 0.932          & 0.956          & 0.896          & 0.801          & 0.822          & 0.860          & 0.736          & 0.807          & 0.818          & 0.838          & 0.865          \\
            \textit{Intrinsic PAPR (Ours)}                          & \textbf{0.989}                       & \textbf{0.943}                  & \textbf{0.946} & \textbf{0.973} & \textbf{0.922} & \textbf{0.914} & \textbf{0.919} & \textbf{0.905} & \textbf{0.811} & \textbf{0.915} & \textbf{0.831} & \textbf{0.870} & \textbf{0.912} \\
            \midrule
            \multicolumn{14}{c}{\textbf{LPIPS}$_{Vgg}$$\downarrow$}                                                                                                                                                                                                                                                                     \\
            \midrule
            \textit{DPIR~\cite{chung2024differentiable}}            & 0.154                                & 0.310                           & 0.251          & 0.101          & 0.124          & 0.348          & 0.404          & 0.298          & 0.388          & 0.311          & 0.391          & 0.395          & 0.290          \\
            \textit{Intrinsic-NeRF~\cite{ye2023intrinsicnerf}}      & 0.074                                & 0.177                           & 0.218          & 0.081          & 0.277          & 0.306          & 0.356          & 0.262          & 0.342          & 0.250          & 0.344          & 0.335          & 0.252          \\
            \textit{MaterialFusion~\cite{litman2025materialfusion}} & 0.048                                & 0.147                           & 0.137          & 0.092          & 0.299          & 0.268          & 0.311          & 0.229          & 0.299          & 0.393          & 0.301          & 0.321          & 0.237          \\
            \textit{GS-IR~\cite{Liang2023GSIR3G}}                   & 0.043                                & 0.142                           & 0.115          & 0.079          & 0.179          & 0.188          & 0.279          & 0.158          & 0.259          & 0.264          & 0.258          & 0.260          & 0.185          \\
            \textit{IRGS~\cite{irgs2025}}                           & 0.033                                & 0.112                           & 0.103          & 0.071          & 0.160          & 0.178          & 0.206          & 0.152          & 0.198          & 0.241          & 0.199          & 0.232          & 0.157          \\
            \textit{DiscretizedSDF~\cite{discretizedsdf2025}}       & 0.030                                & 0.097                           & 0.093          & 0.067          & 0.152          & 0.163          & 0.189          & 0.139          & 0.181          & 0.239          & 0.183          & 0.231          & 0.147          \\
            \textit{PAPR~\cite{zhang2023papr}}                      & 0.072                                & 0.108                           & 0.157          & 0.118          & 0.031          & 0.180          & 0.209          & 0.154          & 0.201          & 0.244          & 0.202          & 0.226          & 0.159          \\
            \textit{Intrinsic PAPR (Ours)}                          & \textbf{0.018}                       & \textbf{0.092}                  & \textbf{0.081} & \textbf{0.045} & \textbf{0.024} & \textbf{0.096} & \textbf{0.112} & \textbf{0.082} & \textbf{0.107} & \textbf{0.171} & \textbf{0.108} & \textbf{0.187} & \textbf{0.094} \\
            \bottomrule
        \end{tabular}
    }
\end{table*}

%% file: tables/supp/ablation_ladder_per_scene.tex
\begin{table*}
  \centering
  \footnotesize
  \caption{Per-component ablation, per scene. Per-scene novel-view-synthesis PSNR, albedo PSNR, and albedo consistency (MACE) for each rung of the component ablation in the main paper, on the synthetic scenes in our paper. The full model attains the best value on every scene and metric.}
  \label{tab:ablation-ladder-per-scene}
  \resizebox{\linewidth}{!}{
    \begin{tabular}{@{}lcccccccccc@{}}
      \toprule
      & Chair & Drums & Ficus & Hotdog & Lego & Materials & Mic & Ship & Armadillo & Avg. \\
      \midrule
      \multicolumn{11}{c}{\textbf{NVS PSNR}$\uparrow$} \\
      \midrule
      Vanilla PAPR & 33.59 & 25.35 & 36.50 & 36.40 & 32.62 & 29.54 & 35.64 & 26.92 & 38.98 & 32.84 \\
      + albedo/shading + 2D prior & 35.24 & 26.33 & 37.45 & 36.60 & 33.55 & 30.11 & 35.92 & 27.42 & 40.73 & 33.71 \\
      + cIMLE-adapted prior ($M{=}10$) & 36.48 & 27.07 & 38.16 & 36.75 & 34.25 & 30.54 & 36.13 & 27.79 & 42.04 & 34.36 \\
      + space-carving loss & 37.45 & 27.65 & 38.65 & 36.86 & 34.72 & 30.80 & 36.27 & 28.00 & 43.05 & 34.83 \\
      + per-view scalar (Full) & \textbf{37.71} & \textbf{27.81} & \textbf{38.87} & \textbf{36.89} & \textbf{34.95} & \textbf{30.96} & \textbf{36.33} & \textbf{28.16} & \textbf{43.35} & \textbf{35.00} \\
      \midrule
      \multicolumn{11}{c}{\textbf{Albedo PSNR}$\uparrow$} \\
      \midrule
      + albedo/shading + 2D prior & 26.10 & 27.60 & 30.00 & 27.00 & 28.60 & 28.60 & 29.50 & 26.70 & 37.85 & 29.11 \\
      + cIMLE-adapted prior ($M{=}10$) & 26.67 & 28.25 & 30.62 & 27.64 & 29.23 & 29.24 & 30.14 & 27.41 & 38.44 & 29.74 \\
      + space-carving loss & 27.40 & 29.20 & 31.45 & 28.55 & 30.10 & 29.95 & 31.10 & 28.20 & 39.30 & 30.58 \\
      + per-view scalar (Full) & \textbf{28.01} & \textbf{29.77} & \textbf{32.08} & \textbf{29.12} & \textbf{30.69} & \textbf{30.74} & \textbf{31.63} & \textbf{29.05} & \textbf{39.80} & \textbf{31.21} \\
      \midrule
      \multicolumn{11}{c}{\textbf{MACE}$\downarrow$} \\
      \midrule
      + albedo/shading + 2D prior & 0.011 & 0.028 & 0.029 & 0.013 & 0.033 & 0.015 & 0.017 & 0.026 & 0.004 & 0.019 \\
      + cIMLE-adapted prior ($M{=}10$) & 0.010 & 0.026 & 0.026 & 0.012 & 0.030 & 0.014 & 0.016 & 0.024 & 0.004 & 0.018 \\
      + space-carving loss & 0.009 & 0.024 & 0.021 & 0.010 & 0.026 & 0.011 & 0.015 & 0.022 & 0.003 & 0.016 \\
      + per-view scalar (Full) & \textbf{0.009} & \textbf{0.023} & \textbf{0.020} & \textbf{0.009} & \textbf{0.025} & \textbf{0.010} & \textbf{0.014} & \textbf{0.021} & \textbf{0.003} & \textbf{0.015} \\
      \bottomrule
    \end{tabular}
  }
\end{table*}

%% file: tables/supp/scl_realworld_nvs.tex
\begin{table}[t]
\centering
\small
\caption{Effect of the space-carving loss on real-world novel-view synthesis (Tanks \& Temples and Mip-NeRF~360). The space-carving loss improves all three metrics on real scenes as well.}
\label{tab:scl-realworld-nvs}
\begin{tabular}{@{}lccc@{}}
  \toprule
  & PSNR$\uparrow$ & SSIM$\uparrow$ & LPIPS$\downarrow$ \\
  \midrule
  No SCL & 29.14 & 0.891 & 0.126 \\
  With SCL (Full) & \textbf{29.99} & \textbf{0.912} & \textbf{0.094} \\
  \bottomrule
\end{tabular}
\end{table}

%% file: figs/supp/fig_depth_pcd_comparison_vs_papr.tex
\begin{figure*}
  \centering
  \includegraphics[width=\linewidth]{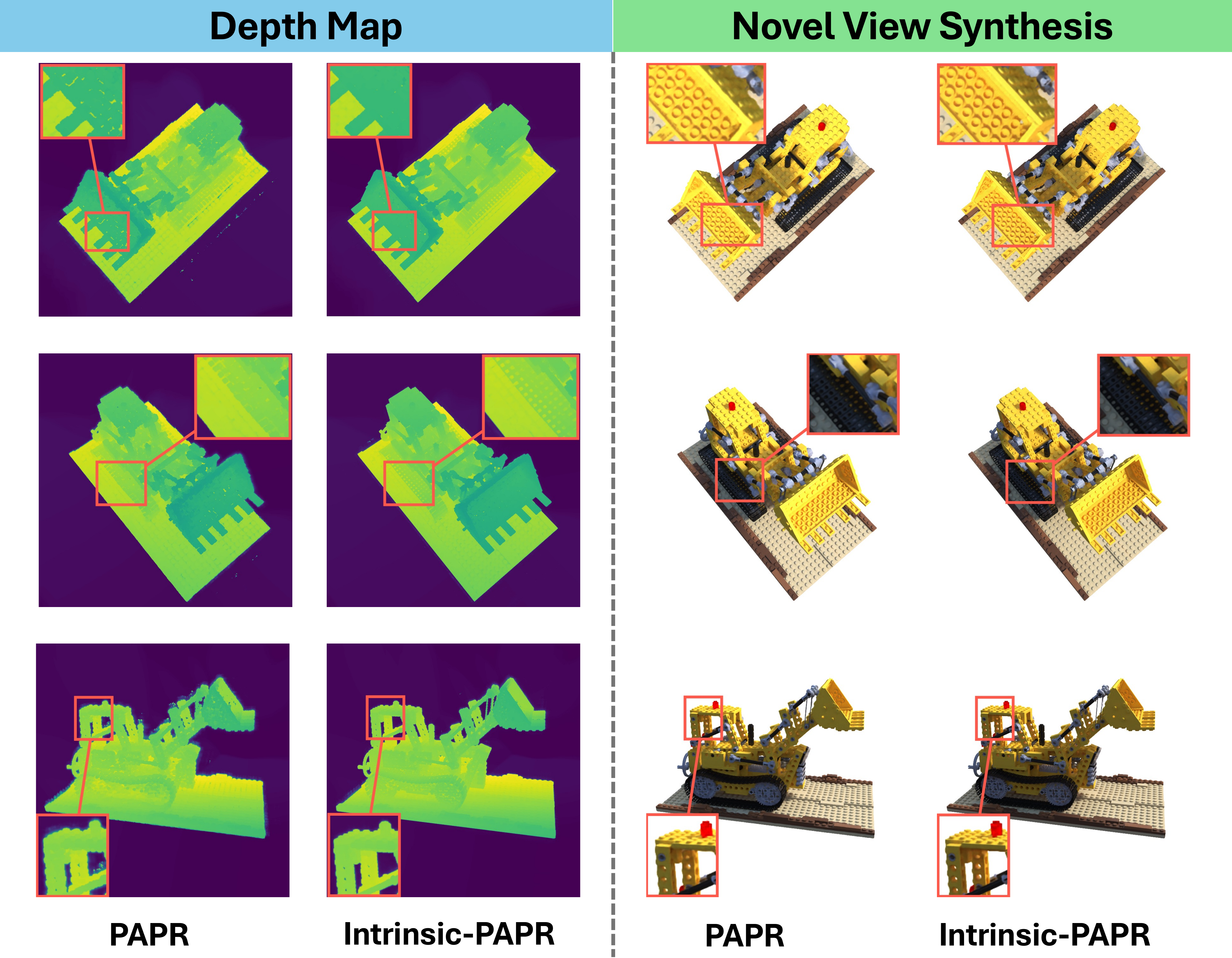}
\caption{Qualitative comparison of depth maps and novel view synthesis between PAPR~\cite{zhang2023papr} and Intrinsic PAPR. Left: depth reconstructions, where Intrinsic PAPR recovers sharper, less blurry geometry that more closely follows the ground-truth structure than PAPR. Right: corresponding novel views, where this more accurate geometry manifests as clearer image details and reduced shading and texture artefacts.}
\Description{Figure with left panels comparing depth reconstructions for PAPR vs.\ Intrinsic PAPR, and right panels comparing their novel view renderings; Intrinsic PAPR exhibits sharper, more accurate geometry and cleaner images.}
  \label{fig:supp-depth-pcd-papr}
\end{figure*}

%% file: tables/supp/albedo_supervision_effect_per_scene.tex
\begin{table*}
    \centering
    \footnotesize
    \caption{\textbf{Detailed per-scene quantitative comparison of baselines with and without albedo supervision:} Training baseline methods with pseudo-ground truth albedo supervision similar to ours does not improve their reconstruction quality.}
    \label{tab:albedo-supervision-effect-per-scene}
    \resizebox{\linewidth}{!}{
    \begin{tabular}{@{}lccccccccccccccc@{}}
        \toprule
        & \multicolumn{3}{c}{Hotdog} & \multicolumn{3}{c}{Lego} & \multicolumn{3}{c}{Materials} & \multicolumn{3}{c}{Ignatius} & \multicolumn{3}{c}{Avg.} \\
        \cmidrule(lr){2-4} \cmidrule(lr){5-7} \cmidrule(lr){8-10} \cmidrule(lr){11-13} \cmidrule(lr){14-16}
        & PSNR$\uparrow$ & SSIM$\uparrow$ & LPIPS$\downarrow$ & PSNR$\uparrow$ & SSIM$\uparrow$ & LPIPS$\downarrow$ & PSNR$\uparrow$ & SSIM$\uparrow$ & LPIPS$\downarrow$ & PSNR$\uparrow$ & SSIM$\uparrow$ & LPIPS$\downarrow$ & PSNR$\uparrow$ & SSIM$\uparrow$ & LPIPS$\downarrow$ \\
        
        \midrule
        \textbf{DPIR~\cite{chung2024differentiable}} \\
        No Albedo Supv. & 25.43 & 0.805 & 0.136 & 27.35 & 0.798 & 0.140 & 26.40 & 0.861 & 0.121 & 18.18 & 0.949 & 0.101 & 24.34 & 0.853 & 0.124 \\
        Albedo Supv. & 22.71 & 0.879 & 0.163 & 25.32 & 0.830 & 0.166 & 24.29 & 0.859 & 0.150 & 16.22 & 0.931 & 0.143 & 22.13 & 0.875 & 0.155 \\
        
        \midrule
        \textbf{Intrinsic-NeRF~\cite{ye2023intrinsicnerf}} \\
        No Albedo Supv. & 34.58 & 0.969 & 0.031 & 29.15 & 0.928 & 0.031 & 26.60 & 0.935 & 0.039 & 26.37 & 0.917 & 0.081 & 29.18 & 0.937 & 0.045 \\
        Albedo Supv. & 34.06 & 0.968 & 0.031 & 30.76 & 0.958 & 0.026 & 24.82 & 0.926 & 0.051 & 26.24 & 0.916 & 0.083 & 28.97 & 0.942 & 0.048 \\
        
        \midrule
        \textbf{GS-IR~\cite{Liang2023GSIR3G}} \\
        No Albedo Supv. & 33.82 & 0.967 & 0.049 & 32.79 & 0.963 & 0.035 & 25.79 & 0.912 & 0.079 & 26.24 & 0.920 & 0.079 & 29.66 & 0.941 & 0.060 \\
        Albedo Supv. & 26.91 & 0.950 & 0.063 & 29.55 & 0.946 & 0.046 & 25.47 & 0.919 & 0.074 & 26.25 & 0.920 & 0.081 & 27.05 & 0.934 & 0.066 \\

        \midrule
        \textbf{IRGS~\cite{irgs2025}} \\
        No Albedo Supv. & 34.02 & 0.973 & 0.032 & 32.85 & 0.969 & 0.027 & 26.46 & 0.922 & 0.049 & 26.91 & 0.934 & 0.071 & 30.06 & 0.950 & 0.045 \\
        Albedo Supv. & 29.23 & 0.963 & 0.046 & 29.54 & 0.945 & 0.056 & 23.87 & 0.852 & 0.068 & 24.56 & 0.921 & 0.097 & 26.80 & 0.920 & 0.067  \\

        \midrule
        \textbf{DiscretizedSDF~\cite{discretizedsdf2025}} \\
        No Albedo Supv. & 34.36 & 0.980 & 0.029 & 32.91 & 0.972 & 0.032 & 27.02 & 0.934 & 0.041 & 27.34 & 0.942 & 0.067 & 30.41 & 0.957 & 0.042 \\
        Albedo Supv. & 31.23 & 0.973 & 0.034 & 30.02 & 0.934 & 0.078 & 25.02 & 0.912 & 0.056 & 25.78 & 0.872 & 0.093 & 28.01 & 0.923 & 0.065 \\

        \bottomrule
    \end{tabular}
    }
\end{table*}

%% file: tables/supp/tab_albedo_consistency.tex
\begin{table*}
  \centering
  \footnotesize
  \caption{\textbf{Per-scene Mean Albedo Consistency Error (MACE) for experiments with and without the Space Carving Loss (SCL):} Lower MACE values indicate higher albedo consistency. The table shows that, for each scene, using SCL results in lower MACE compared to both the model without SCL and the vanilla 2D prior model~\cite{Careaga2024Colorful}, indicating enhanced albedo consistency across different views.}
  \label{tab:abbl-space-carving-albedo-consis-per-scene}
  \resizebox{\linewidth}{!}{%
  \begin{tabular}{@{}lcc|c@{}}
    \toprule
    Scene & Experiment & MACE for Pred. $\downarrow$ & \multicolumn{1}{c}{MACE for vanilla 2D prior} \\
    \midrule
    Lego & No SCL & 0.033 & \multirow{2}{*}{0.032} \\
         & With SCL & \textbf{0.025} & \\
    \cmidrule{1-4}
    Hotdog & No SCL & 0.013 & \multirow{2}{*}{0.016} \\
           & With SCL & \textbf{0.009} & \\
    \cmidrule{1-4}
    Materials & No SCL & 0.015 & \multirow{2}{*}{0.015} \\
              & With SCL & \textbf{0.010} & \\
    \cmidrule{1-4}
    Ficus & No SCL & 0.029 & \multirow{2}{*}{0.029} \\
          & With SCL & \textbf{0.020} & \\
    \cmidrule{1-4}
    Chair & No SCL & 0.011 & \multirow{2}{*}{0.013} \\
          & With SCL & \textbf{0.009} & \\
    \cmidrule{1-4}
    Drums & No SCL & 0.028 & \multirow{2}{*}{0.026} \\
          & With SCL & \textbf{0.023} & \\
    \cmidrule{1-4}
    Mic & No SCL & 0.017 & \multirow{2}{*}{0.016} \\
        & With SCL & \textbf{0.014} & \\
    \cmidrule{1-4}
    Ship & No SCL & 0.026 & \multirow{2}{*}{0.025} \\
         & With SCL & \textbf{0.021} & \\
    \cmidrule{1-4}
    Armadillo & No SCL & 0.004 & \multirow{2}{*}{0.004} \\
              & With SCL & \textbf{0.003} & \\
    \bottomrule
  \end{tabular}
  }
\end{table*}

%% file: figs/supp/render_albedo_shading/fig_render_albedo_shading.tex
\begin{figure*}
  \centering
  \includegraphics[width=\linewidth]{figs/supp/render_albedo_shading/render_albedo_shading.pdf}
  \caption{\textbf{Qualitative Intrinsic Decomposition on Real-world Scenes.} Rendered images alongside their corresponding albedo and shading maps produced by Intrinsic PAPR on representative scenes from the Mip-NeRF 360 dataset~\cite{barron2022mipnerf360}.}
  \Description{Qualitative decomposition results showing rendered images, albedo maps, and shading maps for real-world scenes from the Mip-NeRF 360 dataset, demonstrating plausible intrinsic factorization by Intrinsic PAPR.}
  \label{fig:real-world-decomposition}
\end{figure*}

%% file: figs/supp/fig_albedo_transfer.tex
\begin{figure*}
  \setlength\tabcolsep{1pt}
  \footnotesize
  \centering
  \resizebox{\linewidth}{!}{
  \begin{tabular}{c}
  \includegraphics[width=\linewidth]{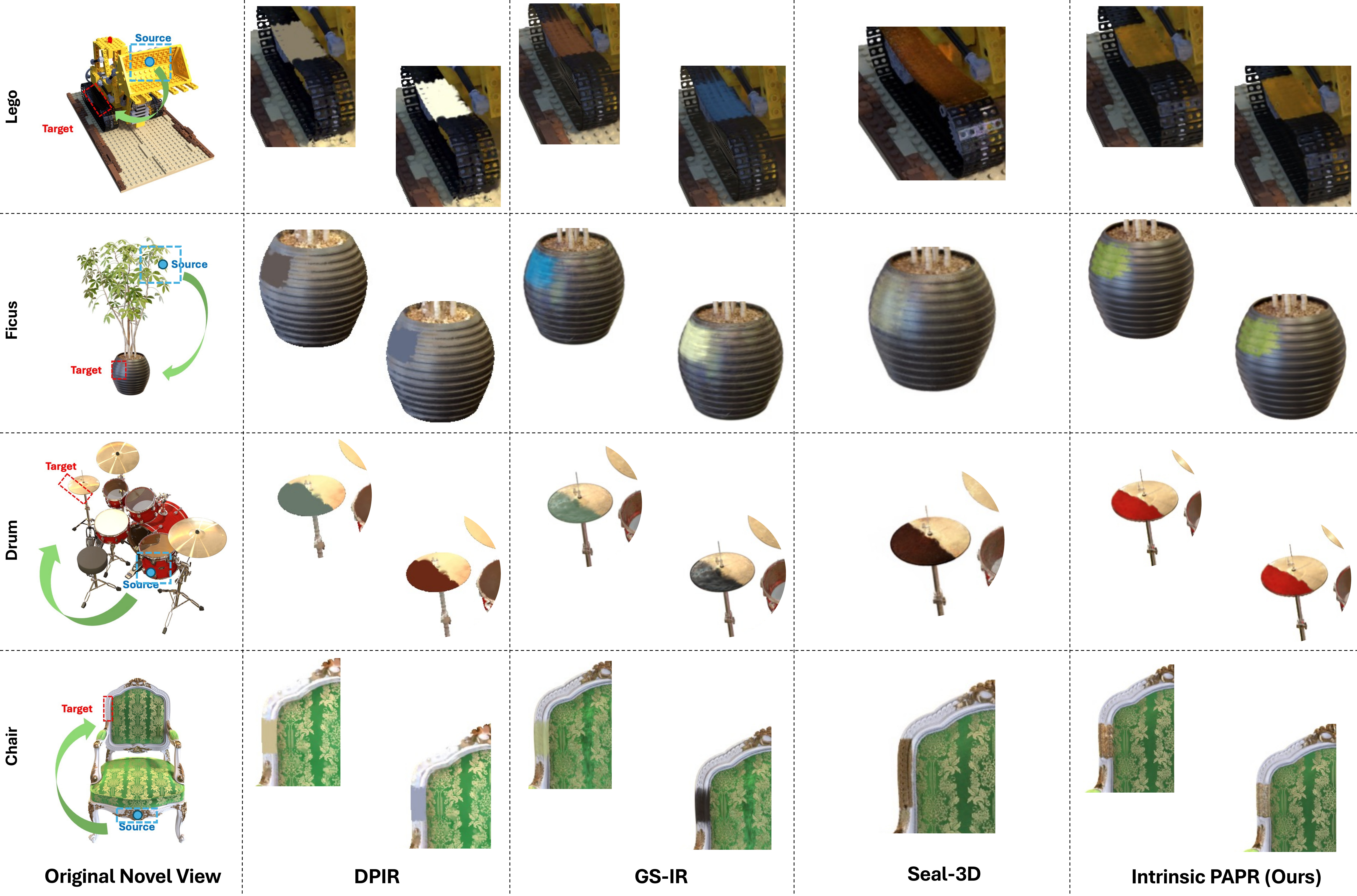}
  \end{tabular}
  }
  \caption{Additional qualitative results for point-level albedo transfer on NeRF Synthetic~\cite{Mildenhall2020NeRFRS}. Our approach accurately assigns albedos at the primitive level, reliably transferring them from several randomly chosen source points to the target region. In contrast, baseline methods often produce visually inconsistent outcomes due to the wrong albedo attribution to the primitives.}
  \Description{Comparison of albedo transfer results showing our method's ability to accurately transfer albedos at the primitive level compared to baseline methods.}
  \label{fig:qualitative-supp-albedo-transfer}
\end{figure*} 

%% file: figs/supp/albedo_editing_MipNeRF/fig_albedo_editing_MipNeRF.tex
\begin{figure*}
    \centering
    \includegraphics[width=\linewidth]{figs/supp/albedo_editing_MipNeRF/garden.pdf} \\
    \includegraphics[width=\linewidth]{figs/supp/albedo_editing_MipNeRF/room.pdf}
    \caption{
        \textbf{Albedo Transfer on Mip-NeRF 360 Scenes}: Qualitative results for point-level albedo transfer on the Mip-NeRF 360 dataset~\cite{barron2022mipnerf360}. In the \textbf{Garden} scene (top), the wooden albedo of specific table panels is replaced with a grass colour. In the \textbf{Room} scene (bottom), a crisscross pattern is transferred onto a section of a plain rug. In both cases, our method successfully alters the surface albedo while perfectly preserving the original shading and geometry, demonstrating coherent editing in complex, unbounded real-world environments.
    }
    \label{fig:qualitative-supp-albedo-transfer-mipnerf}
\end{figure*}

%% file: figs/supp/fig_real_world_albedo_transfer.tex
\begin{figure*}
    \setlength\tabcolsep{1pt}
    \footnotesize
    \centering
    \resizebox{\linewidth}{!}{
        \begin{tabular}{l p{0.3\linewidth} p{0.3\linewidth} p{0.3\linewidth}}
            \rotatebox[origin=c]{90}{Caterpillar}                                                                                 &
            \raisebox{-0.5\height}{\includegraphics[width=\linewidth]{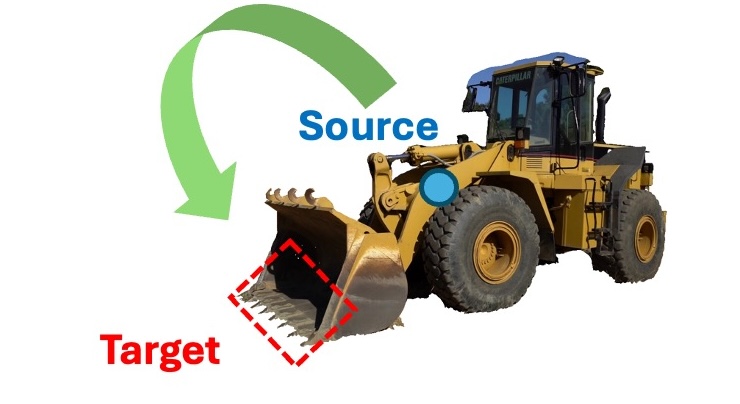}}    &
            \raisebox{-0.5\height}{\includegraphics[width=\linewidth]{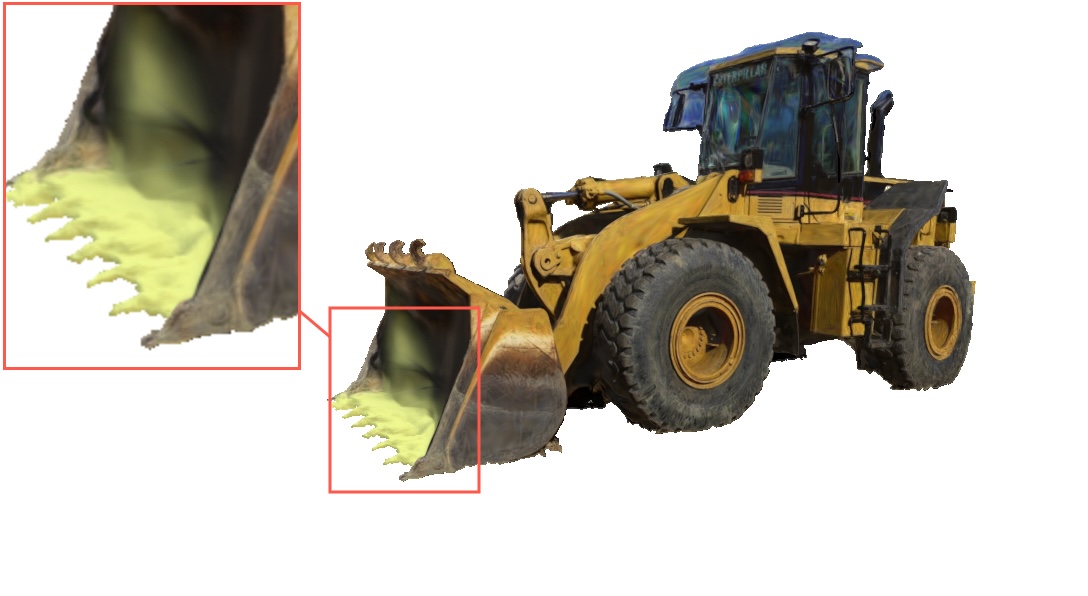}} &
            \raisebox{-0.5\height}{\includegraphics[width=\linewidth]{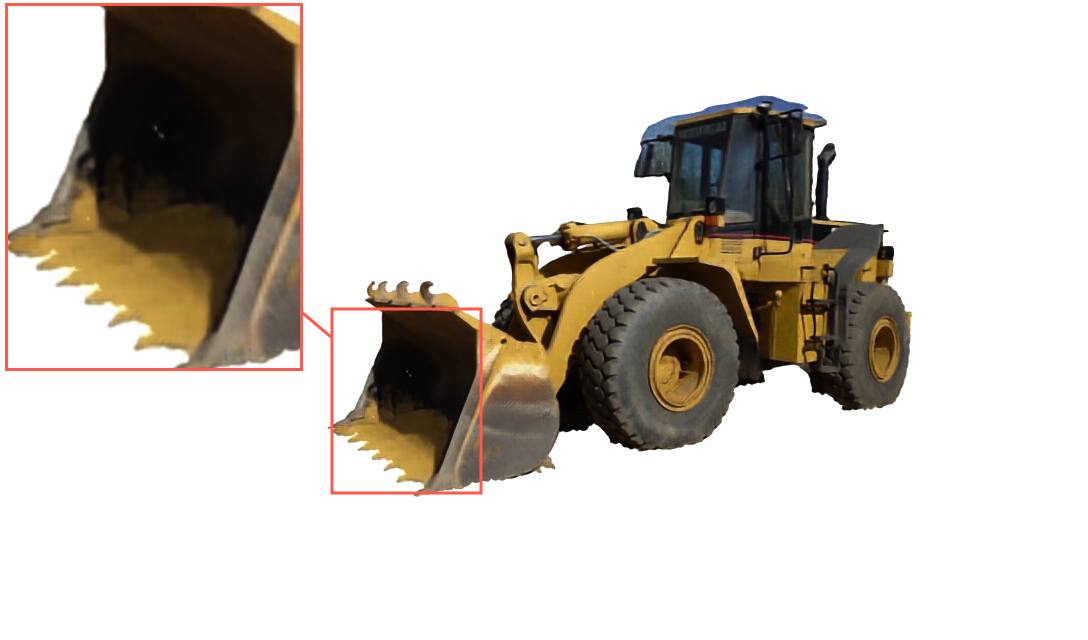}}                                       \\

            \rotatebox[origin=c]{90}{Family}                                                                                      &
            \raisebox{-0.5\height}{\includegraphics[width=\linewidth]{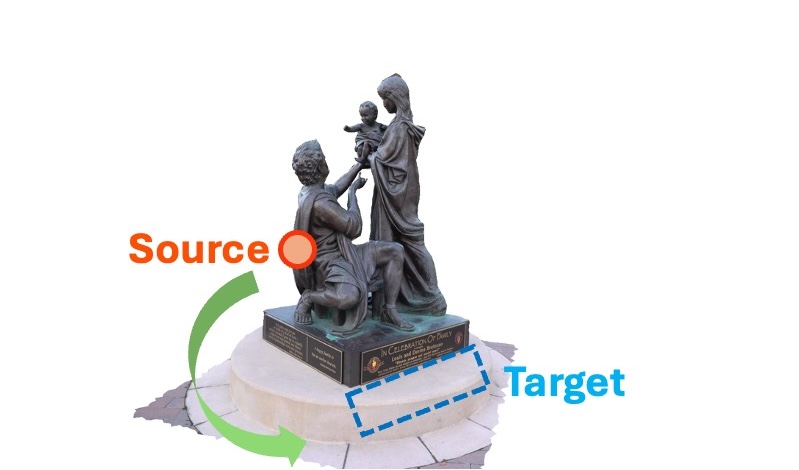}}         &
            \raisebox{-0.5\height}{\includegraphics[width=\linewidth]{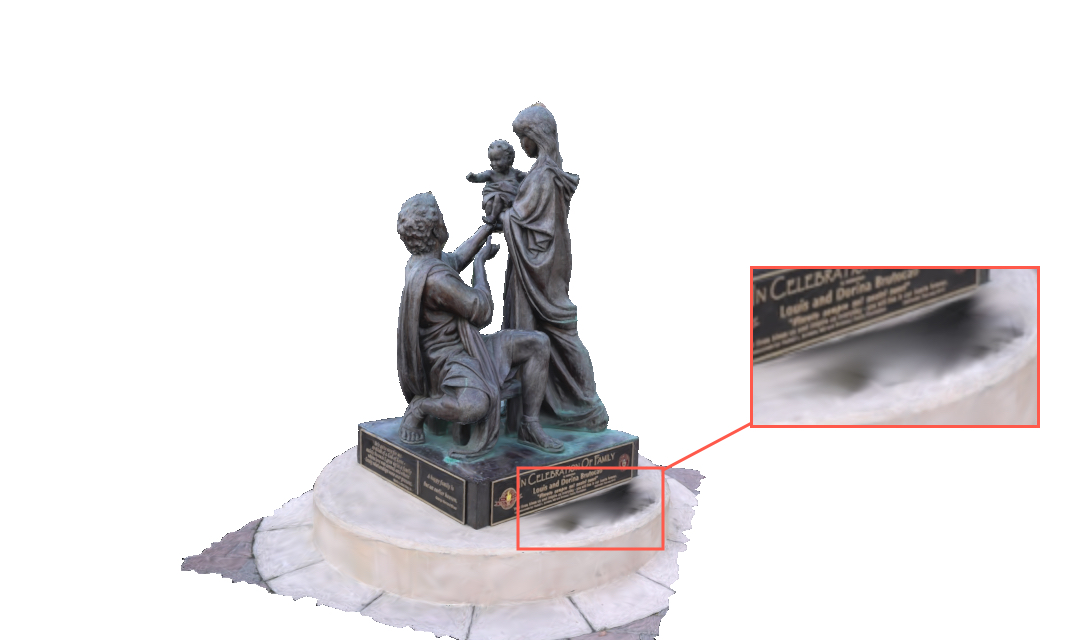}}      &
            \raisebox{-0.5\height}{\includegraphics[width=\linewidth]{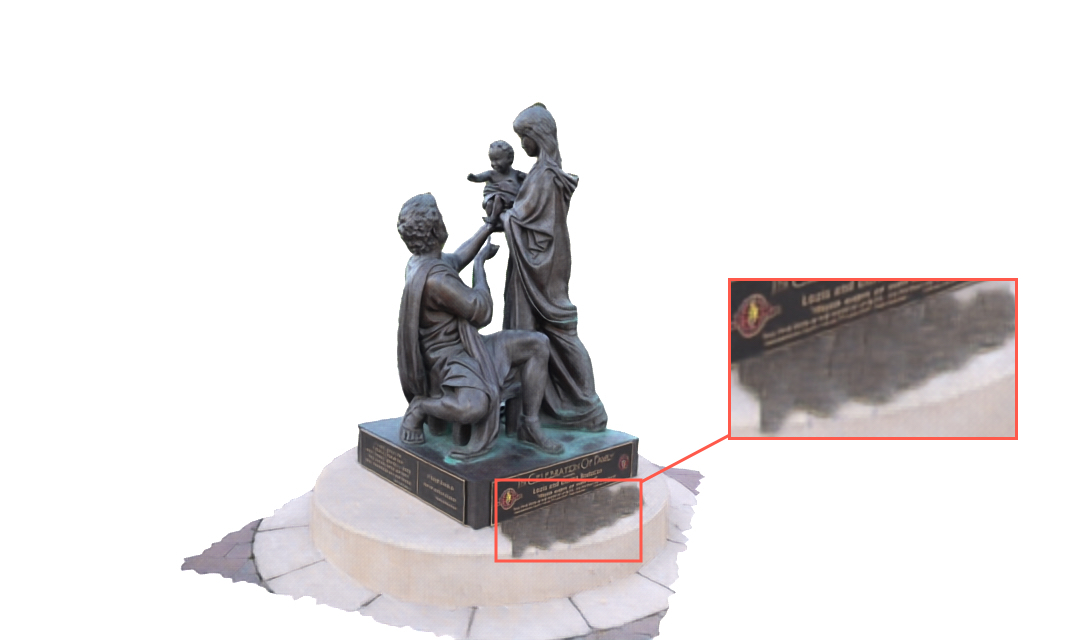}}                                            \\

                                                                                                                                  & \multicolumn{1}{c}{Original View} &
            \multicolumn{1}{c}{GS-IR~\cite{Liang2023GSIR3G}}                                                                      &
            \multicolumn{1}{c}{Intrinsic PAPR (Ours)}                                                                                                                   \\
        \end{tabular}
    }
    \caption{
        \textbf{Albedo Transfer on Real Scenes}: Comparisons for novel view albedo transfer method on the Tanks \& Temples subset~\cite{Knapitsch2017}. The figure shows results for two scenes (Caterpillar and Family), comparing our method with GS-IR in terms of albedo transfer quality and consistency.
    }
    \Description{A comparison table showing albedo transfer results on real-world scenes from the Tanks & Temples dataset. The table displays three columns for each scene (Caterpillar and Family): original view, GS-IR results, and our Intrinsic PAPR method results, demonstrating the effectiveness of our approach in transferring albedo properties across novel views.}
    \label{fig:qualitative-supp-albedo-transfer-TT}
\end{figure*}

%% file: figs/supp/fig_shading_transfer.tex
\begin{figure*}
  \setlength\tabcolsep{1pt}
  \footnotesize
  \centering
  \resizebox{\linewidth}{!}{
  \begin{tabular}{c}
  \includegraphics[width=\linewidth]{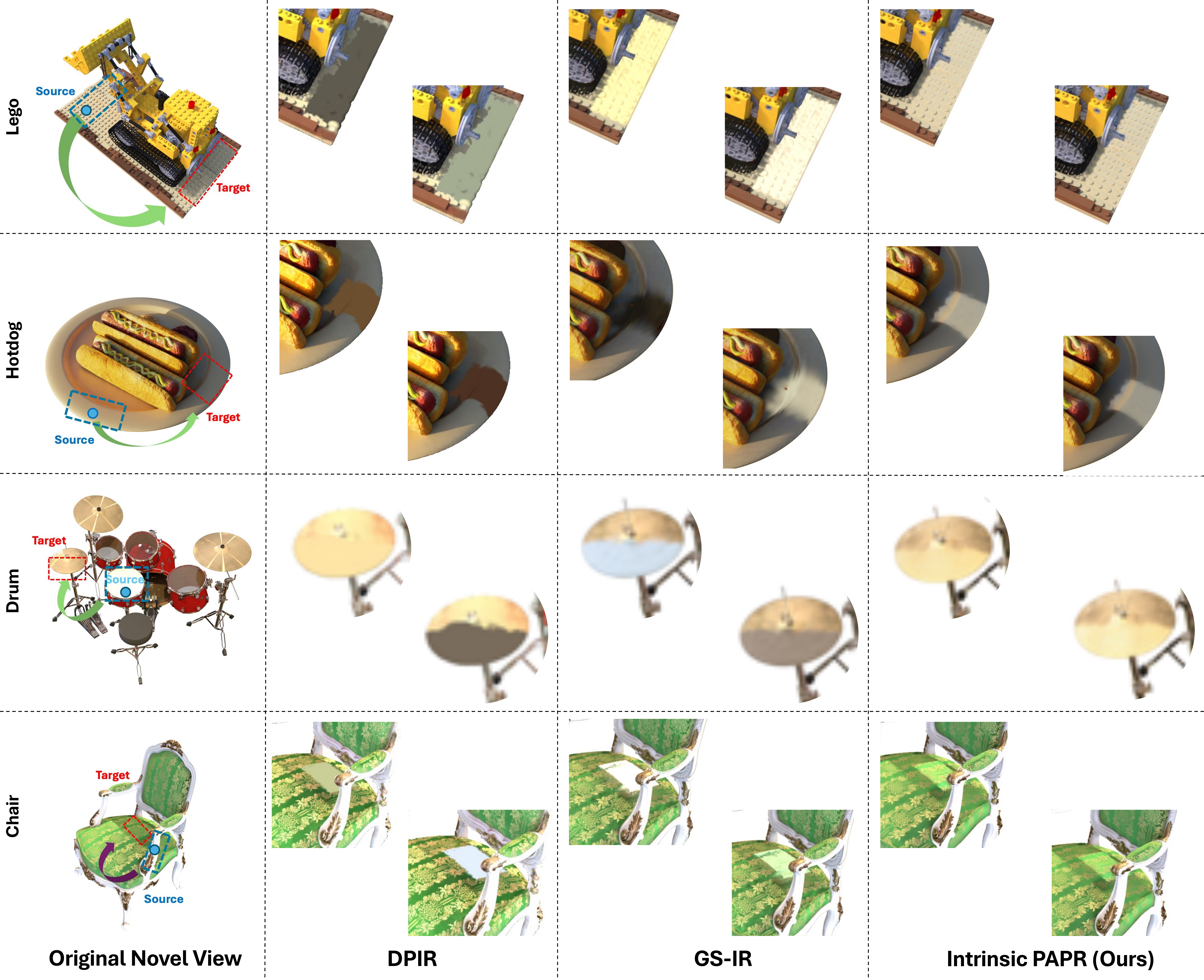}
  \end{tabular}
  }
  \caption{Additional qualitative comparison of shading transfer results on NeRF Synthetic~\cite{Mildenhall2020NeRFRS}. Our approach accurately attributes shading at the primitive level, ensuring consistent and precise transfers, while DPIR and GS-IR exhibit inconsistent patterns and visual artefacts due to the wrong shading attribution to the primitives.}
  \Description{Comparison of shading transfer results showing our method's ability to accurately transfer shading at the primitive level compared to baseline methods.}
  \label{fig:qualitative-supp-shading-transfer}
\end{figure*} 

%% file: figs/supp/fig_real_world_shading_transfer.tex
\begin{figure*}
    \setlength\tabcolsep{1pt}
    \footnotesize
    \centering
    \resizebox{\linewidth}{!}{
        \begin{tabular}{l p{0.3\linewidth} p{0.3\linewidth} p{0.3\linewidth}}
            \rotatebox[origin=c]{90}{Caterpillar}                                                                                  &
            \raisebox{-0.5\height}{\includegraphics[width=\linewidth]{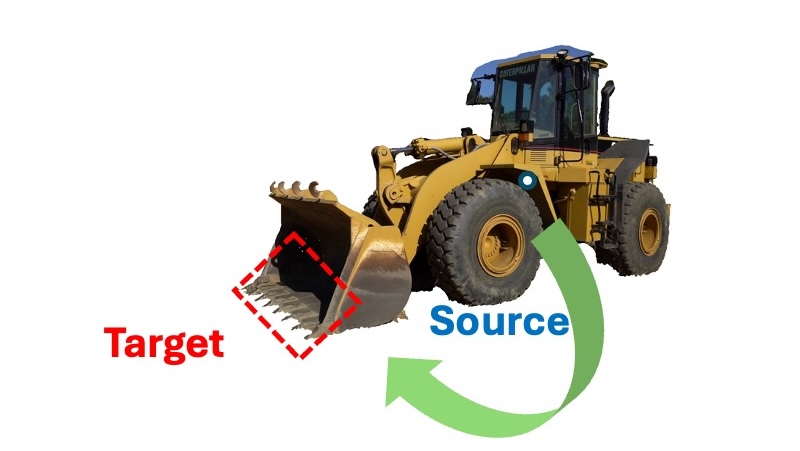}}    &
            \raisebox{-0.5\height}{\includegraphics[width=\linewidth]{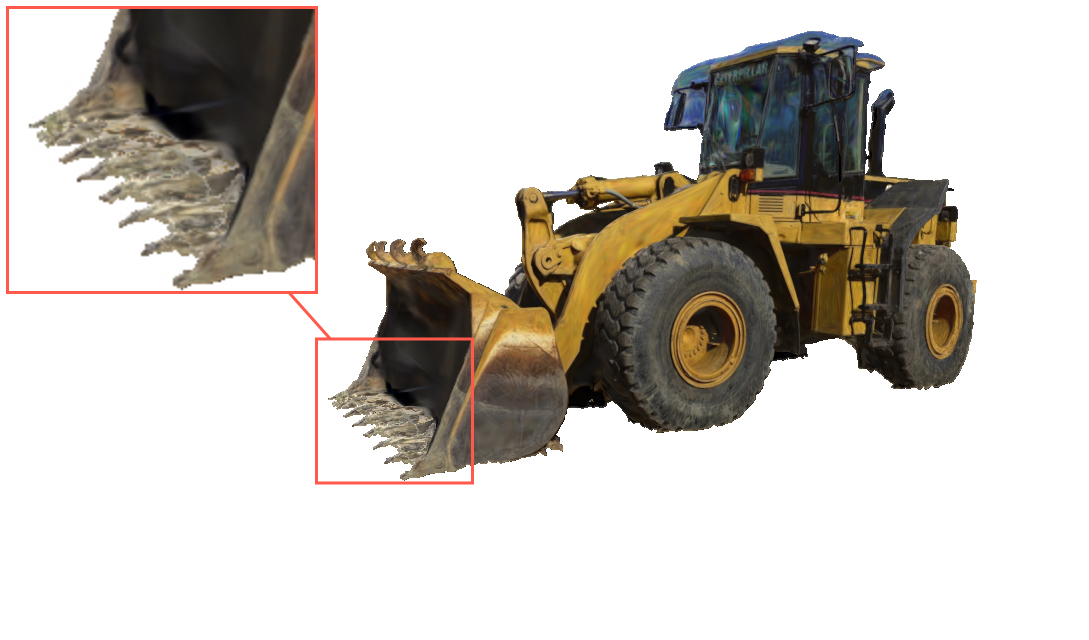}} &
            \raisebox{-0.5\height}{\includegraphics[width=\linewidth]{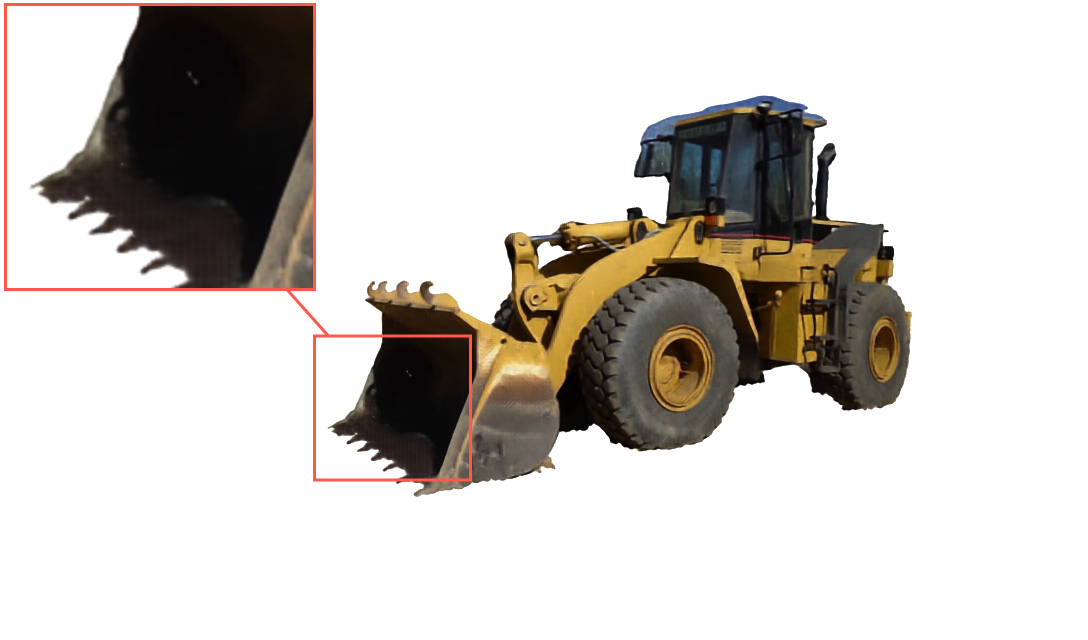}}                                       \\

            \rotatebox[origin=c]{90}{Family}                                                                                       &
            \raisebox{-0.5\height}{\includegraphics[width=\linewidth]{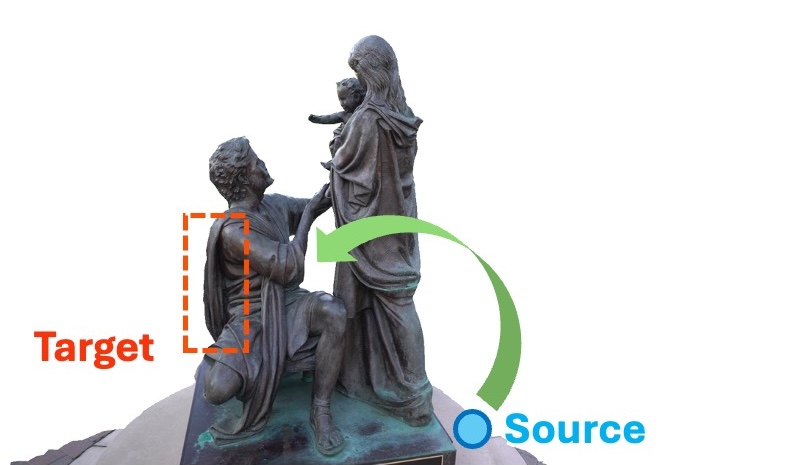}}         &
            \raisebox{-0.5\height}{\includegraphics[width=\linewidth]{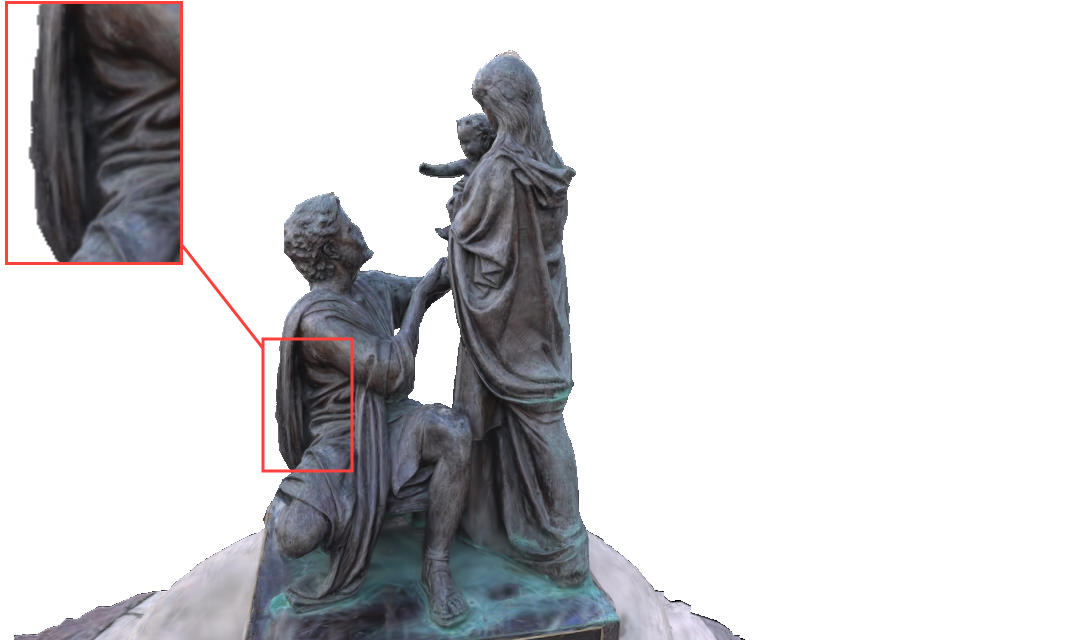}}      &
            \raisebox{-0.5\height}{\includegraphics[width=\linewidth]{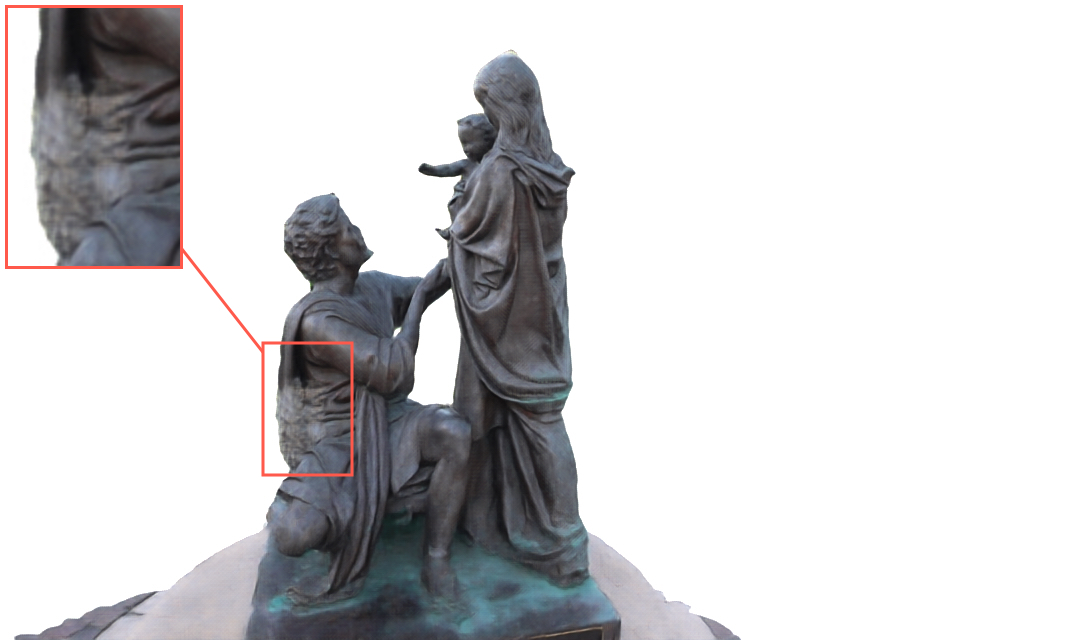}}                                            \\

                                                                                                                                   & \multicolumn{1}{c}{Original View} &
            \multicolumn{1}{c}{GS-IR~\cite{Liang2023GSIR3G}}                                                                       &
            \multicolumn{1}{c}{Intrinsic PAPR (Ours)}                                                                                                                    \\
        \end{tabular}
    }
    \caption{
        \textbf{Shading Transfer on Real Scenes}: Comparisons for novel view shading transfer method on the Tanks \& Temples subset~\cite{Knapitsch2017}. The figure shows results for two scenes (Caterpillar and Family), comparing our method with GS-IR in terms of shading transfer quality and consistency.
    }
    \Description{A comparison table showing shading transfer results on real-world scenes from the Tanks & Temples dataset. The table displays three columns for each scene (Caterpillar and Family): original view, GS-IR results, and our Intrinsic PAPR method results, demonstrating the effectiveness of our approach in transferring shading properties across novel views.}
    \label{fig:qualitative-supp-shading-transfer-TT}
\end{figure*}

%% file: figs/supp/shading_editing_MipNeRF/fig_shading_editing_MipNeRF.tex
\begin{figure*}
    \centering
    \includegraphics[width=\linewidth]{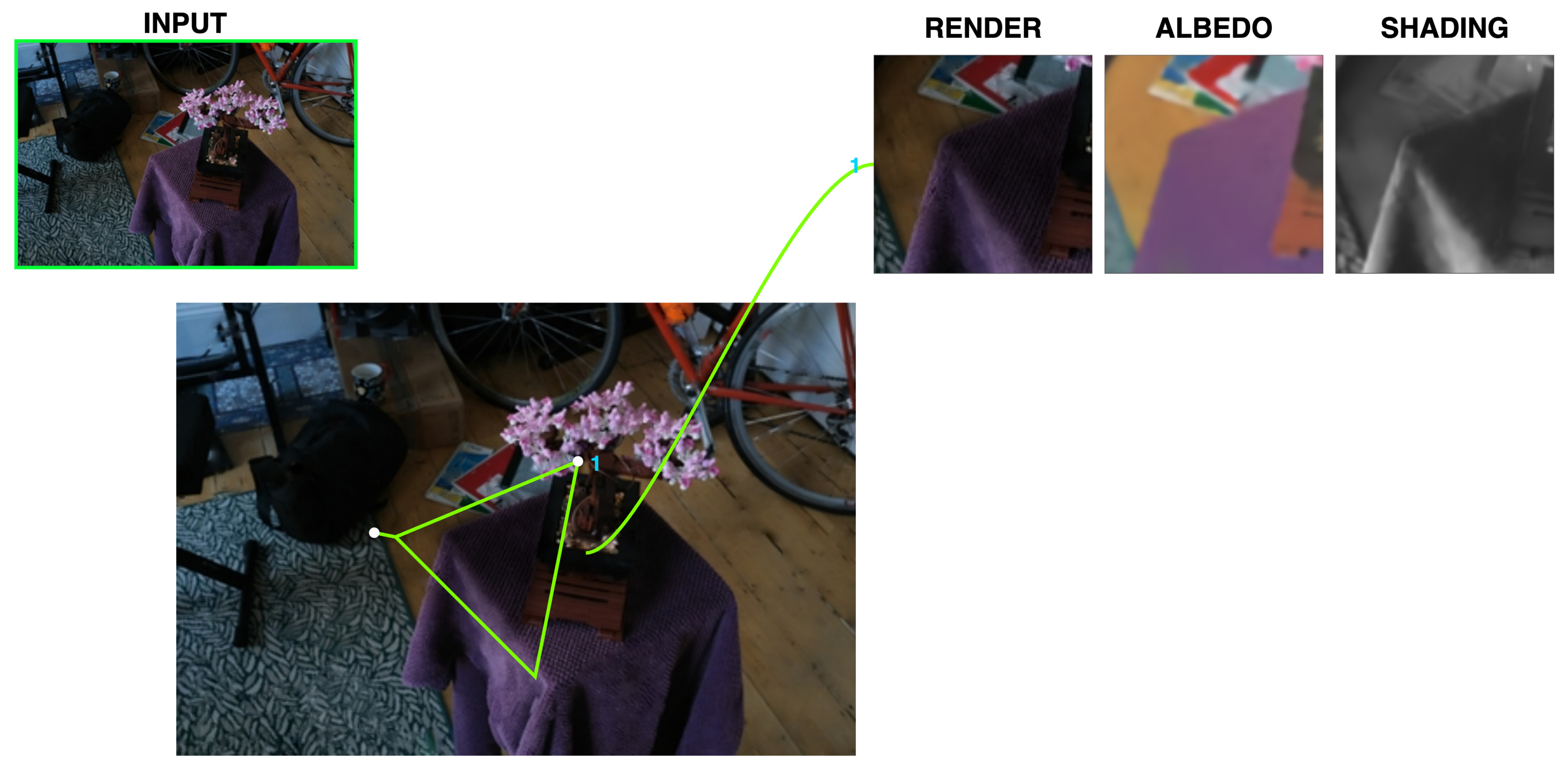} \\
    \includegraphics[width=\linewidth]{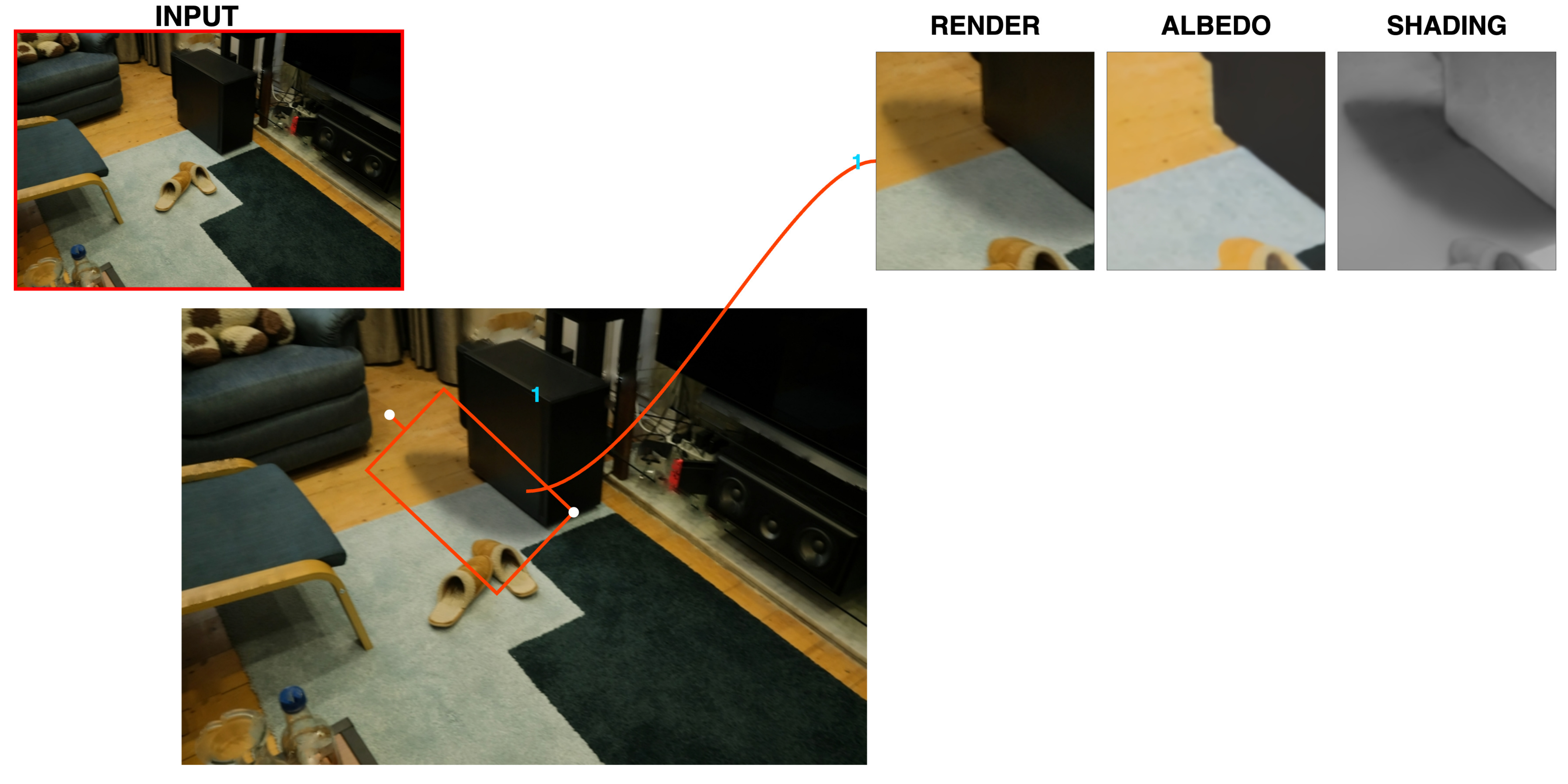}
    \caption{
        \textbf{Shading Transfer on Mip-NeRF 360 Scenes}: Qualitative results for point-level shading transfer on the Mip-NeRF 360 dataset~\cite{barron2022mipnerf360}. In the \textbf{Bonsai} scene (top), a distinct triangular shadow is cast onto the purple cloth base. In the \textbf{Room} scene (bottom), a synthetic shadow cast by the black speaker is extended over both the wooden floor and the light gray rug. In both cases, our method seamlessly composits the new shadow into the shading map while perfectly maintaining the underlying albedo (reflectance and texture), demonstrating coherent editing in complex, unbounded real-world environments.
    }
    \label{fig:qualitative-supp-shading-transfer-mipnerf}
\end{figure*}

%% file: tables/supp/view_count_transfer.tex
\begin{table}[t]
\centering
\small
\caption{Per-primitive transfer error (MSE, averaged over the nine synthetic scenes) versus the number of training views. Our advantage over GS-IR is essentially invariant to view count: the albedo-transfer ratio stays within $5.04$--$5.23\times$, and at 25 views we already beat GS-IR at 200 views, indicating that the misattribution gap is structural rather than a matter of training data.}
\label{tab:view-count-transfer}
\begin{tabular}{@{}llcccc@{}}
  \toprule
  & & 25 views & 50 views & 100 views & 200 views \\
  \midrule
  \multirow{3}{*}{Albedo} & GS-IR & 0.398 & 0.348 & 0.326 & 0.314 \\
   & Ours & \textbf{0.078} & \textbf{0.069} & \textbf{0.064} & \textbf{0.060} \\
   & Ratio & $5.10\times$ & $5.04\times$ & $5.09\times$ & $5.23\times$ \\
  \midrule
  \multirow{3}{*}{Shading} & GS-IR & 0.451 & 0.437 & 0.424 & 0.412 \\
   & Ours & \textbf{0.072} & \textbf{0.063} & \textbf{0.058} & \textbf{0.054} \\
   & Ratio & $6.26\times$ & $6.94\times$ & $7.31\times$ & $7.63\times$ \\
  \bottomrule
\end{tabular}
\end{table}

%% file: figs/supp/fig_shading_control.tex
\begin{figure*}
  \setlength\tabcolsep{1pt}
  \footnotesize
  \centering
  \resizebox{0.9\linewidth}{!}{
  \begin{tabular}{p{0.05\linewidth} p{0.3\linewidth} p{0.3\linewidth} p{0.3\linewidth}}
    & \raisebox{-0.5\height}{\includegraphics[width=\linewidth]{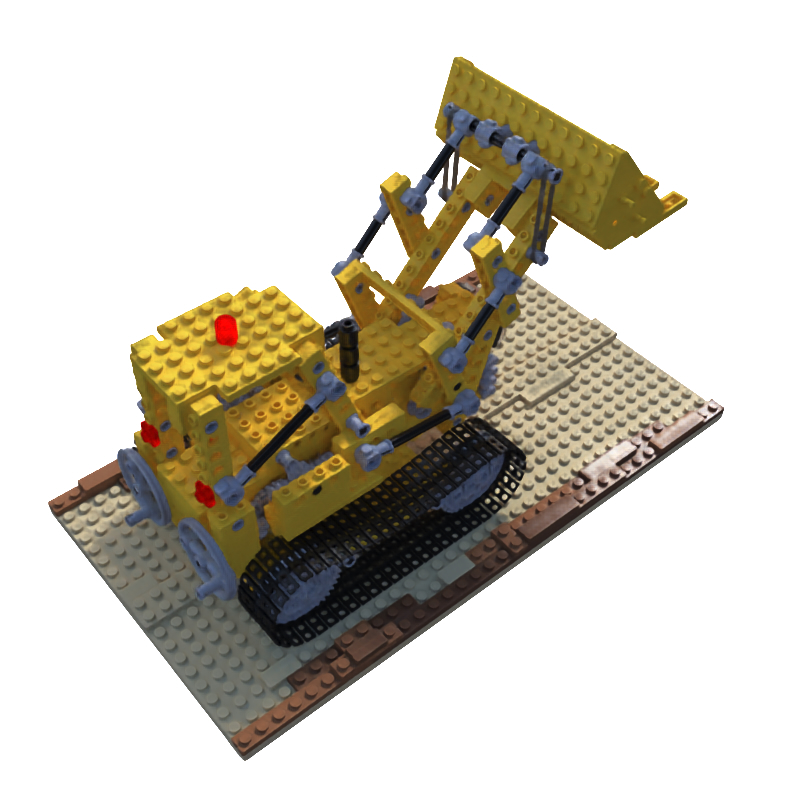}} &
    \raisebox{-0.5\height}{\includegraphics[width=\linewidth]{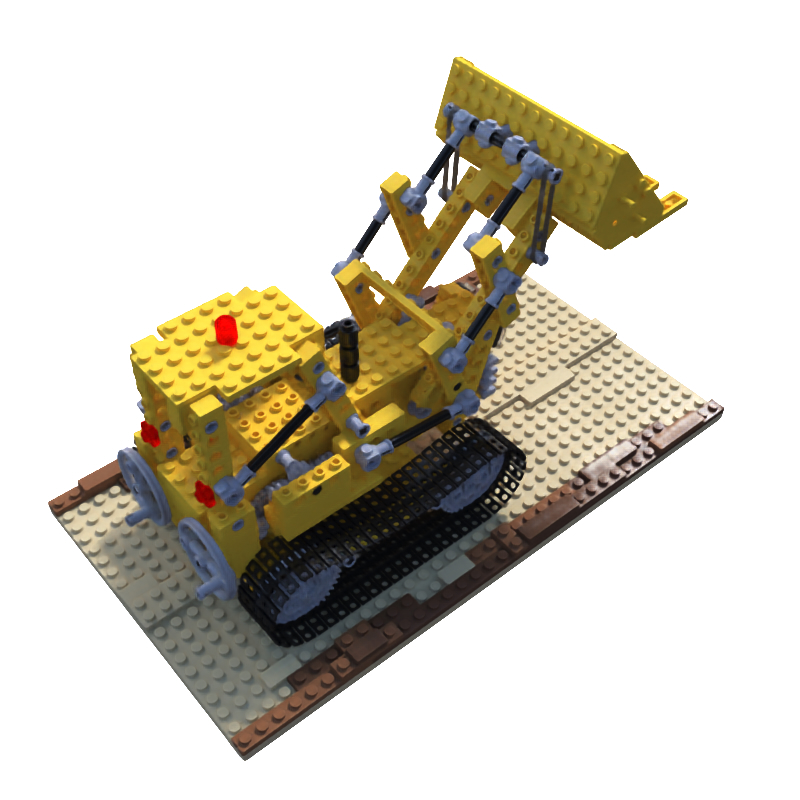}} &
    \raisebox{-0.5\height}{\includegraphics[width=\linewidth]{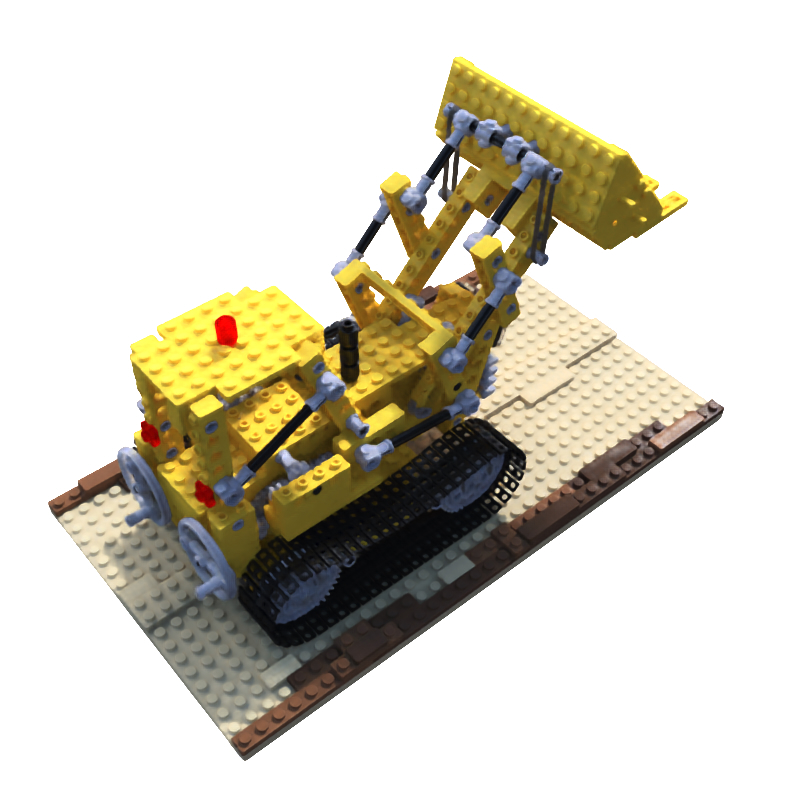}} \\

    & \raisebox{-0.5\height}{\includegraphics[width=\linewidth]{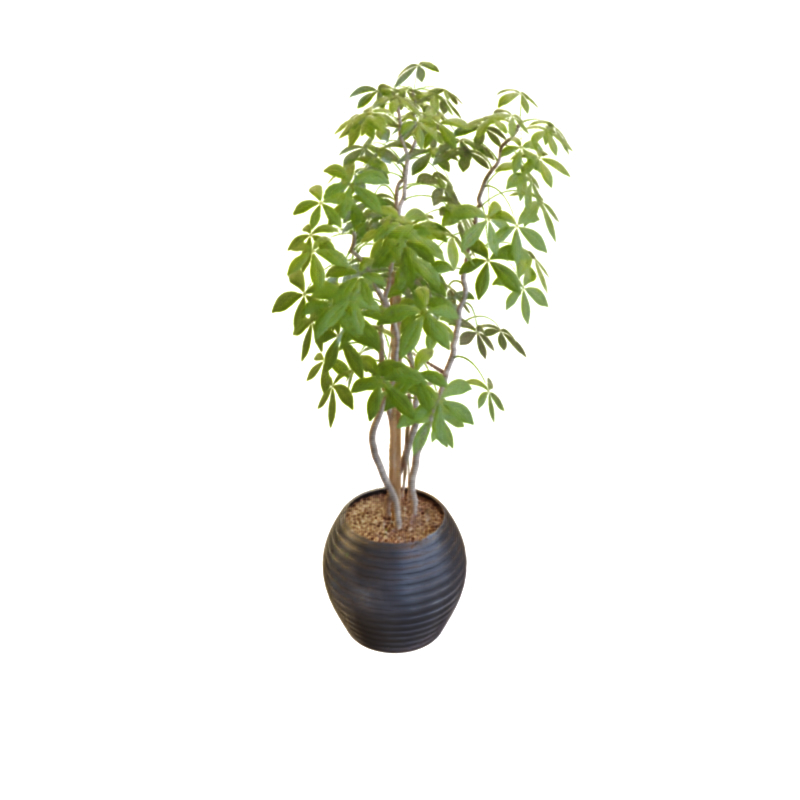}} &
    \raisebox{-0.5\height}{\includegraphics[width=\linewidth]{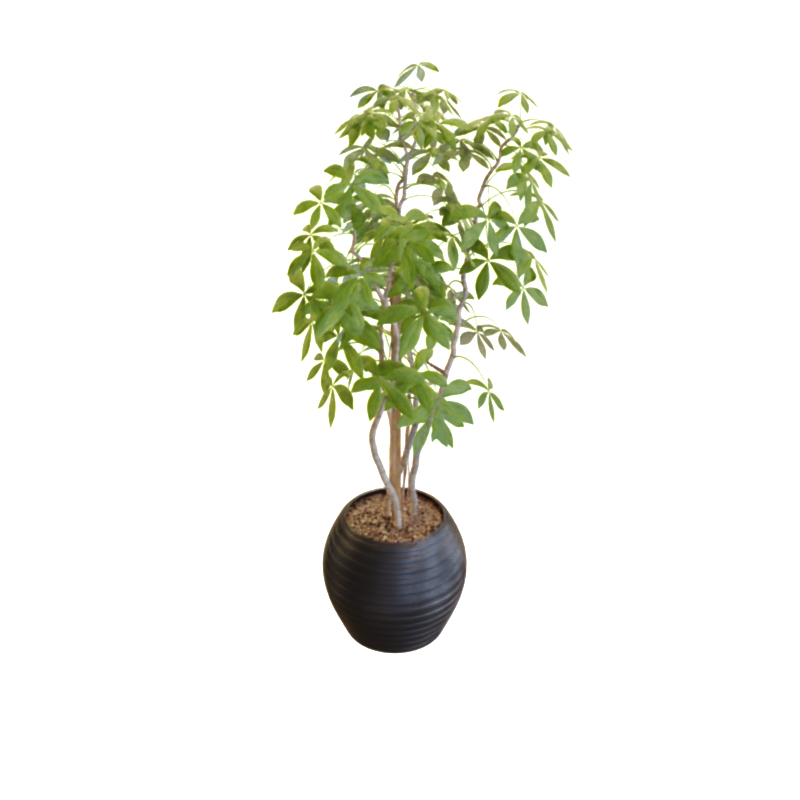}} &
    \raisebox{-0.5\height}{\includegraphics[width=\linewidth]{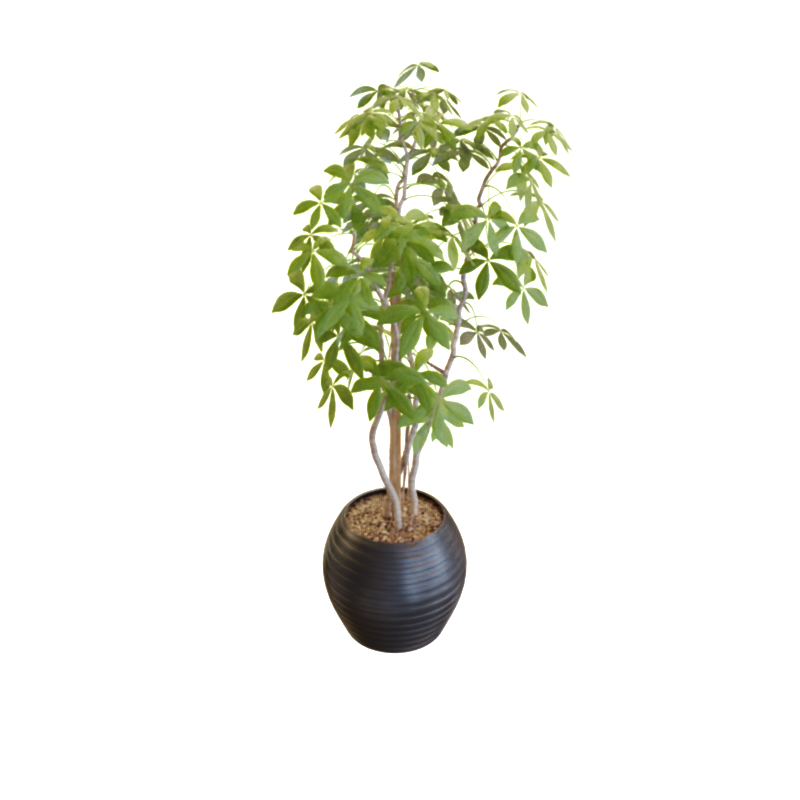}} \\

    & \raisebox{-0.5\height}{\includegraphics[width=\linewidth]{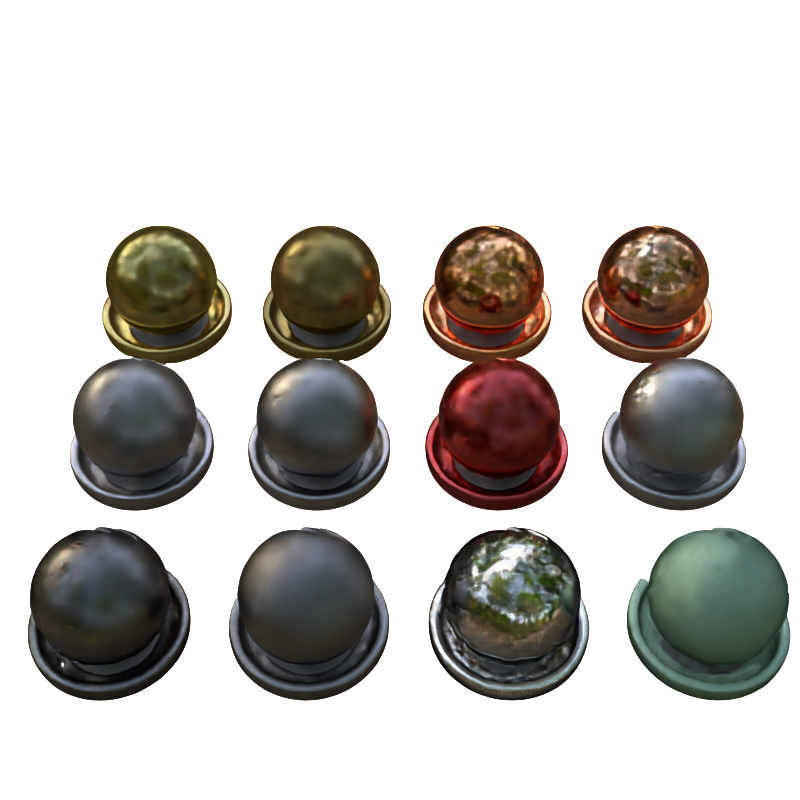}} &
    \raisebox{-0.5\height}{\includegraphics[width=\linewidth]{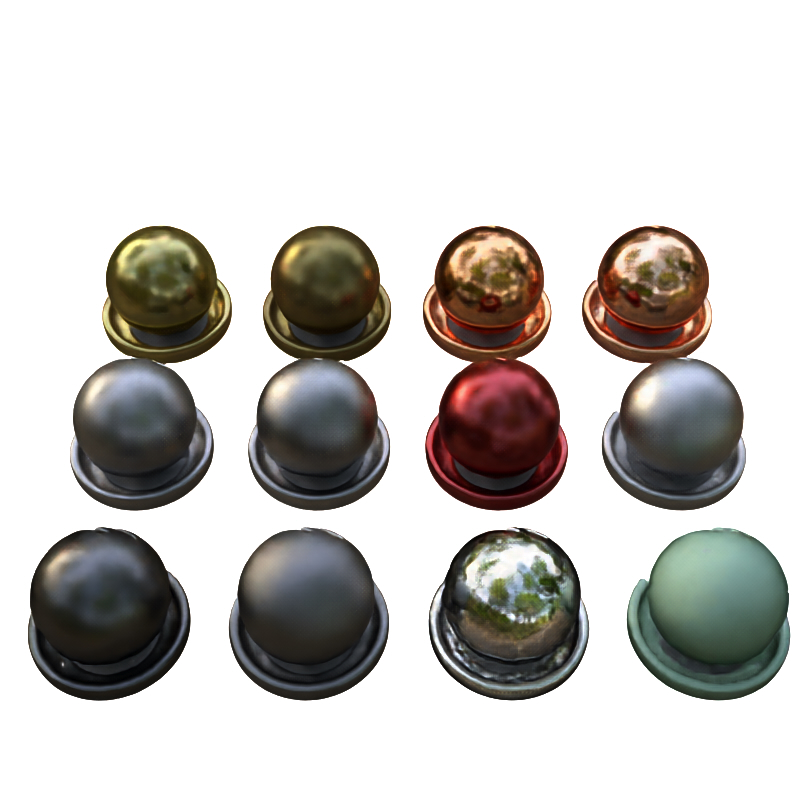}} &
    \raisebox{-0.5\height}{\includegraphics[width=\linewidth]{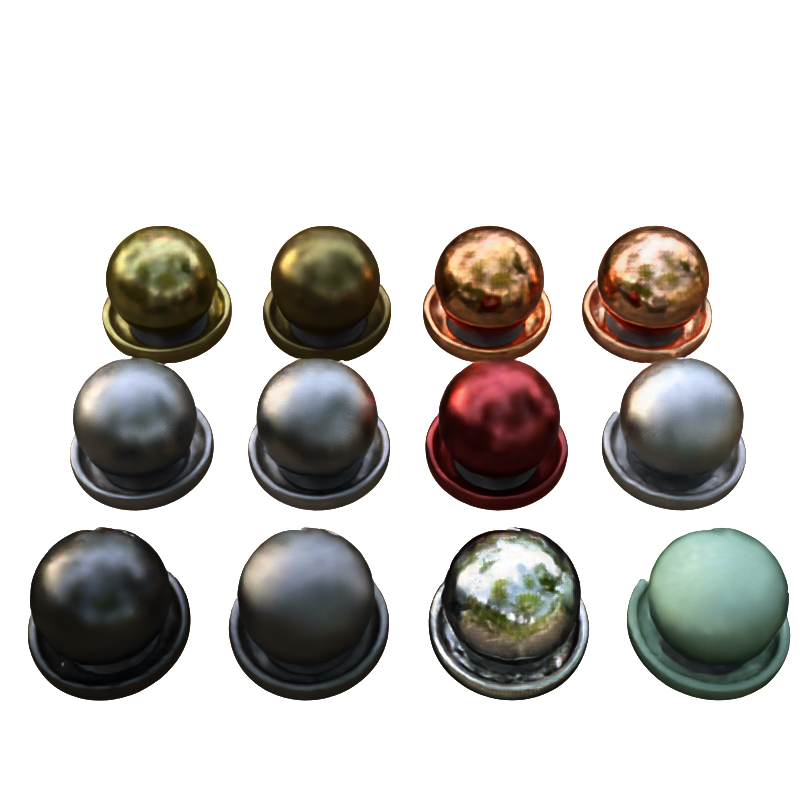}} \\

    & \multicolumn{1}{c}{a. scale = 0.5} & \multicolumn{1}{c}{b. scale = 1.0} & \multicolumn{1}{c}{c. scale = 1.5} \\
  \end{tabular}
  }
  \caption{\textbf{Scene-level Shading Intensity Control}: This figure showcases our method's capability to control the overall shading intensity of an image. By building on our accurate shading attribution in the primitive level, we can now modify the shading intensity across the entire image. This is accomplished by scaling the magnitude of shading features for all points in the image by a consistent factor. Decreasing the scaling factor reduces contrast, while increasing it enhances the contrast within the scene.}
  \Description{Visualization of scene-level shading intensity control showing how our method can adjust the overall shading intensity by scaling factors of 0.5, 1.0, and 1.5.}
  \label{fig:qualitative-supp-shading-control}
\end{figure*} 

%% file: figs/supp/albedo_shading_editing_UI/albedo_shading_editing_tool_fig.tex
\begin{figure*}
    \setlength\tabcolsep{1pt}
    \footnotesize
    \centering
    \resizebox{\linewidth}{!}{
        \begin{tabular}{c}
            \includegraphics[width=\linewidth]{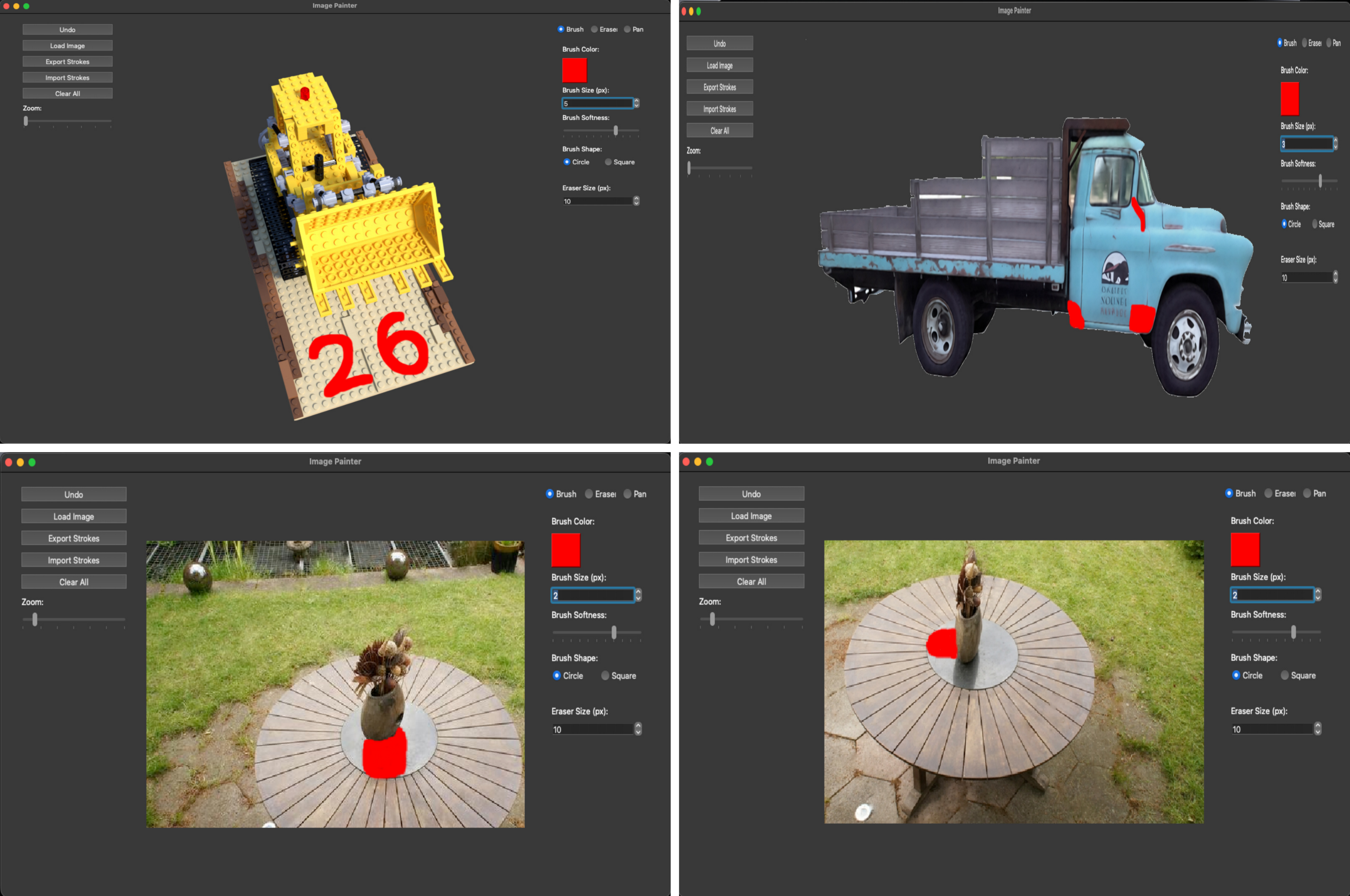}
        \end{tabular}
    }
    \caption{
        \textbf{Custom User Interface for Interactive Scene Editing. }This screenshot illustrates our in-house developed tool for freeform intrinsic decomposition editing. Users can load a scene view, then use brush and eraser tools to define target regions. The tool provides controls for brush colour, size, softness, and shape, enabling precise selection of pixels for feature transfer from a specified source region.
    }
    \Description{\textbf{Custom User Interface for Interactive Scene Editing. }This screenshot illustrates our in-house developed tool for freeform intrinsic decomposition editing. Users can load a scene view, then use brush and eraser tools to define target regions. The tool provides controls for brush colour, size, softness, and shape, enabling precise selection of pixels for feature transfer from a specified source region.}
    \label{fig:albedo-shadow-editing-app}
\end{figure*}

%% file: figs/supp/Special_cases/special_2_cases_fig.tex
\begin{figure*}
    \setlength\tabcolsep{1pt}
    \footnotesize
    \centering
    \resizebox{\linewidth}{!}{
        \begin{tabular}{c}
            \includegraphics[width=\linewidth]{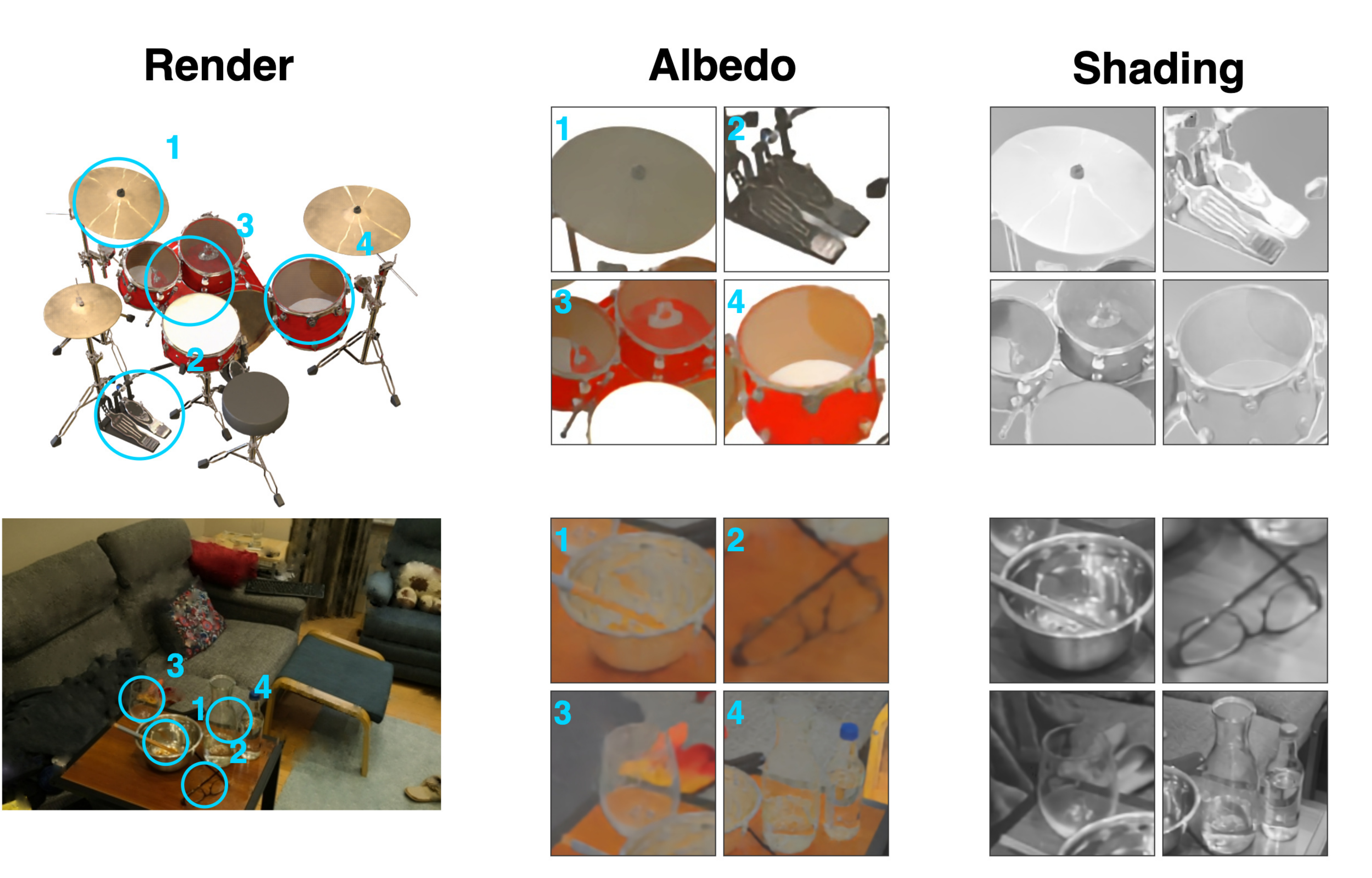}
        \end{tabular}
    }
    \caption{\textbf{Decomposition on Challenging Materials.} We evaluate our method on scenes that stretch the limits of our Lambertian and surface-opacity assumptions. \textbf{Top (Drums):} On glossy-diffuse drum bodies (Regions 3, 4), the decomposition successfully recovers uniform albedo while pushing specular highlights into shading. On near-mirror cymbals (Region 1), the method functionally absorbs the golden reflections into shading, leaving a neutral albedo. \textbf{Bottom (Indoor):} In a complex real-world setting, opaque diffuse surfaces decompose cleanly. Localized artefacts appear at glass boundaries (Region 3) due to the surface-opacity assumption, and coloured inter-reflections (Region 1, 4) occasionally bleed into albedo due to the ordinal shading prior. However, the overall scene composition remains highly coherent.}
    \Description{Renders, albedo, and shading for a Drums scene and a complex indoor scene, highlighting the method's handling of specular surfaces and glass.}
    \label{fig:ood-materials}
\end{figure*}

%% file: figs/supp/fig_limitation.tex
\begin{figure*}
  \setlength\tabcolsep{1pt}
  \footnotesize
  \centering
  \resizebox{\linewidth}{!}{
  \begin{tabular}{c}
  \includegraphics[width=\linewidth]{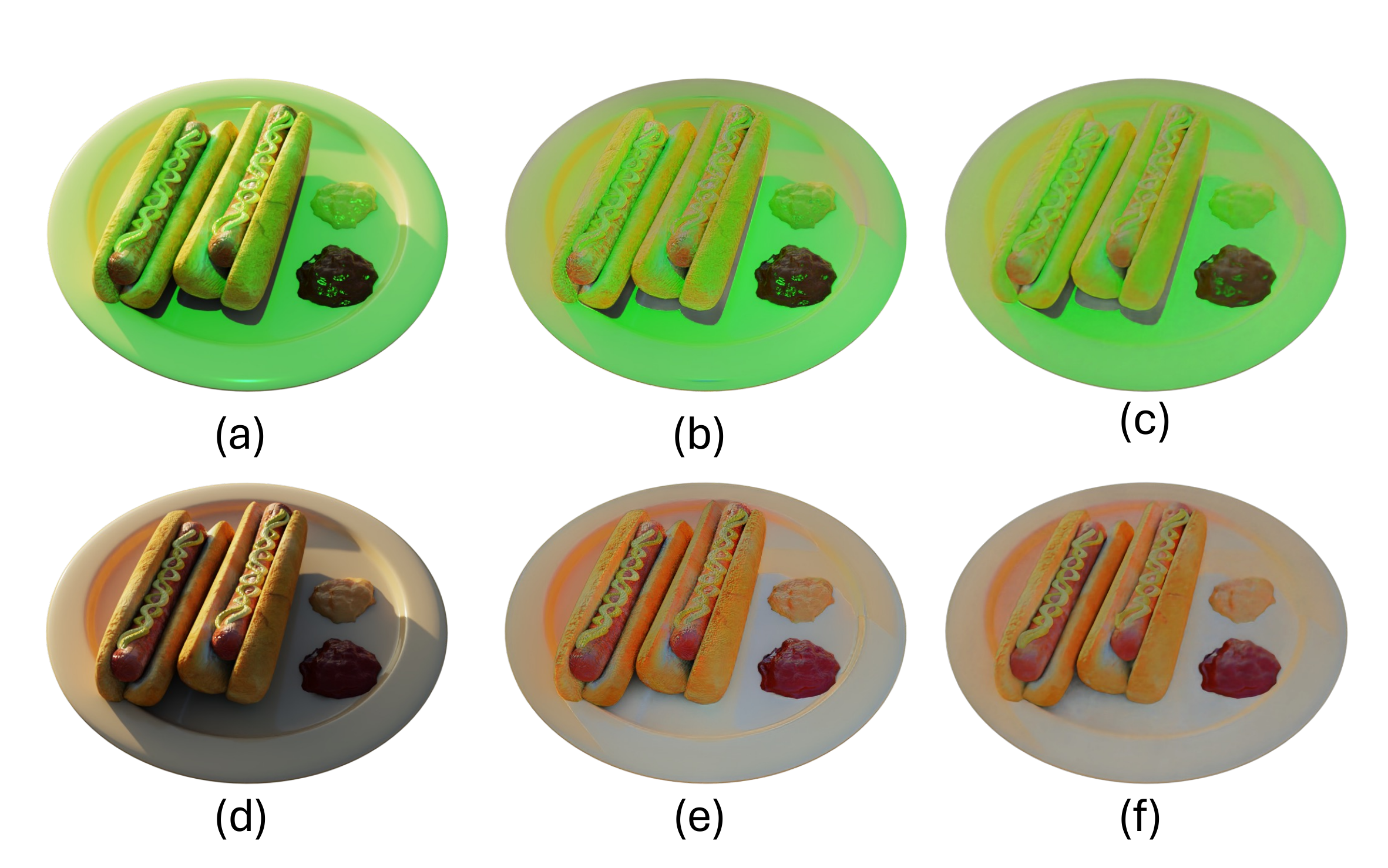}
  \end{tabular}
  }
  \caption{\textbf{Comparison of Albedo Extraction Under Modified and Original Lighting Conditions.} (a) shows the input render with an added green point light. In (b), the vanilla intrinsic decomposition model using ordinal shading baked the green light into the albedo. (c) displays our model's predicted albedo, which inherently baked the light colour to albedo. Under the original lighting setup, panel (d) presents the input view, (e) shows the albedo extracted by the same 2D prior model, and (f) illustrates our model's predicted albedo. This figure highlights how biases in the pseudo-ground-truth data can propagate to our training.}
  \Description{Comparison of albedo extraction results under different lighting conditions, showing how the model's limitations in handling non-monochromatic shading can lead to light colour being baked into albedo.}
  \label{fig:limitation}
\end{figure*}